\documentclass{article}
\pdfoutput=1

\usepackage[final]{corl_2019} 

\usepackage{amsmath}
\usepackage{bm}
\usepackage{amssymb}
\usepackage{graphicx}
\usepackage{subcaption}
\usepackage[makeroom]{cancel}
\usepackage[dvipsnames]{xcolor}

\usepackage{authblk}

\usepackage{natbib}
\newcommand\vf{\mathbf{f}}
\newcommand\vu{\mathbf{u}}
\newcommand\vv{\mathbf{v}}
\newcommand\vw{\mathbf{w}}
\newcommand\vU{\mathbf{U}}

\newcommand\muf{\bm{\mu}_{\mathbf{f}}}
\newcommand\muu{\bm{\mu}_{\mathbf{u}}}
\newcommand\Sff{\bm{\Sigma}_{\mathbf{f}\mathbf{f}}}
\newcommand\Sfu{\bm{\Sigma}_{\mathbf{f}\mathbf{u}}}
\newcommand\Suf{\bm{\Sigma}_{\mathbf{u}\mathbf{f}}}
\newcommand\Suu{\bm{\Sigma}_{\mathbf{u}\mathbf{u}}}

\newcommand\qu{\mathbf{m}_{\mathbf{u}}}
\newcommand\qv{\mathbf{m}_{\mathbf{v}}}
\newcommand\qw{\mathbf{m}_{\mathbf{w}}}
\newcommand\Quu{\mathbf{S}_{\mathbf{u}\mathbf{u}}}
\newcommand\Qww{\mathbf{S}_{\mathbf{w}\mathbf{w}}}

\newcommand\Kff{\textbf{K}_{\mathbf{f}\mathbf{f}}}

\newcommand\Kuu{\textbf{K}_{\mathbf{u}\mathbf{u}}}
\newcommand\Luu{\textbf{L}_{\mathbf{u}\mathbf{u}}}

\newcommand{\kfu}{\textbf{k}_{\cdot\mathbf{u}}}
\newcommand{\kuf}{\textbf{k}_{\mathbf{u}\cdot^\prime}}
\newcommand{\kufalt}{\textbf{k}_{\mathbf{u}\cdot}}

\newcommand{\KL}{\textrm{KL}}
\newcommand{\ELBO}{\textrm{ELBO}}

\DeclareMathOperator{\chol}{chol}
\DeclareMathOperator{\explik}{explik}

\title{A Tutorial on Sparse Gaussian Processes\\and Variational Inference}

\author[1]{Felix Leibfried}
\author[1,2]{Vincent Dutordoir}
\author[1]{ST John}
\author[1]{Nicolas Durrande}
\affil[1]{Secondmind, Cambridge (UK)}
\affil[2]{University of Cambridge, Cambridge (UK)}

\begin{document}
\maketitle


\begin{abstract}
Gaussian processes (GPs) provide a mathematically elegant framework for Bayesian inference and they can offer principled uncertainty estimates for a large range of problems. For example, if we consider certain regression problems with Gaussian likelihoods, a GP model enjoys a posterior in closed form. However, identifying the posterior GP scales cubically with the number of training examples and furthermore requires to store all training examples in memory. In order to overcome these practical obstacles, sparse GPs have been proposed that approximate the true posterior GP with a set of pseudo-training examples (a.k.a.\ inducing inputs or inducing points). Importantly, the number of pseudo-training examples is user-defined and enables control over computational and memory complexity. In the general case, sparse GPs do not enjoy closed-form solutions and one has to resort to approximate inference. In this context, a convenient choice for approximate inference is variational inference (VI), where the problem of Bayesian inference is cast as an optimization problem---namely, to maximize a lower bound of the logarithm of the marginal likelihood. This paves the way for a powerful and versatile framework, where pseudo-training examples are treated as optimization arguments of the approximate posterior that are jointly identified together with hyperparameters of the generative model (i.e.\ prior and likelihood) in the course of training. The framework can naturally handle a wide scope of supervised learning problems, ranging from regression with heteroscedastic and non-Gaussian likelihoods to classification problems with discrete labels, but also problems where the regression or classification targets are multidimensional. The purpose of this tutorial is to provide access to the basic matter for readers without prior knowledge in both GPs and VI\@. It turns out that a proper exposition to the subject enables also convenient access to more recent advances in the field of GPs (like importance-weighted VI as well as interdomain, multioutput and deep GPs) that can serve as an inspiration for exploring new research ideas.
\end{abstract}


\newpage


\tableofcontents
\newpage

\section{Introduction}
\label{sec:intro}

Gaussian Processes (GPs)~\citep{Rasmussen2006} are a natural way to generalize the concept of a multivariate normal distribution. While a multivariate normal distribution describes random variables that are vectors, a GP describes random variables that are real-valued functions defined over some input domain. Imagine the input domain are the real numbers, then the random variable described by a GP can be thought of as a ``vector'' of uncountably infinite extent and with infinite resolution that is ``indexed'' by a real number rather than a discrete index. GPs allow however for a wider range of input domains such as Euclidean vector spaces, but also non-continuous input domains like sets containing graph-theoretical objects or character sequences, and many more.

GPs are a popular tool for regression where the goal is to identify an unknown real-valued function given noisy function observations at some input locations. More precisely, given $N$ input/output tuples $\{(X_n, y_n) \}_{n=1}^N$ where $y_n$ is a scalar and $X_n$ an input from some input domain, the modelling assumption is that the data has been generated by $y_n = f(X_n) + \varepsilon_n$
where $f(\cdot)$ is a real-valued function with input $X_n$ that has been sampled from a GP, and where $\varepsilon_n$ are scalar i.i.d.\ random variables corresponding to observation noise.
In this context, the prior knowledge of the data generation process can be encapsulated in a distribution over $f(\cdot)$ (i.e.\ a prior GP), and the likelihood $p(y_n | f(X_n))$ (i.e.\ the observation model) determines how likely a noisy function observation $y_n$ is given the corresponding noise-free function observation $f(X_n)$. If the observation noise is Gaussian, the posterior process is also a GP that enjoys a closed-form expression. The posterior GP can be inferred via Bayes' rule given the input/output tuples, the likelihood and the prior GP. It turns out that the denominator in Bayes' rule, known as the marginal likelihood that depends on both the prior GP and the likelihood, provides a natural way to identify point estimates for hyperparameters of the generative model that are not subject to inference~\citep{Bishop2006}.

In the case of general regression or classification problems, the exact posterior process is usually no longer a GP\@. In logistic regression for example, the noisy function observations are binary values $y_n \in \{0, 1\}$ (which is why logistic regression is actually a classification problem). The likelihood is a Bernoulli distribution whose mean is obtained by squashing the output of a real-valued function $f(X_n)$, sampled from a GP and evaluated at the input $X_n$, through a sigmoid function $\sigma(f(X_n))$. This yields a probability value $\in [0,1]$ for each input location indicating the probability of the observed noisy function value $y_n$ being $1$.
But even regression problems that have Gaussian likelihoods are not unproblematic. It turns out that computing the exact posterior GP requires to store and invert an $(N \times N)$-matrix that is quadratic in the amount of training data $N$. This means quadratic memory consumption and cubic computational complexity---both of which are infeasible for large data sets.

All of these problems can be addressed with recent advances in the field of GP research: sparse GPs~\citep{Titsias2009}. Sparse GPs limit the amount of pseudo data that is used to represent a (possibly non-Gaussian) posterior process, where the limit is user-defined and determines memory and computational complexity. Intuitively, in a regression problem that has a closed-form solution, the optimal sparse GP should be ``as close as possible'' to the true intractable posterior GP\@. In this regard, ``as close as possible'' can be for example defined as low Kullback-Leibler ($\KL$) divergence between the sparse GP and the true posterior GP, and the goal is to identify the sparse GP's pseudo data in such a way that this $\KL$ becomes minimal. In general, it is not possible to identify such optimal sparse GPs in closed-form solution. One way to approximate optimal sparse GPs is to resort to optimization and gradient-based methods, one particular example of which is variational inference (VI) that is equivalent to minimizing said $\KL$ divergence. Other examples comprise Markov-chain-Monte-Carlo methods and expectation propagation~\citep{Hensman2015b,Bui2017}. However, in this tutorial, we put emphasis on VI due to its popularity and convenience (and to limit the scope).

VI is a particular type of approximate inference technique that translates the problem of Bayesian inference into an optimization objective to be optimized w.r.t.\ the parameters of the approximate posterior. Interestingly, this objective is a lower bound to the logarithm of the aforementioned marginal likelihood, which enables hence convenient joint optimization over hyperparameters (that are not treated as random variables) in addition to the approximate posterior's parameters.
VI does not only provide a principled way to identify approximate posterior processes via sparse GPs when the likelihood is Gaussian, but it also provides a solution to scenarios with arbitrary likelihoods and where the true posterior process is typically not a GP (such as in logistic regression). It turns out that the framework can be readily extended to multimodal likelihood problems, as well as to regression and classification problems where unknown noisy vector-valued functions need to be identified.

The remainder of this manuscript is organized as follows. In Section~\ref{sec:sparse_gps}, we provide an overview over sparse GPs and recent extensions that enable further computational gains. Importantly, Section~\ref{sec:sparse_gps} only explains sparse GPs but outside the scope of approximate inference. This is subject of Section~\ref{sec:vi} giving a general background on VI where we refer to weight space models (such as deep neural networks) to ease the exposition. In Section~\ref{sec:vi_with_svgps}, we combine the previous two sections and elucidate how to do VI with sparse GPs (that are function space models), but we also provide some tricks for practitioners. In the end, we conclude with a summary in Section~\ref{sec:conclusion}.

\section{Sparse Gaussian Processes}
\label{sec:sparse_gps}

Informally, GPs can be imagined as a generalization of multivariate Gaussians that are indexed by a (possibly continuous) input domain rather than an index set. Exact and approximate inference techniques with GPs leverage conditioning operations that are conceptually equivalent to those in multivariate Gaussians. In Section~\ref{sec:mvi_identities}, we therefore provide an overview of the most important conditioning operations in multivariate Gaussians and present their GP counterparts in Section~\ref{sec:gps_and_conditioning}. It turns out that these conditioning operations provide a natural way to express sparse GPs and generalize readily to interdomain GPs (Section~\ref{sec:interdomain_gps}), GPs with multiple outputs (Section~\ref{sec:mo_gps}) and deep GPs consisting of multioutput GPs stacked on top of one another (Section~\ref{sec:deep_gps}).

\subsection{Multivariate Gaussian Identities for Conditioning}
\label{sec:mvi_identities}

The identities presented in this section might evoke the impression of being a bit out of context at first sight but will turn out to be essential for understanding sparse GPs as presented in Section~\ref{sec:gps_and_conditioning} and subsequent sections. We start by noting that conditionals of multivariate Gaussians are Gaussian as well. For that purpose, imagine a multivariate Gaussian $\mathcal{N}$ whose random variables are partitioned into two vectors $\vf$ and $\vu$ respectively. The joint distribution then assumes the following form:
\begin{equation}
\label{eq:joint_mvg}
 \begin{pmatrix} \vf \\ \vu \end{pmatrix} \ \sim \ \mathcal{N} \Bigg(  \begin{pmatrix} \muf \\ \muu \end{pmatrix}, \begin{pmatrix} \Sff & \Sfu \\ \Suf & \Suu \end{pmatrix}  \Bigg),
\end{equation}
where $\muf$ and $\muu$ refer to the marginal means of $\vf$ and $\vu$ respectively, and $\Sff$, $\Sfu$, $\Suf$ and $\Suu$ to (cross-)covariance matrices. The conditional distribution of $\vf$ given $\vu$ can then be expressed as:
\begin{equation}
\label{eq:cond_mvg}
\vf|\vu \ \sim \ \mathcal{N} \Big(  \muf + \Sfu \Suu^{-1} (\vu - \muu), \Sff - \Sfu \Suu^{-1} \Suf \Big).
\end{equation}
Let's imagine that other than the marginal distribution $p(\vu)$ with mean $\muu$ and covariance $\Suu$ from Equation~\eqref{eq:joint_mvg}, there is another Gaussian distribution over $\vu$ with mean $\qu$ and covariance $\Quu$:
\begin{equation}
\label{eq:other_marginal_u}
\vu \ \sim \ \mathcal{N} \Big( \qu, \Quu \Big).
\end{equation}
Denoting the distribution from Equation~\eqref{eq:cond_mvg} as $p(\vf|\vu)$, and the distribution from Equation~\eqref{eq:other_marginal_u} as $q(\vu)$, one obtains a marginal distribution over $\vf$ as $q(\vf) = \int p(\vf|\vu) q(\vu) \mathrm{d} \vu$ that is again Gaussian:
\begin{equation}
\label{eq:sparse_mvg}
\vf \ \sim \ \mathcal{N} \Big(  \muf + \Sfu \Suu^{-1} (\qu - \muu), \Sff - \Sfu \Suu^{-1} (\Suu - \Quu) \Suu^{-1} \Suf \Big).
\end{equation}
A quick sanity check reveals that if we had integrated $p(\vf|\vu)$ with the marginal distribution $p(\vu)$ from Equation~\eqref{eq:joint_mvg} instead of $q(\vu)$ from Equation~\eqref{eq:other_marginal_u}, we would have recovered the original marginal distribution $p(\vf)$ with mean $\muf$ and covariance $\Sff$ from Equation~\eqref{eq:joint_mvg} as expected.

Importantly, Equations~\eqref{eq:joint_mvg} to~\eqref{eq:sparse_mvg} remain valid if we define $\vu$ as $\vu = \bm{\Phi} \vf$ via a linear transformation $\bm{\Phi}$ of the random variable $\vf$. In this case, since $\muf$ and $\Sff$ are given, the only remaining quantities to be identified are the mean $\muu$ and the (cross)-covariance matrices $\Sfu$, $\Suf$ and $\Suu$, yielding:
\begin{eqnarray}
\muu &=&  \bm{\Phi} \muf, \label{eq:muu} \\
\Sfu &=& \Sff \bm{\Phi}^\top = (\bm{\Phi} \Sff)^\top = \Suf^\top, \label{eq:sfu} \\
\Suu &=& \bm{\Phi} \Sff \bm{\Phi}^\top. \label{eq:suu}
\end{eqnarray}
Note that the joint covariance matrix of $\vf$ and $\vu$ is degenerate (i.e.\ singular) because $\vu$ is the result of a linear transformation of $\vf$ and hence completely determined by $\vf$ (which leads to the occurrence of covariance matrix eigenvalues that are 0).

\subsection{Gaussian Processes and Conditioning}
\label{sec:gps_and_conditioning}

A GP represents a distribution denoted as $\mathcal{GP}$ over real-valued functions $f(\cdot): \mathcal{X} \rightarrow \mathbb{R}$ defined over an input domain $\mathcal{X}$ that we assume to be continuous throughout this tutorial (although this does not need to be the case). While multivariate Gaussians represent a distribution over finite-dimensional vectors, GPs represent a distribution over uncountably infinite-dimensional functions---vector indexes in multivariate Gaussian random variables conceptually correspond to specific evaluation points $X \in \mathcal{X}$ in GP random functions. Formally, a GP is defined through two real-valued functions: a mean function $\mu(\cdot): \mathcal{X} \rightarrow \mathbb{R}$ and a symmetric positive-definite covariance function $k(\cdot, \cdot^\prime): \mathcal{X} \times \mathcal{X} \rightarrow \mathbb{R}$ a.k.a.\ kernel~\citep{Rasmussen2006}:
\begin{equation}
\label{eq:def_gp}
f(\cdot) \ \sim \ \mathcal{GP} \Big( \mu(\cdot), k(\cdot, \cdot^\prime) \Big) ,
\end{equation}
where $\mu(\cdot) = \mathbb{E}[f(\cdot)]$ and $k(\cdot, \cdot^\prime) = \mathbb{E}[(f(\cdot) -  \mu(\cdot)) (f(\cdot^\prime) - \mu(\cdot^\prime))] = \text{cov}(f(\cdot), f(\cdot^\prime))$ in accordance with multivariate Gaussian notation. Importantly, if we evaluate the GP at any finite subset $\{X_1, X_2, ..., X_N\}$ of $\mathcal{X}$ with cardinality $N$, we would obtain an $N$-dimensional multivariate Gaussian random variable $\vf$:
\begin{equation}
\label{eq:gp_eval}
\vf \ \sim \ \mathcal{N} \Big(  \muf, \Kff \Big),
\end{equation}
where $\muf$ and $\Kff$ are the mean and covariance matrix obtained by evaluating the mean and covariance function respectively at $\{X_1, X_2, ..., X_N\}$, i.e.\ $\muf[n] = \mu(X_n)$ and $\Kff[n, m] = k(X_n, X_m)$. Here, square brackets conveniently refer to \texttt{numpy} indexing notation for vectors and matrices: $\muf[n]$ refers to the entry at index $n$ of the vector $\muf$, and $\Kff[n, m]$ refers to the entry at row index $n$ and column index $m$ of the matrix $\Kff$. While the notation with subscript $\mathbf{f}$ puts emphasis on the random variable and is standard in the literature, it unfortunately hides away the explicit ``dependence'' on the evaluation points $X_n$ which might confuse readers new to the subject.

Similarly to Equation~\eqref{eq:joint_mvg} from the previous section on multivariate Gaussians, we can partition the uncountably infinite set of random variables represented by a GP into two sets: one that contains a finite subset denoted as $\vu$ evaluated at a finite set of evaluation points $\{Z_1, Z_2, ..., Z_M\} \in \mathcal{X}$, such that $\vu[m]=f(Z_m)$, with mean $\muu$ and covariance matrix $\Kuu$; and one that contains the remaining uncountably infinite set of random variables denoted as $f(\cdot)$ evaluated at all locations in $\mathcal{X}$ except for the finitely many points $Z_m$:
\begin{equation}
\label{eq:joint_gp}
 \begin{pmatrix} f(\cdot) \\ \vu \end{pmatrix} \ \sim \ \mathcal{GP} \Bigg(  \begin{pmatrix} \mu(\cdot) \\ \muu \end{pmatrix}, \begin{pmatrix} k(\cdot, \cdot^\prime) & \kfu \\ \kuf & \Kuu \end{pmatrix}  \Bigg),
\end{equation}
where $\kfu$ and $\kuf$ denote vector-valued functions that express the cross-covariance between the finite-dimensional random variable $\vu$ and the uncountably infinite-dimensional random variable~$f(\cdot)$, i.e.\ $\kfu[m] = k(\cdot,Z_m)$ and $\kuf[m] = k(Z_m, \cdot^\prime)$. Note that $\kfu$ is row-vector-valued whereas $\kuf$ is column-vector-valued, and that $\kfu=\kufalt^\top$ holds. 

One might ask why we have chosen the same notation $f(\cdot)$ in Equation~\eqref{eq:joint_gp} as in Equation~\eqref{eq:def_gp} although both random variables have different evaluation domains ($\mathcal{X}$ without inducing points $Z_m$ versus all of $\mathcal{X}$). The answer is a more careful notation could have been adopted~\citep{Matthews2016} but, technically, Equation~\eqref{eq:joint_gp} is correct and poses a degenerate distribution over $f(\cdot)$ and $\vu$, because $\vu$ is completely determined by $f(\cdot)$ (since $\vu$ is the GP evaluated at the inducing points $Z_m$).

The conditional GP of $f(\cdot)$ conditioned on $\vu$ corresponding to the joint from Equation~\eqref{eq:joint_gp} is obtained in a similar fashion as the conditional multivariate Gaussian from Equation~\eqref{eq:cond_mvg} is obtained from the joint multivariate Gaussian in Equation~\eqref{eq:joint_mvg}, resulting in:
\begin{equation}
\label{eq:cond_gp}
f(\cdot)|\vu \ \sim \ \mathcal{GP} \Big( \mu(\cdot) + \kfu \Kuu^{-1} (\vu - \muu), k(\cdot, \cdot^\prime) - \kfu \Kuu^{-1} \kuf \Big).
\end{equation}
Similarly to the multivariate Gaussian case, one can assume another marginal distribution $q(\vu)$ over $\vu$ as in Equation~\eqref{eq:other_marginal_u} (other than $p(\vu)$ which is the marginal distribution over $\vu$ with mean $\muu$ and covariance $\Kuu$ in accordance with Equation~\eqref{eq:joint_gp}). Denoting the conditional GP from Equation~\eqref{eq:cond_gp} as $p(f(\cdot)|\vu)$ and integrating out $\vu$ with $q(\vu)$ yields $q(f(\cdot)) = \int p(f(\cdot)|\vu) q(\vu) \mathrm{d} \vu$, which is a GP:
\begin{equation}
\label{eq:sparse_mgp}
f(\cdot) \ \sim \ \mathcal{GP} \Big(  \mu(\cdot) + \kfu \Kuu^{-1} (\qu - \muu), k(\cdot, \cdot^\prime) - \kfu \Kuu^{-1} (\Kuu - \Quu) \Kuu^{-1} \kuf \Big),
\end{equation}
and conceptually equivalent to its multivariate Gaussian counterpart from Equation~\eqref{eq:sparse_mvg}. With Equation~(\ref{eq:sparse_mgp}), we have arrived at the definition of a sparse GP as used in contemporary literature~\citep{Titsias2009}. In this context, the evaluation points $Z_m$ are usually called ``inducing points'' or ``inducing inputs'' that refer to pseudo-training examples, and $\vu$ is called ``inducing variable'' conceptually referring to noise-free pseudo-outputs observed at the inducing inputs. 

The number $M$ of inducing points $Z_m$ governs the expressiveness of the sparse GP---more inducing points mean more pseudo-training examples and hence a more accurate approximate representation of a function posterior. However, since $M$ determines the dimension of $\vu$, more inducing points also mean higher memory requirements (as $\Kuu$ needs to be stored) and higher computational complexity (as $\Kuu$ needs to be inverted which is a cubic operation $\mathcal{O}(M^3)$). Practical limitations therefore incentivise low $M$ explaining why the formulation is referred to as a ``sparse'' GP in the first place. Note that this line of reasoning assumes a naive approach for dealing with covariance matrices---there are more recent advances that enable approximate computations based on conjugate-gradient-type algorithms~\citep{Gardner2018}, but this is outside the scope of this tutorial.

At this point, we need to highlight that the notation $q(f(\cdot))$ related to the approximate posterior process (but also the notation $p(f(\cdot)|\vu)$ related to the conditional process) is mathematically sloppy since it colloquially refers to a distribution over functions for which no probability density exists. We will nevertheless continue with this notation occasionally---or with the notation $p(f(\cdot))$ to refer to the prior distribution over $f(\cdot)$---where we feel it makes the subject more easily digestible.

In practice, $q(f(\cdot))$ is used to approximate an intractable posterior process through variational inference---we will learn more about what this means in Section~\ref{sec:vi_with_svgps}. In short and on a high level, variational inference phrases an approximate inference problem as an optimization problem where inducing points $Z_m$, as well as the mean $\qu$ and the covariance $\Quu$ of the inducing variable distribution $q(\vu)$, are treated as optimization arguments that are automatically identified in the course of training. The importance of Equation~(\ref{eq:sparse_mgp}) is substantiated by the fact that it remains valid for interdomain GPs (Section~\ref{sec:interdomain_gps}) and multioutput GPs (Section~\ref{sec:mo_gps}), and forms the central building block of modern deep GPs (Section~\ref{sec:deep_gps}), all of which can be practically trained with variational inference.

\subsection{Interdomain Gaussian Processes}
\label{sec:interdomain_gps}

In the preceding section, when introducing sparse GPs, we have defined $\vu$ as a random variable obtained when evaluating the GP on a set of $M$ inducing points $\{Z_1, Z_2, ..., Z_M\} \in \mathcal{X}$. In line with the second part of Section~\ref{sec:mvi_identities}, we could have alternatively defined $\vu$ differently as $\vu[m] = \int f(X) \phi_m(X) \mathrm{d}X$ through a linear functional on $f(\cdot)$ with help of a set of ``inducing features'' $\{\phi_1(\cdot), \phi_2(\cdot), ..., \phi_M(\cdot) \}$ that are real-valued functions $\phi_m(\cdot): \mathcal{X} \rightarrow \mathbb{R}$~\citep{Lazaro2009}. It turns out that in this case, Equation~\eqref{eq:sparse_mgp} remains valid and the only quantities to be identified are the mean $\muu$ for inducing variables $\vu$ as well as the (cross-)covariances $\kfu$ and $\Kuu$. The mean $\muu$ is a vector of size $M$ and its value $\muu[m]$ at a particular index $m$ is computed as follows:
\begin{eqnarray}
\muu[m] &=& \mathbb{E}\big[\vu[m]\big] \; = \; \mathbb{E} \Bigg[ \int f(X) \phi_m(X) \; \mathrm{d}X  \Bigg] \nonumber =  \int \mathbb{E} \big[ f(X) \big] \; \phi_m(X) \; \mathrm{d}X \nonumber \\ &=& \int \mu(X) \phi_m(X) \; \mathrm{d}X.
\label{eq:muui_id}
\end{eqnarray}

The cross-covariance $\kfu$ is a vector-valued function with $M$ outputs, and the scalar-valued function $\kfu[n]$ at a particular output index $n$ is computed as:
\begin{eqnarray}
\kfu[n] &=& \mathbb{E} \Bigg[ \Bigg( f(\cdot) - \mu(\cdot) \Bigg) \Bigg( \vu[n] - \muu[n] \Bigg) \Bigg] \nonumber \\
&=& \mathbb{E} \Bigg[ \Bigg( f(\cdot) - \mu(\cdot) \Bigg) \Bigg( \int (f(X^\prime) - \mu(X^\prime))  \phi_n(X^\prime) \; \mathrm{d}X^\prime \Bigg) \Bigg] \nonumber \\
&=& \int \mathbb{E} \big[ (f(\cdot) - \mu(\cdot)) (f(X^\prime) - \mu(X^\prime)) \big] \; \phi_n(X^\prime) \; \mathrm{d}X^\prime \nonumber \\
&=& \int k(\cdot, X^\prime) \phi_n(X^\prime) \; \mathrm{d}X^\prime. \label{eq:sfu:i_id}
\end{eqnarray}

Similarly, the covariance $\Kuu$ is an $(M \times M)$-matrix and an individual entry $\Kuu[m, n]$ at row index $m$ and column index $n$ is computed as: 
\begin{eqnarray}
\Kuu[m, n] &=& \mathbb{E} \Bigg[ \Bigg( \vu[m] - \muu[m] \Bigg) \Bigg( \vu[n] - \muu[n] \Bigg) \Bigg] \nonumber \\
&=& \mathbb{E} \Bigg[ \Bigg( \int (f(X) - \mu(X)) \phi_m(X) \; \mathrm{d}X \Bigg) \Bigg( \int (f(X^\prime) - \mu(X^\prime)) \phi_n(X^\prime) \; \mathrm{d}X^\prime \Bigg) \Bigg] \nonumber \\
&=& \int \int \mathbb{E} \big[ (f(X) - \mu(X)) (f(X^\prime) - \mu(X^\prime)) \big] \; \phi_m(X) \phi_n(X^\prime) \; \mathrm{d}X^\prime \; \mathrm{d}X \nonumber \\
&=& \int \int k(X, X^\prime) \phi_m(X) \phi_n(X^\prime) \; \mathrm{d} X^\prime \; \mathrm{d}X . \label{eq:suuij_id}
\end{eqnarray}
We ask the reader at this stage to pause and carefully compare the definition of $\vu$ in this section and Equations~\eqref{eq:muui_id} to~\eqref{eq:suuij_id}, with the definition of $\vu$ at the end of Section~\ref{sec:mvi_identities} on multivariate Gaussians and Equations~\eqref{eq:muu} to~\eqref{eq:suu}. Equations~\eqref{eq:muu} to~\eqref{eq:suu} can be rewritten in \texttt{numpy} indexing notation as:
\begin{eqnarray}
\muu[m] &=&  \sum_i \muf[i] \bm{\phi}_m[i], \label{eq:muui} \\
\Sfu[:, n] &=& \sum_j \Sff[:, j] \bm{\phi}_n[j], \label{eq:sfu:i} \\
\Suu[m, n] &=& \sum_i \sum_j \Sff[i, j] \bm{\phi}_m[i] \bm{\phi}_n[j]. \label{eq:suuij}
\end{eqnarray}
This provides a good intuition of how interdomain GPs relate to linear transformations of random variables in the multivariate Gaussian case: mean functions and covariance functions correspond to mean vectors and covariance matrices, inducing features to feature vectors, and integrals over a continuous input domain to sums over discrete indices. Mathematically, $\int f(X) \phi_m(X) \mathrm{d}X$ is a real-valued linear integral operator (which maps functions $f(\cdot)$ to real numbers) that generalizes the concept of a real-valued linear transformation operating on finite-dimensional vector spaces.

After obtaining a mathematical intuition for interdomain GPs and how they relate to linear transformations in multivariate Gaussians, it can be insightful to get a conceptual intuition with concrete examples. An important characteristic of the interdomain formulation is that inducing points can live in a domain different from the one in which the GP operates (which is $\mathcal{X}$), hence the naming. Of practical importance is also the question how to choose the features $\phi_m(\cdot)$ such that the covariance formulations from Equations~\eqref{eq:sfu:i_id} and~\eqref{eq:suuij_id} have closed-form expressions (the mean formulation from Equation~\eqref{eq:muui_id} evaluates trivially to zero for a zero-mean function that assigns zero to every input location $X$, which is a common practical choice). 

In the following, we present four examples: Dirac, Fourier, kernel-eigen and derivative features. Dirac features recover ordinary inducing points as a special case of the interdomain formulation. Fourier features enable inducing points to live in a frequency domain (different from $\mathcal{X}$ that is considered as time/space domain in this context). Kernel-eigen features are conceptually equivalent to principal components of a finite-dimensional covariance matrix but for uncountably-infinite dimensional kernel functions. Derivative features enable inducing points to evaluate the derivative of~$f(\cdot)$ w.r.t.\ specific dimensions of $X$, as opposed to the the ordinary inducing point formulation that enables inducing points to evaluate $f(\cdot)$.

\paragraph{Dirac Features.}
Dirac features are trivially defined as $\phi_m(\cdot) = \delta_{Z_m}(\cdot)$ through the Dirac delta function $\delta_{Z_m}(\cdot)$ that puts all probability mass on $Z_m$. This makes inducing points live in $\mathcal{X}$ as expected, and the inducing mean $\muu$, the cross covariance function $\kfu$ as well as the covariance matrix $\Kuu$ recover the expressions from Section~\ref{sec:gps_and_conditioning} for ordinary sparse GPs. Since Dirac features recover the ordinary inducing point formulation through a linear integral operator, it becomes clear that we can choose the same notation $f(\cdot)$ for both random variables in Equations~\eqref{eq:joint_gp} and~\eqref{eq:def_gp}, without worrying too much about one input domain being $\mathcal{X}$ excluding finitely many inducing points and the other one all of $\mathcal{X}$. We want to stress here once again that Dirac features are of no practical purpose and we only use them to outline how the interdomain formulation generalizes the concept of inducing points $Z_m$ through inducing features $\phi_m(\cdot)$.

\paragraph{Fourier Features.}
Fourier features are defined as $\phi_m(\cdot) = \exp(-\mathrm{i} \bm{\omega}_m^\top \cdot)$ where $\bm{\omega}_m$ refers to an inducing frequency vector and~$\mathrm{i}$ to the complex unit. Note that, practically, boundary conditions need to be defined for Fourier features. Otherwise, $\Kuu$ would have infinite-valued entries on its diagonal for any stationary kernel (stationary kernels are a specific type of kernel where the covariance between two input locations $X$ and $X^\prime$ only depends on the distance between $X$ and $X^\prime$). Also note that for specific stationary kernels, $\kfu$ and $\Kuu$ have real-valued closed-form expressions---see~\cite{Hensman2018} for details. Fourier features underpin the term ``interdomain'' because the integral operator enables inducing points $\bm{\omega}_m$ to live in a frequency domain different from the time/space domain $\mathcal{X}$. While \cite{Hensman2018} chose fixed inducing frequencies $\bm{\omega}_m$ arranged in a grid-wise fashion, an interesting future research direction is to treat $\bm{\omega}_m$ as optimization arguments in the context of approximate inference (e.g.\ variational inference as explained later in Section~\ref{sec:vi}). We need to stress that the original Fourier feature formulation from~\cite{Hensman2018} does not use the $L^2$ inner product between $f(\cdot)$ and $\phi_m(\cdot)$ to define inducing variables~$\vu$ as we did at the beginning of this section, but introducing alternative ways how to define the inner product between two functions is outside the scope of this tutorial.

\paragraph{Kernel-Eigen Features.}
Kernel-eigen features are given as $\phi_m(\cdot) = v_m(\cdot)$ where $v_m(\cdot)$ refers to the $m$-th eigenfunction of the kernel $k(\cdot, \cdot^\prime)$ with eigenvalue $\lambda_m$. Eigenfunctions of kernels are defined as $\int k(\cdot, X^\prime) v_m(X^\prime) \mathrm{d}X^\prime = \lambda_m v_m(\cdot)$, similarly to eigenvectors of square matrices when replacing integrals over $X^\prime$ with sums over indexes. Equation~\eqref{eq:sfu:i_id} then trivially evaluates to:
\begin{eqnarray}
\kfu[n] &=& \int k(\cdot, X^\prime) v_n(X^\prime) \; \mathrm{d}X^\prime = \lambda_n v_n(\cdot) ,
\label{eq:sfu:i_id_ef}
\end{eqnarray}
and is conceptually equivalent to rotating finite-dimensional vectors via principal component analysis. The covariance matrix from Equation~\eqref{eq:suuij_id} is computed as follows: 
\begin{eqnarray}
\Kuu[m, n] &=& \int \int k(X, X^\prime) v_m(X) v_n(X^\prime) \; \mathrm{d} X^\prime \; \mathrm{d}X \nonumber \\
&=& \int v_m(X) \int k(X, X^\prime) v_n(X^\prime) \; \mathrm{d} X^\prime \; \mathrm{d}X \nonumber \\
&=& \int v_m(X) \lambda_n v_n(X) \; \mathrm{d}X = \lambda_n \int v_m(X)  v_n(X) \; \mathrm{d}X \nonumber \\
&=& \lambda_n \textrm{ if } m==n \textrm{ else } 0,
\label{eq:suuij_id_ef}
\end{eqnarray}
which is a diagonal matrix~\citep{Burt2019}. The step from the second to the third line merely utilizes the eigenfunction definition. The step from the third to the fourth line leverages that eigenfunctions are orthonormal, which means $\int v_m(X) v_n(X) \mathrm{d}X$ equals one if $m$ equals $n$ and is zero otherwise.
This is of great practical importance since a diagonal $\Kuu$ is more memory-efficient and invertible in linear rather than cubic time. Identifying eigenfunctions and eigenvalues for arbitrary kernels in closed-form is non-trivial, but solutions do exist for some~\citep{Rasmussen2006,Borovitskiy2020,Burt2020,Dutordoir2020,Riutort2020}.

\paragraph{Derivative Features.}
In addition to the linear integral operator formulation from above in Equations~\eqref{eq:muui_id} to~\eqref{eq:suuij_id} that contains Dirac, Fourier and kernel-eigen features as special cases, there is another possibility to define interdomain variables via the derivatives of a GP's function values~\citep{Adam2020,vanderWilk2020}. In this case, the inducing variable $\vu$ is expressed as $\vu[m] = \frac{\partial}{\partial X_{d(m)}} f(X)\Big|_{X=Z_m}$ where $Z_m$ refers to an inducing point and $d(m)$ to the input dimension of $X$ over which the partial derivative is performed (as determined by the interdomain variable $\vu[m]$ with index $m$). Every interdomain variable $\vu[m]$ needs to specify a particular input dimension $d(m)$ of $X$ the derivative of $f(\cdot)$ is taken with respect to. Equations~\eqref{eq:muui_id} to~\eqref{eq:suuij_id} then become:
\begin{eqnarray}
\muu[m] &=&  \frac{\partial}{\partial X_{d(m)}} \mu(X)\Bigg|_{X=Z_m} , \label{eq:muui_id_d}\\
\kfu[n] &=& \frac{\partial}{\partial X^\prime_{d(n)}} k(\cdot, X^\prime) \Bigg|_{X^\prime=Z_n} , \label{eq:sfu:i_id_d}\\
\Kuu[m, n] &=& \frac{\partial^2}{\partial X_{d(m)} \partial X^\prime_{d(n)}}  k(X, X^\prime) \Bigg|_{X=Z_m, X^\prime = Z_n}, \label{eq:suuij_id_d}
\end{eqnarray}
presupposing differentiable mean and kernel functions. This can be useful, e.g.\ when $\mathcal{X}$ is a time domain and $f(\cdot)$ represents a space domain, but one seeks to express interdomain variables in a velocity domain~\citep{OHagan1992,Rasmussen2006}. Mathematically, the results above are not surprising since differentiation is a linear operator over function spaces.


\subsection{Multioutput Gaussian Processes}
\label{sec:mo_gps}

A multioutput GP extends a GP to a distribution over vector-valued functions $\vf(\cdot): \mathcal{X} \rightarrow \mathbb{R}^D$, where $D$ refers to the number of outputs~\citep{Alvarez2012}. While an ordinary GP outputs a scalar Gaussian random variable when evaluated at a specific input location $X \in \mathcal{X}$, a multioutput GP outputs a $D$-dimensional multivariate Gaussian random variable when evaluated at a specific input location $X$. Formally, a multioutput GP is defined similarly to an ordinary singleoutput GP as in Equation~\eqref{eq:def_gp}, but with a vector-valued mean function $\bm{\mu}(\cdot): \mathcal{X} \rightarrow \mathbb{R}^D$ and a matrix-valued cross-covariance function $\textbf{K}(\cdot, \cdot^\prime): \mathcal{X} \times \mathcal{X} \rightarrow \mathbb{R}^{D \times D}$~\citep{Micchelli2005}. Note that $\textbf{K}(\cdot, \cdot^\prime)$ needs to output a proper cross-covariance matrix for every $(X,X^\prime)$-pair and must hence satisfy the following symmetry relation $\textbf{K}(\cdot, \cdot^\prime) = \textbf{K}(\cdot^\prime, \cdot)^\top$, i.e.\ $\textbf{K}(\cdot, \cdot^\prime)[i, j] = \textbf{K}(\cdot^\prime, \cdot)[j, i]$ in \texttt{numpy} indexing notation.
We would like to remind the reader at this stage to not confuse multioutput notation with singleoutput notation from earlier for random vectors $\vf$, mean vectors $\muf$ and covariance matrices $\Kff$ that result from evaluating a singleoutput GP in multiple locations $\{X_1, X_2, ..., X_N\}$.

One might ask how to evaluate a multioutput GP since this would naively lead to a random variable of extent $N \times D$ where $N$ refers to the number of evaluation points. The answer is that we can flatten the $N \times D$ random variable into a vector of size $N  D$. This results in a multivariate Gaussian random variable $\vf$ as described in Equation~\eqref{eq:gp_eval} with a mean vector $\muf$ of size $N D$ and a covariance matrix $\Kff$ of size $N D \times N D$. While the choice of flattening is up to the user, one could e.g.\ concatenate all $D$-dimensional random variables for each evaluation point. This yields a partitioned mean vector with $N$ partitions of size $D$ and a block-covariance matrix with $N \times N$ blocks each holding a $(D \times D)$-matrix. The block covariance matrix is required to be a proper covariance matrix, i.e.\ it needs to be symmetric (which is guaranteed by the imposed symmetry relation from earlier).

The ``flattening trick'' already hints at the fact that a multioutput GP can be indeed defined differently as a distribution over real-valued functions with a real-valued mean function and a real-valued covariance function. This is made possible through the ``output-as-input'' view where a multioutput GP's input domain is extended through an index set $\mathcal{I} = \{1, 2, ..., D \}$ to index output dimensions~\citep{vanderWilk2020}. More formally, this yields $f(\cdot): (\mathcal{X}, \mathcal{I}) \rightarrow \mathbb{R}$ distributed according to the mean function $\mu(\cdot): (\mathcal{X}, \mathcal{I}) \rightarrow \mathbb{R}$ and the covariance function $k(\cdot, \cdot^\prime): (\mathcal{X}, \mathcal{I}) \times (\mathcal{X}, \mathcal{I}) \rightarrow \mathbb{R}$, where the dot notation $\cdot$ now refers to a pair (the first element of the pair being the input $X \in \mathcal{X}$ and the second the output index $i \in \mathcal{I}$). In this notation, Equation~\eqref{eq:def_gp} for singleoutput GPs remains applicable (under the extended input domain).

The advantage of defining a multioutput GP this way is that it makes the evaluation at arbitrary input-output-index pairs $(X,i)$ more convenient, i.e.\ it is not required to evaluate the multioutput GP at all outputs $i$ for a specific input $X$. The latter comes in handy e.g.\ in (approximate) Bayesian inference problems that have multidimensional targets and where training examples can have missing labels (i.e.\ the set of labels for some training inputs $X$ can be incomplete).
The output-as-input view not only ensures that Equation~\eqref{eq:def_gp} remains valid for multioutput GPs but also ensures that Equation~\eqref{eq:sparse_mgp} remains valid for sparse multioutput GPs under an output-index-extended input domain. Naively, one could specify $M D$ inducing points $Z$ (or inducing features $\phi(\cdot)$ in the interdomain formulation) and $M D$ inducing variables $\vu$---$M$ for each output head respectively. In Equation~\eqref{eq:sparse_mgp}, this would lead $\muu$ and $\qu$ to be $M D$-dimensional vectors, $\Kuu$ and $\Quu$ to be $(M D \times M D)$-matrices, and $\kfu$ and $\kuf$ to be $M D$-dimensional vector-valued functions.

At this point, we have acquired an understanding of sparse multioutput GPs, but it can be insightful to continue with how to design computationally efficient sparse multioutput GPs in practice. The biggest computational chunk in Equation~\eqref{eq:sparse_mgp} is the inversion of $\Kuu$ which poses a cubic operation in $M D$---i.e.\ $\mathcal{O}(M^3D^3)$---under the naive specification from earlier. This can be addressed two-fold: by a separate independent multioutput GP that sacrifices the ability to model output correlations for the sake of computational efficiency~\citep{vanderWilk2020}, or by squashing a latent separate independent multioutput GP through a linear transformation to couple outputs---the latter can be achieved with a linear model of coregionalization~\citep{Journel1978} or convolutional GPs~\citep{Alvarez2010,vanderWilk2017}.

In the following, we are going to provide some specific examples of multioutput GPs and elucidate briefly how they can be leveraged to construct efficient sparse multioutput GPs. We refer the interested reader once more to \cite{vanderWilk2020} for a more in-depth discussion. We start with a separate independent multioutput GP, which is essentially the result of defining multiple separate ordinary singleoutput GPs over the same input domain, one for each output head---this gives rise to a multivariate Gaussian random variable with a diagonal covariance matrix at each input location. A simple way to obtain non-diagonal covariance matrices is via a linear model of coregionalization, that is the result of squashing the multivariate random variable of a separate independent multioutput GP through a fixed linear transformation at each input location. Subsequently, we will also cover convolutional and image-convolutional GPs that are not as related as the name indicates, and conclude the section with derivative GPs which are multioutput GPs that naturally arise when considering the derivative of a singleoutput GP's random function.

\paragraph{Separate Independent Multioutput GP.}
A separate independent multioutput GP specifies $D$ separate singleoutput GPs---one for each output dimension---which makes different output dimensions have zero-covariance, i.e.\ $\textbf{K}(\cdot, \cdot^\prime)$ is a diagonal matrix-valued function under the output-as-output view from earlier. Under the $M D$-flatting outlined previously, a separate independent multioutput sparse GP makes $\Kuu$ be a block matrix with $M \times M$ blocks each of which contains a diagonal matrix of size $D \times D$. However, we could have alternatively chosen a $D M$-flattening view rather than an $M D$-flattening view, in which case $\Kuu$ would be a block-diagonal matrix with $D \times D$ blocks, where each diagonal block is a full $(M \times M)$-matrix but each off-diagonal block contains only zeros. The latter rearrangement enables to invert $\Kuu$ by inverting the $D$ diagonal blocks (each of size $M \times M$) separately, yielding an improved computational complexity of $\mathcal{O}(M^3 D)$.


\paragraph{Linear Model of Coregionalization.}
A linear model of coregionalization~\citep{Journel1978} provides a simple approach to construct a sparse multioutput GP that is both computationally efficient and ensures correlated output heads. Resorting back to the output-as-output view, the idea is to squash a latent separate independent multioutput GP with a diagonal matrix-valued kernel $\textbf{K}_g(\cdot, \cdot^\prime)$ through a linear transformation~\citep{Dutordoir2018}. The linear transformation $\textbf{W}$ is defined to be a $D \times D_g$ matrix where $D_g$ denotes the number of latent outputs. Since $D_g$ is user-defined, it enables control over computational complexity because it determines the computational burden of inverting latent covariance matrices. This allows for efficient fully-correlated multioutput GPs with high-dimensional outputs where $d \gg d_g$. Denoting a generic latent GP in the output-as-output view as:
\begin{equation}
\label{eq:def_latent_mogp}
\textbf{g}(\cdot) \ \sim \ \mathcal{GP} \Big(  \bm{\mu}_g(\cdot), \textbf{K}_g(\cdot, \cdot^\prime) \Big) ,
\end{equation}
where $\textbf{g}(\cdot)$ refers to the latent vector-valued random function and $\bm{\mu}_g(\cdot)$ to the latent vector-valued mean function, a proper multioutput GP can be obtained via $\textbf{f}(\cdot) = \textbf{W} \textbf{g}(\cdot)$, resulting in:
\begin{equation}
\label{eq:def_loc_mogp}
\textbf{f}(\cdot) \ \sim \ \mathcal{GP} \Big( \textbf{W} \bm{\mu}_g(\cdot), \textbf{W} \textbf{K}_g(\cdot, \cdot^\prime) \textbf{W}^\top \Big) .
\end{equation}
Rather than defining the latent multioutput GP with Equation~\eqref{eq:def_latent_mogp} that is generic, we could have alternatively used the corresponding output-as-output view of a sparse multioutput GP according to Equation~\eqref{eq:sparse_mgp} in combination with the kernel function $\textbf{K}_g(\cdot, \cdot^\prime)$, in which case Equation~\eqref{eq:def_loc_mogp} would have represented a fully correlated but sparse multioutput GP with efficient matrix inversion in latent space. A more detailed discussion on the topic can be found in~\cite{vanderWilk2020}.

\paragraph{Convolutional GP.} In similar vein to a linear model of coregionalization, we can construct coupled output heads from a latent separate independent multioutput GP as given in Equation~\eqref{eq:def_latent_mogp} via a convolutional GP. The idea is to define $\textbf{f}(\cdot) = \int \textbf{G}(\cdot - X) \textbf{g}(X) \textrm{d} X$ where $\textbf{G}(\cdot)$ is a matrix-valued function that outputs a $D \times D_g$ matrix for every input $X$. The corresponding process over $\textbf{f}(\cdot)$ is a proper multioutput GP because of the linearity of the convolution operator, and is given by:
\begin{equation}
\label{eq:def_conv_gp}
\textbf{f}(\cdot) \ \sim \ \mathcal{GP} \Big( \int \textbf{G}(\cdot - X) \bm{\mu}_g(X) \; \textrm{d} X, \int \int \textbf{G}(\cdot - X) \textbf{K}_g(X, X^\prime) \textbf{G}^\top(\cdot^\prime - X^\prime) \; \textrm{d} X^\prime \; \textrm{d} X \Big) ,
\end{equation}
where the matrix-valued function $\textbf{G}(\cdot)$ is usually chosen such as to yield tractable integrals. The presentation of convolutional GPs here follows the descriptions in~\cite{Alvarez2010} and~\cite{vanderWilk2020}, but similar models have been proposed in earlier work~\citep{Higdon2002,Boyle2004,Alvarez2008,Alvarez2009}. 
Note that Equation~\eqref{eq:def_conv_gp} builds upon a generic latent GP according to Equation~\eqref{eq:def_latent_mogp} for conceptual convenience, but we could have used a sparse GP according to Equation~\eqref{eq:sparse_mgp} instead.

\paragraph{Image-Convolutional GP.} Despite the similarity in naming, an image-convolutional GP is different from a convolutional GP. Let's imagine a domain of images and that a single image $X$ is subdivided into a set of (possibly overlapping) patches, all of equal size and indexed by $p$. For notational convenience, we define the $p$-th patch of $X$ as $X[p]$. We can then define an ordinary singleoutput GP operating in a latent patch space with random functions denoted as $g(\cdot[p])$ and with mean function $\mu_g(\cdot[p])$ and kernel $k_g(\cdot[p], \cdot^\prime[p^\prime])$, where the notation $\cdot[p]$ refers to the $p$-th patch of an input image and where the input image is indicated by the usual dot notation~$\cdot$. 
It turns out that this latent singleoutput GP defined in patch space induces a multioutput GP over vector-valued functions $\textbf{f}(\cdot)$ that operates in image space (where the number of outputs equals the number of patches). The vector-valued random function $\textbf{f}(\cdot)$ then relates to the latent real-valued function $g(\cdot[p])$ as $\textbf{f}(\cdot)[p] = g(\cdot[p])$. Similarly, the multioutput mean function can be described as $\bm{\mu}(\cdot)[p] = \mu_g(\cdot[p])$ and the multioutput kernel as $\textbf{K}(\cdot, \cdot^\prime)[p, p^\prime] = k_g(\cdot[p], \cdot^\prime[p^\prime])$. This design was inspired by convolutional neural networks and a first description can be found in~\cite{vanderWilk2017} which was later extended to deep architectures by~\cite{Blomqvist2019} and~\cite{Dutordoir2020b}. The difference between a convolutional and an image-convolutional GP is hence that the former performs a convolution operation on the input domain, whereas the latter performs an operation that resembles a discrete two-dimensional convolution on a single element from an image input domain.

\paragraph{Derivative GP.}
We conclude this section with derivative GPs because they provide a natural example of multioutput GPs. Earlier, we have seen how to define interdomain variables that are partial derivatives of a singleoutput GP's random function (evaluated at specific locations). It turns out that for any singleoutput GP with mean function $\mu(\cdot)$ and kernel $k(\cdot, \cdot^\prime)$, the derivative of $f(\cdot)$ w.r.t.\ $X$ gives rise to a proper multioutput GP where the number of output dimensions $D$ equals the dimension of $\mathcal{X}$---presupposing that $\mu(\cdot)$ and $k(\cdot, \cdot^\prime)$ are differentiable. More precisely, in the output-as-output view, we obtain $\textbf{f}(\cdot) = \nabla_X f(X) \big|_{X = \cdot}$ with the corresponding multioutput mean function $\bm{\mu}(\cdot) = \nabla_X \mu(X) \big|_{X=\cdot}$ and the multioutput kernel function $\textbf{K}(\cdot, \cdot^\prime) = \nabla_X \nabla_{X^\prime} k(X, X^\prime) \big|_{X=\cdot, X^\prime = \cdot^\prime}$. Analysing the derivative of a multioutput GP, the resulting random function and its mean function are matrix-valued---in Jacobian notation $\textbf{J}_{\textbf{f}}(\cdot)$ and $\textbf{J}_{\bm{\mu}}(\cdot)$---, and the kernel function is a hypercubical tensor of dimension four (we refrain from a mathematical notation for preserving a clear view). However, by applying the flattening trick from earlier, we can again obtain a multioutput GP in the output-as-output view---e.g.\ by flattening the matrix-valued random function $\textbf{J}_{\textbf{f}}(\cdot)$ and its matrix-valued mean function $\textbf{J}_{\bm{\mu}}(\cdot)$ into vector-valued functions, which induces a flattening of the four-dimensional hypercubical tensor-valued kernel function into a matrix-valued kernel function.

\subsection{Deep Gaussian Processes}
\label{sec:deep_gps}

\begin{figure}[h!]
\centering
\includegraphics[trim=150 200 280 100,clip,width=0.5\textwidth]{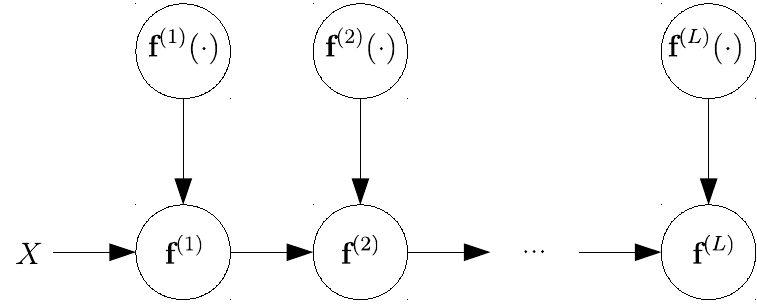}
\caption{Graphical model of a deep GP. A single input $X$ determines where to evaluate the first GP over vector-valued functions $\textbf{f}^{(1)}(\cdot)$ yielding a multivariate random variable $\textbf{f}^{(1)}$. A single sample from $\textbf{f}^{(1)}$ determines where to evaluate the next GP over $\textbf{f}^{(2)}(\cdot)$ resulting in $\textbf{f}^{(2)}$. Repeating this process until the last GP with index $L$ yields $\textbf{f}^{(L)}$ from which a random output for $X$ can be sampled.}
\label{fig:graphical_model_deep_gp}
\end{figure}

A deep GP is obtained by stacking multioutput GPs on top of each other~\citep{Damianou2013}. The output of one GP determines where the next GP is evaluated, i.e.\ the output dimension (=~number of outputs) of the GP below needs to conform with the input dimension of the GP on top. More formally, imagine $L$ multioutput GPs with random vector-valued functions denoted in the output-as-output notation as $\{ \textbf{f}^{(1)}(\cdot), \textbf{f}^{(2)}(\cdot), ..., \textbf{f}^{(L)}(\cdot) \}$. A single input $X$ is propagated through the deep GP as follows. The first GP with index $1$ is evaluated at $X$ yielding a vector-valued random variable $\textbf{f}^{(1)}$. Drawing a single sample from $\textbf{f}^{(1)}$ determines where to evaluate the second GP with index $2$. This yields another vector-valued random variable $\textbf{f}^{(2)}$ that determines where to evaluate the third GP, and so forth. This process is repeated until we arrive at the last GP with index $L$ yielding the random variable $\textbf{f}^{(L)}$. A single sample from $\textbf{f}^{(L)}$ results in one output sample of the deep GP for the input $X$. The graphical model illustrating this process is depicted in Figure~\ref{fig:graphical_model_deep_gp}. 

Note however that while the random variables $\{ \textbf{f}^{(1)}, \textbf{f}^{(2)}, ..., \textbf{f}^{(L)} \}$ are all multivariate normal, the marginal distribution over $\textbf{f}^{(L)}$ when integrating out $\{ \textbf{f}^{(1)}, \textbf{f}^{(2)}, ..., \textbf{f}^{(L-1)} \}$ for one given input $X$ is no longer multivariate normal and can assume multiple modes. The same is true for the marginal distributions of $\{ \textbf{f}^{(2)}, ..., \textbf{f}^{(L-1)} \}$ for a given $X$. Only $\textbf{f}^{(1)}$ is marginally multivariate normal for a given $X$ because it sits at the beginning of the hierarchy.

Note also that for deep GPs, the notation $\textbf{f}$ refers to a vector-valued random variable as a result of evaluating all output heads of a multioutput GP at a single input location $X$. This clashes with the same notation used for shallow singleoutput GPs where $\textbf{f}$ refers to a vector-valued random variable as a result of evaluating a singleoutput GP in multiple locations $\{X_1, X_2, ..., X_N\}$. We apologize for the confusion but there are only so many ways to represent vector-valued quantities. 
Also note that if we propagate $N$ samples $\{X_1, X_2, ..., X_N\}$ through a deep GP (instead of just one as depicted above), we would need to evaluate the first GP at all $N$ inputs yielding a random variable of size $N \times D^{(1)}$ in the output-as-output view (where $D^{(1)}$ refers to the number of outputs of the first GP). Sampling precisely once from this $N \times D^{(1)}$ random variable results in $N$ vector-valued samples of size $D^{(1)}$ each, that would be used to evaluate the second GP, and so on. The final output would be a random variable of size $N \times D^{(L)}$, where $D^{(L)}$ refers to the number of outputs of the last GP.

\begin{figure}[h!]
\centering
\includegraphics[trim=0 0 0 0,clip,width=\textwidth]{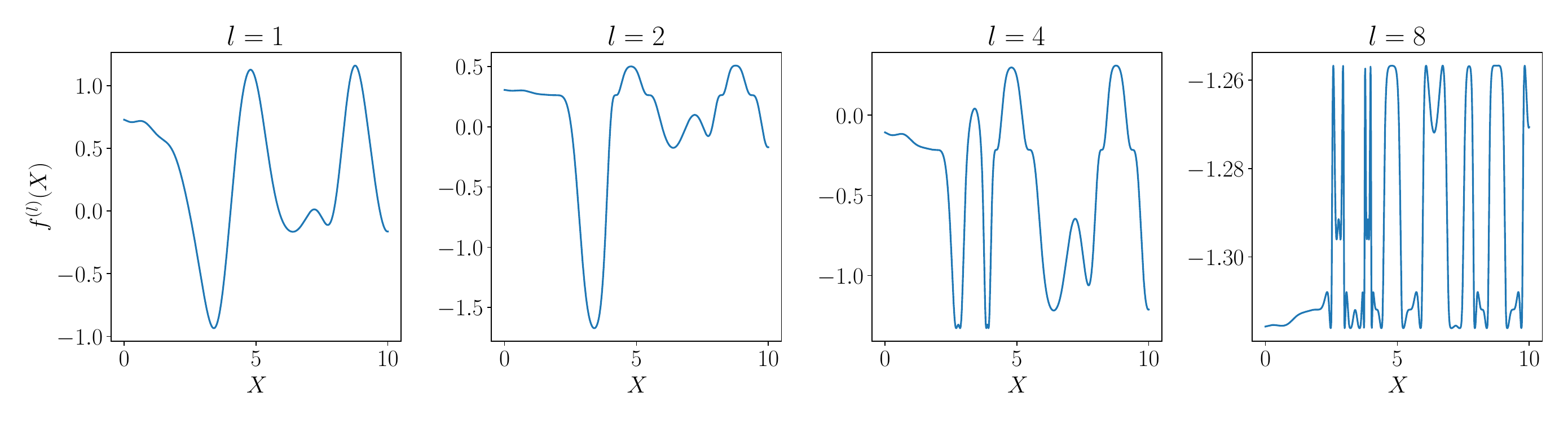}
\caption{Samples from a deep GP at various depths. The deep GP building block for this figure is a singleoutput GP operating on a one-dimensional input domain and using an RBF kernel (lengthscale 0.7) with a zero-mean function. An equally spaced set of evaluation points $X$ from the interval $[0, 10]$ is propagated through the deep GP and the function values $f^{(l)}(X)$ are shown at increasing depths ($l \in \{1, 2, 4, 8\}$). With depth increasing, function values rapidly change and assume values from a narrow range. The reason is that once samples are mapped to similar values in intermediate layers, they won't assume significantly different values in subsequent layers---see~\cite{Duvenaud2014}.}
\label{fig:deep_rbf}
\end{figure}

The behavior of a deep GP is illustrated in Figure~\ref{fig:deep_rbf} for singleoutput GP building blocks that have one-dimensional input domains (for reasons of interpretability). With increasing depth, function values become more narrow-ranged and the function changes from smooth to becoming more abrupt ~\citep{Duvenaud2014}. The latter is not surprising since once samples at intermediate layers are mapped to similar function values, they won't assume very different function values in subsequent layers. While this enables to potentially model challenging functions that are less smooth (which may be difficult with an ordinary shallow GP), the marginal distributions over function values in every layer (except for the first one) are no longer Gaussian (as alluded to earlier) and hence impede analytical uncertainty estimates, which is usually considered a hallmark of GP models.

We conclude this section by noting that the definition of a deep GP is independent of the type of shallow GP used as a building block. Modern deep GPs stack sparse GPs, as presented in Equation~\eqref{eq:sparse_mgp}, on top of each other to be computationally efficient~\citep{Salimbeni2017,Salimbeni2019}. Up to this point, we haven't yet addressed how to do inference in (deep) sparse GPs. The reason is that exact inference is not possible and one has to resort to approximate inference techniques. Variational inference (VI) is a convenient tool used by contemporary literature in this context. We therefore dedicate the next section (Section~\ref{sec:vi}) to explain general VI, before coming back to VI in shallow and deep sparse GPs later on in Section~\ref{sec:vi_with_svgps}.

\section{Variational Inference}
\label{sec:vi}

VI is a specific type of approximate Bayesian inference. As the name indicates, approximate Bayesian inference deals with approximating posterior distributions that are computationally intractable. The goal of this section is to provide an overview of VI, where we resort mostly to the parameter space view and where we assume a supervised learning scenario. How to combine sparse GPs (that are function space models) with VI is subject of Section~\ref{sec:vi_with_svgps} later on. We begin with Section~\ref{sec:vanilla_vi}, where we derive vanilla VI. In Section~\ref{sec:iw_vi}, we demonstrate an alternative way to derive VI with importance weighting that provides a more accurate solution at the expense of increased computational complexity. After that, in Section~\ref{sec:lv_vi}, we introduce latent-variable VI to enable more flexible models. In Section~\ref{sec:iwlv_vi}, we combine latent-variable VI with importance weighting to trade computational cost for accuracy. Finally, in Section~\ref{sec:bayesian_layers}, we present how to do VI in a hierarchical and compositional fashion giving rise to a generic framework for Bayesian deep learning via the concept of Bayesian layers~\citep{Tran2019}.

\begin{figure}[h!]
\centering
\includegraphics[trim=0 200 50 0,clip,width=\textwidth]{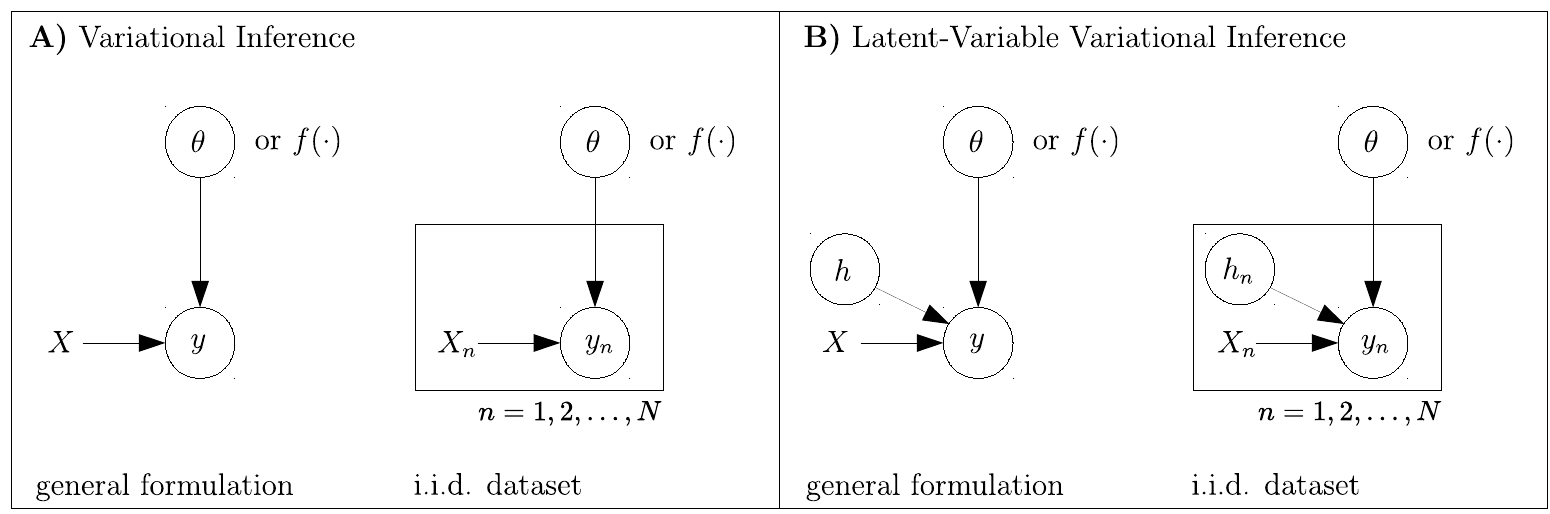}
\caption{Graphical models for VI in supervised learning settings. $X$ refers to inputs and $y$ to labels. Unknown functions are depicted via the variable $\theta$ (in parameter space view) or $f(\cdot)$ (in function space view). Vanilla VI is illustrated in \textbf{A)} and subject of Section~\ref{sec:vanilla_vi}, whereas latent-variable VI is illustrated in \textbf{B)} and subject of Section~\ref{sec:lv_vi}. Latent-variable VI introduces another latent variable $h$ (in addition to $\theta$) to enable more expressive generative models. For the sake of clarity, we present in both cases one general formulation and another one in which the dataset is i.i.d.\ which is a common assumption in most scenarios---individual training examples are then indexed with $n$.}
\label{fig:vi}
\end{figure}

\subsection{Vanilla Variational Inference}
\label{sec:vanilla_vi}

Let's start be revisiting Bayesian inference for supervised learning. Imagine some input $X$, some observed variable $y$ and the probability of observing $y$ given $X$ via a parametric distribution $p_\gamma(y | \theta, X)$ with hyperparameters $\gamma$ and unknown parameters $\theta$. Our goal is to infer $\theta$ and we have some prior belief over $\theta$ through the distribution $p_\gamma(\theta)$, where we assume for notational convenience that both $p_\gamma(y | \theta, X)$ and $p_\gamma(\theta)$ are hyperparameterized by $\gamma$. Inference over $\theta$ is then obtained via the posterior distribution over $\theta$ after observing $y$ and $X$:
\begin{equation}
\label{eq:bayes_rule}
p_\gamma(\theta|y, X) = \frac{p_\gamma(y | \theta, X) p_\gamma(\theta)}{\int p_\gamma(y | \theta, X) p_\gamma(\theta) \mathrm{d}\theta},
\end{equation}
where $p_\gamma(y | \theta, X)$ is referred to as likelihood, $p_\gamma(\theta)$ as prior and $p_\gamma(y|X) = \int p_\gamma(y | \theta, X) p_\gamma(\theta) \mathrm{d}\theta$ as marginal likelihood (or evidence). The corresponding graphical model for this inference problem is depicted in Figure~\ref{fig:vi} A) on the left-hand side (denoted as ``general formulation''). The graphical model represents the joint distribution $p_\gamma(y | \theta, X) p_\gamma(\theta)$ between $y$ and $\theta$ given $X$, which is also referred to as ``generative model''. The challenge in computing the posterior is that the marginal likelihood usually does not have a closed form solution except for special cases, e.g.\ when the prior is conjugate to the likelihood (which we won't consider in this tutorial). When the marginal likelihood does have a closed form solution, it is usually maximized w.r.t.\ to the hyperparemeters~$\gamma$ of the generative model before the exact posterior is computed~\citep{Bishop2006,Rasmussen2006}. Note that the hyperparameters $\gamma$ are also referred to as ``generative parameters''.

The idea in VI is to introduce an approximation $q_\psi(\theta)$, parameterized via $\psi$, to the intractable posterior $p_\gamma(\theta|y, X)$, and to optimize for $\psi$ such that the approximate posterior becomes close to the true posterior. In this regard, the approximate posterior $q_\psi(\theta)$ is also referred to as ``variational model'' and $\psi$ as ``variational parameters''. The question is which optimization objective to choose to identify optimal variational parameters $\psi$. We are going to respond to this question shortly but for now, we commence with the negative Kullback-Leibler divergence ($\KL$) between the approximate and the true posterior, which can be written as (by applying Bayes' rule to the true posterior):
\begin{equation}
\label{eq:neg_kl}
-\KL\Big(q_\psi(\theta) \Big|\Big| p_\gamma(\theta|y, X) \Big) = \int q_\psi(\theta) \ln p_\gamma(y | \theta, X) \; \mathrm{d}\theta -\KL\Big(q_\psi(\theta) \Big|\Big| p_\gamma(\theta) \Big) - \ln p_\gamma(y|X) .
\end{equation}
Rearranging by bringing the log marginal likelihood term $\ln p_\gamma(y|X)$ to the left yields:
\begin{equation}
\label{eq:elbo}
\ln p_\gamma(y|X) - \KL\Big(q_\psi(\theta) \Big|\Big| p_\gamma(\theta|y, X) \Big) = \underbrace{\int q_\psi(\theta) \ln p_\gamma(y | \theta, X) \; \mathrm{d}\theta -\KL\Big(q_\psi(\theta) \Big|\Big| p_\gamma(\theta) \Big)}_{=:\ELBO(\gamma,\psi)}.
\end{equation}
The term on the right-hand side is referred to as the evidence lower bound $\ELBO(\gamma,\psi)$~\citep{Rasmussen2006} since it poses a lower bound to the log marginal likelihood (a.k.a.\ log evidence)---``log evidence lower bound'' might hence be a more suitable description but omitting ``log'' is established convention. The $\ELBO$ is a lower bound because the $\KL$ between the posterior approximation and the true posterior is non-negative. Since the log marginal likelihood does not depend on the variational parameters $\psi$, the $\ELBO$ assumes its maximum when the approximate posterior equals the true one, i.e.\ $q_\psi(\theta) = p_\gamma(\theta|y, X)$, in which case the $\KL$ term on the left is zero and the $\ELBO$ recovers the log marginal likelihood exactly. 

Note how the formulation for the $\ELBO$ does not require to know the true posterior in its functional form a priori in order to identify an optimal approximation, because Equation~\eqref{eq:elbo} was obtained via decomposing the intractable posterior via Bayes' rule. Also note that the log marginal likelihood is usually the preferred objective to maximize for the generative hyperparameters $\gamma$, as mentioned earlier. Contemporary VI methods therefore maximize the evidence lower bound $\max_{\gamma,\psi}\ELBO(\gamma,\psi)$ jointly w.r.t.\ both generative parameters $\gamma$ and variational parameters $\psi$. Some current methods with deep function approximators choose a slight modification of Equation~\eqref{eq:elbo} by multiplying the $\KL$ term between the approximate posterior and the prior with a positive $\beta$-parameter. This is called ``$\beta$-VI'' and recovers a maximum expected log likelihood objective as a special case when $\beta \rightarrow 0$. It has been proposed by~\cite{Higgins2017}, and \cite{Wenzel2020} provide a recent discussion.

Assuming an optimal approximate posterior has been identified after optimizing the $\ELBO$ w.r.t.\ both variational parameters $\psi$ and generative parameters $\gamma$, the next question is how to use it, namely how to predict $y^\star$ for a new data point $X^\star$ that is not part of the training data. The answer is:
\begin{equation}
\label{eq:pred}
p(y^\star | X^\star) =  \int p_\gamma(y^\star | \theta, X^\star) q_\psi(\theta) \; \mathrm{d}\theta ,
\end{equation}
by forming the joint between the likelihood $p_\gamma(y^\star | \theta, X^\star)$ and the approximate posterior $q_\psi(\theta)$, and integrating out $\theta$. If the integration has no closed form, one has to resort to Monte Carlo methods---i.e.\ replace the integral over $\theta$ with an empirical average via samples obtained from $q_\psi(\theta)$.

So far, we haven't made any assumptions on how the generative model looks like precisely. In supervised learning, it is however common to assume an i.i.d.\ dataset in the sense that the training set comprises $N$ i.i.d.\ training examples in the form of ($X_n, y_n$)-pairs. The corresponding graphical model is depicted in Figure~\ref{fig:vi} A) on the right denoted as ``i.i.d.\ dataset''. In this case, the likelihood is given by $\prod_{n=1}^N p_\gamma(y_n | \theta, X_n)$ and the $\ELBO$ becomes:
\begin{equation}
\label{eq:elbo_iid}
\ELBO(\gamma,\psi) = \sum_{n=1}^N \int q_\psi(\theta) \ln p_\gamma(y_n | \theta, X_n) \; \mathrm{d}\theta -\KL\Big(q_\psi(\theta) \Big|\Big| p_\gamma(\theta) \Big).
\end{equation}
An interesting fact to note is that in case of large $N$, Equation~\eqref{eq:elbo_iid} can be approximated by Monte Carlo using minibatches, which can be used for parameter updates without exceeding potential memory limits~\citep{Hensman2013}.
The corresponding predictions $\{y_n^\star\}_{n=1,..,N^\star}$ for new data points $\{X_n^\star\}_{n=1,..,N^\star}$ in the i.i.d.\ setting are:
\begin{equation}
\label{eq:pred_iid}
p(y_1^\star, ..., y_{N^\star}^\star | X_1^\star, ..., X_{N^\star}^\star) =  \int \prod_{n=1}^{N^\star} p_\gamma(y^\star_n | \theta, X^\star_n) q_\psi(\theta) \; \mathrm{d}\theta .
\end{equation}

Note that we have deliberately not made any assumptions on the dimensions of $y$, $X$, $\theta$, $\gamma$ and $\psi$ to keep the notation light (which doesn't mean that these quantities need to be scalars). We also chose the weight space view by using $\theta$ instead of the function space view, although both are conceptually equivalent. The function space view can be obtained by replacing $\theta$ with $f(\cdot)$ in all formulations and equations above. Practically, one would need to be careful with expectations and $\KL$ divergences between infinite-dimensional random variables. We are going to address this issue later in Section~\ref{sec:vi_with_svgps} when talking about VI in sparse GPs (that naturally assume the function space view).

Since the probability distributions have been held generic so far, it can be insightful to provide some examples. To this end, imagine an i.i.d.\ regression problem with one-dimensional labels. Assume the prior $p(\bm{\theta})$ is a mean field multivariate Gaussian over the vectorized weights $\bm{\theta}$ in a neural network, with mean vector $\bm{\mu}_{\bm{\theta}}$ and variance vector $\bm{\upsilon}_{\bm{\theta}}$. Let the likelihood $p_\gamma(y_n | \bm{\theta}, X_n)$ be a homoscedastic Gaussian with variance $\upsilon^{(\gamma)}_{\textrm{lik}}$, whose mean depends on the neural net's output---i.e.\ $\mu_{\textrm{lik}}(X_n) = f_{\bm{\theta}}(X_n)$ where $f_{\bm{\theta}}(X_n)$ denotes the output of the neural net for the input $X_n$. In this context, the superscript $(\gamma)$ marks the likelihood variance as a generative parameter. The variational approximation $q_\psi(\bm{\theta})$ could then be a mean field multivariate Gaussian as well, with mean vector $\textbf{m}_{\bm{\theta}}^{(\psi)}$ and variance vector $\textbf{v}_{\bm{\theta}}^{(\psi)}$, where the superscript $(\psi)$ marks variational parameters. We have just arrived at a vanilla Bayesian neural network as in~\cite{Blundell2015}. 

We can increase the expressiveness of the likelihood by making it heteroscedastic, i.e.\ by letting the neural net output a two-dimensional vector $\textbf{f}_{\bm{\theta}}(\cdot)$ instead of a scalar to encode both the mean and the variance. This is achieved by defining $\mu_{\textrm{lik}}(X_n) = \textbf{f}_{\bm{\theta}}(X_n)[1]$ and $\upsilon_{\textrm{lik}}(X_n) = g(\textbf{f}_{\bm{\theta}}(X_n)[2])$ where $1$ and $2$ index both network outputs and $g(\cdot)$ is a strictly positive function (because the neural net's output is unbounded in general). In the latter case, there wouldn't be any generative parameter $\gamma$ anymore because the likelihood variance has become a function of the neural net's output. 

In practice, during optimization with gradient methods, the reparameterization trick~\citep{Kingma2014, Rezende2014} is applied to the random variable $\bm{\theta}$ in order to establish a differentiable relationship between $\bm{\theta}$ and the parameters of the distribution $q_\psi(\bm{\theta})$ from which $\bm{\theta}$ is sampled, i.e.\ the mean and the variance parameters $\textbf{m}_{\bm{\theta}}^{(\psi)}$ and $\textbf{v}_{\bm{\theta}}^{(\psi)}$. This is known to produce parameter updates with lower variance leading to better optimization. If we replace the prior and the approximate posterior with (multioutput) GPs and replace the notation $f_{\bm{\theta}}(\cdot)$~and~$\textbf{f}_{\bm{\theta}}(\cdot)$ with $f(\cdot)$~and~$\textbf{f}(\cdot)$ accordingly in the likelihood, we would obtain the GP equivalents of the homoscedastic and heteroscedastic Bayesian neural networks respectively. However, in GPs, one usually treats certain kernel hyperparameters as generative parameters $\gamma$, which means the prior is subject to optimization as opposed to the Bayesian neural network case.

We conclude by mentioning a related approximate inference scheme called expectation propagation (EP)~\citep{Bishop2006,Bui2017} that also encourages an approximate posterior to be close to the true posterior via a $\KL$ objective, similar to VI\@. In fact, EP chooses a similar objective as in Equation~\eqref{eq:neg_kl} but with swapped arguments in the $\KL$. The practical difference is that VI tends to provide mode-centered solutions whereas EP tends to provide support-covering solutions at the price of potentially significant mode mismatch~\citep{Bishop2006} (if the approximation is unimodal but the true posterior multimodal). There are at least two more advantages of VI over vanilla EP\@. First, in VI, the expectation is w.r.t.\ to the approximate posterior and hence amenable to sampling and stochastic optimization with gradient methods, whereas the expectation in EP is w.r.t.\ to the unknown optimal posterior. And second, VI is principled in that it lower-bounds the log marginal likelihood therefore encouraging convenient optimization not only over variational but also generative parameters.

\subsection{Importance-Weighted Variational Inference}
\label{sec:iw_vi}

Importance-weighted VI provides a way of lower-bounding the log marginal likelihood more tightly and with less estimation variance at the expense of increased computational complexity. We start by showing that there is an alternative way to derive the $\ELBO$, in addition to the derivation from the previous section, according to the following formulation:
\begin{eqnarray}
\ln p_\gamma(y|X) &=& \ln \int p_\gamma(y | \theta, X) p_\gamma(\theta) \; \mathrm{d}\theta = \ln \int q_\psi(\theta) \frac{p_\gamma(y | \theta, X) p_\gamma(\theta)}{q_\psi(\theta)} \; \mathrm{d}\theta \label{eq:iw_relevant}\\
&\geq& \int q_\psi(\theta) \ln \frac{p_\gamma(y | \theta, X) p_\gamma(\theta)}{q_\psi(\theta)} \; \mathrm{d}\theta = \ELBO(\gamma,\psi), \label{eq:elbo_recovered}
\end{eqnarray}
where the inequality comes from applying Jensen's inequality that swaps the logarithm with the expectation over $q_\psi(\theta)$. While this derivation is straightforward, it has the disadvantage of only demonstrating that the $\ELBO$ lower-bounds the log marginal likelihood but not by how much, namely the $\KL$ between the approximate and the true posterior, as shown in Equation~\eqref{eq:elbo}.

In order to obtain an importance-weighted formulation that bounds the log marginal likelihood more tightly~\citep{Burda2016, Domke2018}, we need to proceed from Equation~\eqref{eq:iw_relevant} before applying Jensen:
\begin{eqnarray}
\ln p_\gamma(y|X) &=&  \ln \int q_\psi(\theta) \frac{p_\gamma(y | \theta, X) p_\gamma(\theta)}{q_\psi(\theta)} \; \mathrm{d}\theta =\ln \mathbb{E}_{q_\psi(\theta)} \Bigg[ \frac{p_\gamma(y | \theta, X) p_\gamma(\theta)}{q_\psi(\theta)} \Bigg] \label{eq:iw_vi_1}  \\
&=& \ln \frac{1}{S} \sum_{s=1}^S \mathbb{E}_{q_\psi(\theta^{(s)})} \Bigg[  \frac{p_\gamma(y | \theta^{(s)}, X) p_\gamma(\theta^{(s)})}{q_\psi(\theta^{(s)})} \Bigg] \label{eq:iw_vi_1.5} \\
&=& \ln \mathbb{E}_{\prod_{s=1}^S q_\psi(\theta^{(s)})} \Bigg[ \frac{1}{S} \sum_{s=1}^S  \frac{p_\gamma(y | \theta^{(s)}, X) p_\gamma(\theta^{(s)})}{q_\psi(\theta^{(s)})} \Bigg] \label{eq:iw_vi_2} \\
&\geq& \mathbb{E}_{\prod_{s=1}^S q_\psi(\theta^{(s)})} \Bigg[ \ln \frac{1}{S} \sum_{s=1}^S  \frac{p_\gamma(y | \theta^{(s)}, X) p_\gamma(\theta^{(s)})}{q_\psi(\theta^{(s)})} \Bigg] =: \ELBO_S(\gamma,\psi). \; \; \label{eq:iw_vi_3}
\end{eqnarray}
In Equation~\eqref{eq:iw_vi_1.5}, the expectation in Equation~\eqref{eq:iw_vi_1} is replicated $S$ times by introducing $S$ i.i.d.\ variables $\theta^{(s)}$ and computing the average over those. In Equation~\eqref{eq:iw_vi_2}, the expectation over $\theta^{(s)}$ is swapped with the sum before applying Jensen in Equation~\eqref{eq:iw_vi_3}. The final importance-weighted $\ELBO$ is denoted as $\ELBO_S(\gamma,\psi)$ with an explicit dependence on the number of replicates $S$ and where importance weights are given by the fraction between $p_\gamma(\theta^{(s)})$ and $q_\psi(\theta^{(s)})$.

It is straightforward to verify that the ordinary $\ELBO$ from Equation~\eqref{eq:elbo} in the previous section is recovered as a special case of the importance-weighted $\ELBO_S$ for $S=1$. It turns out that the other extreme, when $S \rightarrow \infty$, recovers the log marginal likelihood, demonstrated as follows:
\begin{eqnarray}
\lim_{S \rightarrow \infty} \ELBO_S(\gamma,\psi) &=& \lim_{S \rightarrow \infty} \mathbb{E}_{\prod_{s=1}^S q_\psi(\theta^{(s)})} \Bigg[ \ln \frac{1}{S} \sum_{s=1}^S  \frac{p_\gamma(y | \theta^{(s)}, X) p_\gamma(\theta^{(s)})}{q_\psi(\theta^{(s)})} \Bigg] \label{eq:ml_rec_1} \\
&=& \mathbb{E}_{\prod_{s=1}^S q_\psi(\theta^{(s)})} \Bigg[ \ln \int q_\psi(\theta) \frac{p_\gamma(y | \theta, X) p_\gamma(\theta)}{q_\psi(\theta)} \; \mathrm{d} \theta \Bigg] \label{eq:ml_rec_2} \\
&=& \ln \int \cancel{q_\psi(\theta)} \frac{p_\gamma(y | \theta, X) p_\gamma(\theta)}{\cancel{q_\psi(\theta)}} \; \mathrm{d} \theta = \ln p_\gamma(y|X). 
\end{eqnarray}
It can be furthermore shown that the following sequence of inequalities holds in accordance with~\cite{Burda2016} and~\cite{Domke2018}:
\begin{equation}
\label{eq:elbo_ineq}
\ELBO(\gamma,\psi) = \ELBO_1(\gamma,\psi) \leq \ELBO_2(\gamma,\psi) \leq ... \leq \lim_{S \rightarrow \infty} \ELBO_S(\gamma,\psi) = \ln p_\gamma(y|X),
\end{equation}
where the computational complexity is determined by the number of replicates $S$ and increases from left to right. In the limit of infinite computational resources, $\ln p_\gamma(y|X)$ is recovered exactly. Note that $\ELBO_S$ is not only a tighter bound for large $S$, but also empirical estimates of $\ELBO_S$ (via sampling the outer expectation from $\theta^{(1)}$ to $\theta^{(S)}$) become more accurate and have less variance as $S$ increases. This becomes apparent in the limit of $S \rightarrow \infty$ when every sample of the expectation over~$\theta^{(1)}$ up to $\theta^{(S)}$ yields the same result, which is the exact log marginal likelihood.

For the sake of completeness, we provide here the importance-weighted formulation in case of a dataset with i.i.d.\ training samples $\{(y_n, X_n)\}_{n=1,..,N}$ by adjusting the likelihood accordingly:
\begin{equation}
\label{eq:iw_vi_iid}
\ELBO_S(\gamma,\psi) = \mathbb{E}_{\prod_{s=1}^S q_\psi(\theta^{(s)})} \Bigg[ \ln \frac{1}{S} \sum_{s=1}^S  \frac{\prod_{n=1}^N p_\gamma(y_n | \theta^{(s)}, X_n) p_\gamma(\theta^{(s)})}{q_\psi(\theta^{(s)})} \Bigg],
\end{equation}
that can be approximated with samples $\theta^{(s)}$ from each of the replicated distributions $q_\psi(\theta^{(s)})$, just like the non-i.i.d.\ formulation.
Note that the way prediction is performed for new samples $X^\star$ is the same for importance-weighted VI as for vanilla VI, and Equations~\eqref{eq:pred} and~\eqref{eq:pred_iid} from the previous section apply (for both the general and the i.i.d.\ case respectively).

\subsection{Latent-Variable Variational Inference}
\label{sec:lv_vi}

The idea behind latent-variable VI is to introduce another latent variable $h$ in addition to $\theta$ as illustrated by the graphical model in Figure~\ref{fig:vi} B) on the left (``general formulation''). The reason for this is to construct generative models that are more flexible as discussed shortly. To that end, it is assumed that the prior over $\theta$ and $h$ factorizes into $p_\gamma(\theta)$ and $p_\gamma(h)$, and the likelihood $p_\gamma(y | \theta, h, X)$ is conditioned on $h$ in addition to $\theta$ and $X$. We again indicate with $\gamma$ the entirety of all generative parameters for notational convenience. Since $h$ is latent, we need to do joint inference over $\theta$ and $h$. Under the typical assumption of a factorized approximate posterior $q_\psi(\theta) q_\psi(h)$, where $\psi$ indicates all variational parameters, we arrive at the latent-variable $\ELBO$:
\begin{eqnarray}
\ELBO(\gamma, \psi) &=& \int \int q_\psi(\theta) q_\psi(h) \ln p_\gamma(y | \theta, h, X)  \; \mathrm{d} h \;  \mathrm{d}\theta \nonumber \\
&& -\KL\Big(q_\psi(h) \Big|\Big| p_\gamma(h) \Big) -\KL\Big(q_\psi(\theta) \Big|\Big| p_\gamma(\theta) \Big), \label{eq:elbo_lv}
\end{eqnarray}
where the two $\KL$ terms are a result of the assumed factorization between $\theta$ and $h$ in both the prior and the approximate posterior.
Predicting $y^\star$ for previously unseen $X^\star$ is then achieved via:
\begin{equation}
\label{eq:pred_lv}
p(y^\star | X^\star) =  \int \int p_\gamma(y^\star | \theta, h, X^\star) p_\gamma(h) \; \mathrm{d} h \; q_\psi(\theta) \; \mathrm{d}\theta ,
\end{equation}
where, importantly, the prior $p_\gamma(h)$ over the latent variable $h$ is used and not the approximate posterior $q_\psi(h)$ for reasons that become apparent shortly.
At this point, one might wonder why we have introduced the latent variable $h$ in the first place as it seems notationally redundant to $\theta$. We shed light into this by providing a more concrete example for the functional form of the likelihood. To that end, imagine the vanilla homoscedastic Bayesian neural net example from Section~\ref{sec:vanilla_vi} where boldface $\bm{\theta}$ represents the vectorized weights of a neural net with a mean field multivariate prior $p(\bm{\theta})$. Under the latent-variable formulation, we need to introduce another prior over $h$ and the homoscedastic Gaussian likelihood needs to be conditioned on $h$ as well. This is where the difference between $\bm{\theta}$ and $h$ becomes apparent: the mean is then defined as $\mu_{\textrm{lik}}(h, X) = f_{\bm{\theta}}(h, X)$ where $\bm{\theta}$ parameterizes the mean function as a neural net (indicated by the subscript $\bm{\theta}$) but $h$ serves as additional neural network input. Ordinarily, without an additional latent variable $h$, the distribution over $y$ for a given $\bm{\theta}$ and $X$ is a unimodal Gaussian. However, by introducing $h$ and adding it to the neural net input, the distribution over $y$ for a given $\bm{\theta}$ and $X$ becomes non-Gaussian (when integrated over $h$) and can assume multiple modes. A multimodal distribution is more expressive in the sense that it can model more challenging relationships between labels $y$ and the corresponding inputs $X$.

The latent-variable formulation is typically combined with the assumption of an i.i.d.\ training dataset $\{(y_n, X_n)\}_{n=1,..,N}$---see the graphical model in Figure~\ref{fig:vi} B) on the right (``i.i.d.\ dataset''). Since the latent variable $h$ is considered as additional likelihood input in addition to $X$, it also assumed i.i.d.\ across training examples and gets an index $n$. Under a factorized likelihood, the $\ELBO$ becomes:
\begin{eqnarray}
\ELBO(\gamma, \psi) &=& \sum_{n=1}^N \int \int q_\psi(\theta) q_\psi(h_n) \ln p_\gamma(y_n | \theta, h_n, X_n) \; \mathrm{d} h_n \;  \mathrm{d}\theta \nonumber \\
&& -\sum_{n=1}^N \KL\Big(q_\psi(h_n) \Big|\Big| p_\gamma(h_n) \Big) -\KL\Big(q_\psi(\theta) \Big|\Big| p_\gamma(\theta) \Big), \label{eq:elbo_lv_iid}
\end{eqnarray}
with a separate integral and $\KL$ term for each $h_n$. Predictions $\{y_n^\star\}_{i=1,..,N^\star}$ for new data points $\{X_n^\star\}_{n=1,..,N^\star}$ are then accomplished via the following formulation:
\begin{equation}
\label{eq:pred_lv_iid}
p(y_1^\star, ..., y_{N^\star}^\star | X_1^\star, ..., X_{N^\star}^\star) =  \int \prod_{n=1}^{N^\star}  \int p_\gamma(y_n^\star | \theta, h_n, X_n^\star) p_\gamma(h_n) \; \mathrm{d} h_n \; q_\psi(\theta) \; \mathrm{d}\theta ,
\end{equation}
where, importantly, $h_n$ is integrated out with the prior $p_\gamma(h_n)$ instead of with the approximate posterior. The reason for integrating $h_n$ with the prior is that, naively, in the course of training, there is a separate approximate posterior $q_\psi(h_n)$ for each individual training example $(X_n, y_n)$. The approximate posterior $q_\psi(h_n)$ can therefore be interpreted as auxiliary training tool that does not readily generalize to unseen $X^\star_n$, and is typically ``thrown away'' after the training phase because it is no longer needed for prediction.

For illuminating purposes, let's provide a more concrete example for an i.i.d.\ regression problem with one-dimensional labels $y_n$. Imagine, similarly to earlier, that the prior $p(\bm{\theta})$ is a mean field multivariate Gaussian over vectorized weights $\bm{\theta}$ of a neural net. Let's furthermore imagine that $p(\textbf{h}_n)$ is a multivariate standard normal Gaussian. Let the likelihood $p_\gamma(y_n|\bm{\theta},\textbf{h}_n,X_n)$ be a homoscedastic Gaussian with variance $\upsilon_{\textrm{lik}}^{(\gamma)}$---where $\gamma$ indicates that the variance is a generative parameter---and a neural net mean function $\mu_{\textrm{lik}}(\textbf{h}_n,X_n) =f_{\bm{\theta}}(\textbf{h}_n, X_n)$ that has as input both $\textbf{h}_n$ and $X_n$. Let the variational approximation $q_\psi(\bm{\theta})$ be a mean field multivariate Gaussian, and importantly, let's assume a multivariate Gaussian approximate posterior over $\textbf{h}_n$ for every data point $n$ denoted as $q_\psi(\textbf{h}_n)$---where the variational parameters are mean-covariance-pairs for each $n$. We could alternatively parameterize the approximate posterior over $\textbf{h}_n$ differently, e.g.\ as $q_\psi(\textbf{h}_n | y_n, X_n)$ via a neural net that maps a $(y_n, X_n)$-tuple to a mean vector and covariance matrix for $\textbf{h}_n$ (in which case the variational parameters were the weights of the neural net mapping, rather than an individual mean-covariance-pair for each training example $n$). The latter is called ``amortized VI'' and the variational neural net referred to as ``recognition model'' or ``encoder''. In this context, the mean function of the likelihood $\mu_{\textrm{lik}}(\textbf{h}_n,X_n) =f_{\bm{\theta}}(\textbf{h}_n, X_n)$, that is part of the generative model, is called the ``decoder''.

The example from the previous paragraph might sound familiar to some readers and provides indeed a conceptual generalization of a conditional variational autoencoder~\citep{Kingma2015,Sohn2015}. In a vanilla conditional variational autoencoder however, the setting is slightly simplified: the decoder---i.e.\ the neural net $\bm{\theta}$ parameterizing the likelihood mean $\mu_{\textrm{lik}}(\textbf{h}_n,X_n) =f_{\bm{\theta}}(\textbf{h}_n, X_n)$---is treated as a generative parameter $\gamma$ (for which the optimization procedure finds a point estimate) rather than a latent variable over which one seeks to do inference. Coming back to the more general formulation where one seeks to do inference over $\bm{\theta}$. If we replace the neural net weight prior $p(\bm{\theta})$ and the approximate posterior $q_\psi(\bm{\theta})$ with GPs that operate on the concatenated domain of $X_n$ and $\textbf{h}_n$, and replace $f_{\bm{\theta}}(\textbf{h}_n, X_n)$ with $f(\textbf{h}_n, X_n)$ accordingly where $f(\cdot)$ denotes a GP random function, we would obtain the GP equivalent of the example from the previous paragraph. The latter is similar to the work of~\cite{Dutordoir2018}.

Note that if we parameterize the approximate posterior over $\textbf{h}_n$ as $q_\psi(\textbf{h}_n|X_n)$ with a neural net that maps from $X_n$ only (ignoring $y_n$) to a mean vector and covariance matrix for $\textbf{h}_n$, we would sacrifice the label information during training but could predict $y^\star_n$ for unseen $X^\star_n$ context-dependently:
\begin{equation}
\label{eq:pred_lv_iid_alt}
p(y_1^\star, ..., y_{N^\star}^\star | X_1^\star, ..., X_{N^\star}^\star) =  \int \prod_{n=1}^{N^\star}  \int p_\gamma(y_n^\star | \theta, h_n, X_n^\star) q_\psi(h_n| X^\star_n) \;\mathrm{d} h_n \; q_\psi(\theta) \; \mathrm{d}\theta ,
\end{equation}
where we could now make use of the approximate posterior $q_\psi(h_n| X^\star_n)$ that is conditioned on $X^\star_n$ as opposed to Equation~\eqref{eq:pred_lv_iid} where we were forced to make use of the less informative prior $p_\gamma(h_n)$ instead. Also note that we have deliberately reverted our notation in Equation~\eqref{eq:pred_lv_iid_alt} back from boldface $\bm{\theta}$~and~$\textbf{h}_n$ to $\theta$~and~$h_n$ in order to be notationally consistent with Equation~\eqref{eq:pred_lv_iid}.

While we have chosen a supervised learning example as running theme in this tutorial to present multiple VI objectives, latent-variable VI is also often used in the context of unsupervised learning where there are no inputs $X$ but only ``labels'' $y$ whose generative process one seeks to do inference over. In Appendix~\ref{sec:unsupervised_vi}, we have therefore also added a section for unsupervised latent-variable VI for the sake of completeness, but in a nutshell, the formulations there are essentially the same as in this section just ignoring the explicit ``dependence'' on the input $X$.

\subsection{Importance-Weighted Latent-Variable Variational Inference}
\label{sec:iwlv_vi}

Following Equation~\eqref{eq:iw_vi_3}, we can straightforwardly go ahead and combine the idea of latent variables with the importance weighting trick in order to arrive at a tighter lower bound to the $\ELBO$ (that has less estimation variance):
\begin{equation}
\label{eq:iw_vi_lv}
\ELBO_S(\gamma,\psi) = \mathbb{E}_{\substack\prod_{s=1}^S q_\psi(\theta^{(s)})  q_\psi(h^{(s)})} \Bigg[ \ln \frac{1}{S} \sum_{s=1}^S  \frac{p_\gamma(y | \theta^{(s)}, h^{(s)}, X) p_\gamma(\theta^{(s)}) p_\gamma(h^{(s)})}{q_\psi(\theta^{(s)})q_\psi(h^{(s)})} \Bigg],
\end{equation}
from which we could obtain the corresponding formulation for an i.i.d.\ dataset $\{(y_n, X_n)\}_{n=1,..,N}$ by replacing the likelihood and the posterior approximation for $h$ (as well as the prior for $h$) with their factorized counterparts.

However, there is an alternative way to combine importance weighting with the latent-variable formulation. Since $h$ can be considered as additional likelihood input in addition to $\theta$ and $X$, as explained in the previous section, we can imagine the term $\int p(y|\theta,h,X)p_\gamma(h) \textrm{d} h$ as actual likelihood of $y$ given $X$ and $\theta$. We can then proceed with the ordinary VI formulation via an approximate posterior $q_\psi(\theta)$. This leaves us with a likelihood term that contains an integral over $h$, which we can lower-bound via importance weighting through approximate inference over $h$. The maths behind this idea is detailed as follows:
\begin{eqnarray}
\ln p_\gamma(y|X) &=& \ln \int \int p(y|\theta, h, X) p_\gamma(h) \; \mathrm{d} h \; p_\gamma(\theta) \; \mathrm{d} \theta \label{eq:iw_lv_marg_lik} \\
&\geq& \int q_\psi(\theta) \ln \int p(y|\theta, h, X) p_\gamma(h) \; \mathrm{d} h \; \mathrm{d} \theta - \KL\Big(q_\psi(\theta) \Big|\Big| p_\gamma(\theta) \Big) \label{eq:iw_lv_marg_lik_2} \\
&\geq& \int q_\psi(\theta) \mathbb{E}_{\prod_{s=1}^S q_\psi(h^{(s)})} \Bigg[ \ln \frac{1}{S} \sum_{s=1}^S \frac{p(y|\theta, h^{(s)}, X) p_\gamma(h^{(s)})}{q_\psi(h^{(s)})} \Bigg] \; \mathrm{d} \theta \nonumber \\
&& - \KL\Big(q_\psi(\theta) \Big|\Big| p_\gamma(\theta) \Big) \label{eq:iw_lv_marg_lik_3},
\end{eqnarray}
where in Equation~\eqref{eq:iw_lv_marg_lik}, we have applied Jensen's inequality at the $\theta$ level, and in Equation~\eqref{eq:iw_lv_marg_lik_2}, we have applied the importance-weighting trick at the level of the marginal term $\int p(y|\theta, h, X) p_\gamma(h) \mathrm{d} h$. This type of derivation encourages explicitly to counteract increased estimation variance of the $\ELBO$ as a consequence of introducing the additional latent variable $h$. We are going to come back to something similar in Section~\ref{sec:lv_shallow_iw} in the context of VI with sparse latent-variable GPs.

We refrain at this stage from the formulation for an i.i.d.\ dataset that can be obtained readily. Also remember that the type of VI chosen (whether importance-weighted or vanilla VI) does not have an impact on how to predict new labels $y^\star$ given previously unseen data examples $X^\star$. Equations~\eqref{eq:pred_lv} and~\eqref{eq:pred_lv_iid} from the previous section on vanilla latent-variable VI remain still valid for importance-weighted latent-variable VI expressions.

\subsection{Bayesian Deep Learning and Bayesian Layers}
\label{sec:bayesian_layers}

It turns out that VI can be applied hierarchically with building blocks that are stacked on top of each other. To that end, imagine $L$ random functions denoted in weight space view as $\{ \theta^{(1)}, \theta^{(2)}, ..., \theta^{(L)} \}$ each of which is sampled from its own respective prior distribution $p_\gamma(\theta^{(l)})$ where $\gamma$ indicates generative parameters and $l \in \{1,2,...,L\}$. The first random function $\theta^{(1)}$ receives as input the data point~$X$ and a sample from a latent variable $h^{(1)}$ distributed according to the prior $p_\gamma(h^{(1)})$, resulting in the random variable $f^{(1)}$. The second random function $\theta^{(2)}$ receives a sample from $f^{(1)}$ as input together with a sample from another latent variable $h^{(2)} \sim p_\gamma(h^{(2)})$, yielding $f^{(2)}$, and so forth. This process is repeated until the last layer $L$ where the random variable $f^{(L)}$ is obtained. The random variable $f^{(L)}$ evaluates the likelihood for the label $y$ as given by $p_\gamma(y|f^{(L)})$. 

\begin{figure}[h!]
\centering
\includegraphics[trim=140 0 90 0,clip,width=0.8\textwidth]{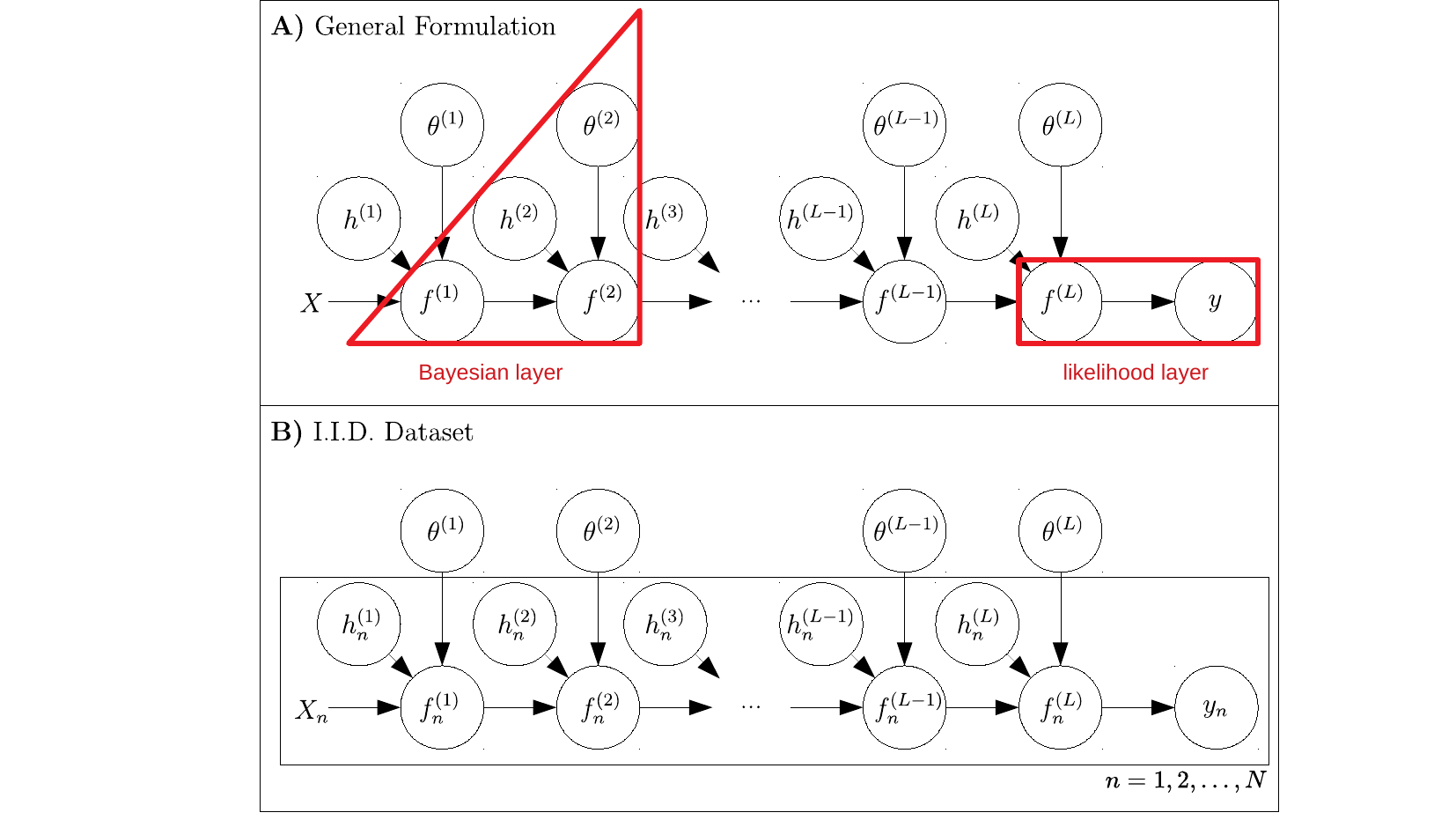}
\caption{Graphical models for deep VI in supervised learning settings. $X$ refers to inputs and $y$ to labels. Random functions are depicted in weight space view through $\theta$-notation (although we could have used alternatively the function space view in $f(\cdot)$-notation). The general formulation is depicted in \textbf{A)} and the formulation for an i.i.d.\ dataset in \textbf{B)} where an index $n$ is added to variables that factorize over samples. The input $X$ and the first latent variable $h^{(1)}$ determine where to evaluate the first random function $\theta^{(1)}$. This yields a random variable $f^{(1)}$ determining together with $h^{(2)}$ where to evaluate the second random function $\theta^{(2)}$, and so on. The last layer depicts the likelihood of $y$ conditioned on $f^{(L)}$. The red triangle on the left indicates (the generative part of) a Bayesian layer. Bayesian layers are stacked on top of one another and serve as building blocks for deep VI. The red rectangle on the right indicates the likelihood layer on top of the last Bayesian layer. Bayesian layers also need to maintain an approximate posterior for approximate inference over their corresponding latent variables (i.e.\ $\theta^{(l)}$ as well as $h^{(l)} \ \forall l$ in the general formulation, and $\theta^{(l)}$ as well as $h_n^{(l)} \ \forall l \text{ and } \forall n$ in the i.i.d.\ formulation), as explained in the main text in more detail.}
\label{fig:vi_deep}
\end{figure}

The graphical model behind the generative process of this hierarchical formulation is depicted in Figure~\ref{fig:vi_deep} A) where the ``likelihood layer'' is highlighted through a red rectangle on the right. All the previous layers before the likelihood layer are building blocks with their own random function (that receives as input the output from the previous layer, as well as a block-specific latent random variable). These blocks are referred to as ``Bayesian layers''~\citep{Tran2019}, the second of which is highlighted through a red triangle in Figure~\ref{fig:vi_deep} A) on the left. 
Note that the latent variables $h^{(l)}$ are optional for each layer and having one $h^{(l)}$ sitting at each layer $l$ might be practically an overkill in terms of model flexibility, but we chose to represent the most general case.

In order to perform inference over the latent functions $\theta$ and variables $h$, we need to introduce a posterior approximation for the joint distribution over all $\theta$ and $h$. A common way to model the posterior approximation is as pairwise independent in latent variables and functions, yielding $q_\psi(\theta^{(1)}) q_\psi(h^{(1)}) \cdot \cdot \cdot q_\psi(\theta^{(L)}) q_\psi(h^{(L)})$ where $\psi$ indicates variational parameters as earlier. Each $(q_\psi(\theta^{(l)}), q_\psi(h^{(l)}))$-pair is part of the corresponding Bayesian layer $l$. Under this assumption, the $\ELBO$ looks as follows:
\begin{eqnarray}
\ELBO(\gamma, \psi) &=&  \int q_\psi(f^{(L)})  \ln p_\gamma(y | f^{(L)}) \;  \mathrm{d} f^{(L)} \nonumber \\
&& -\sum_{l=1}^L \KL\Big(q_\psi(h^{(l)}) \Big|\Big| p_\gamma(h^{(l)}) \Big) -\sum_{l=1}^L \KL\Big(q_\psi(\theta^{(l)}) \Big|\Big| p_\gamma(\theta^{(l)}) \Big), \label{eq:elbo_lv_deep}
\end{eqnarray}
where $q_\psi(f^{(L)})$ refers to the approximate posterior marginal over $f^{(L)}$ marginalized over all $\theta^{(l)}$ and $h^{(l)}$ for all layers $l$ ranging from $1$ to $L$, but also marginalized over all random variables $f^{(l)}$ for all layers ranging from $1$ to $L-1$ (except the last of course). Note that each layer has its own respective latent-variable and latent-function $\KL$ term. 

While propagating a single input $X$ through the posterior approximations of the Bayesian layers (and the likelihood layer) in order to evaluate the expected log likelihood term in Equation~\eqref{eq:elbo_lv_deep}, every layer needs to ``keep track'' of its contribution to the summed $\KL$ terms (that are part of the final $\ELBO$ objective). Furthermore, note that an end-to-end differentiable system is obtained via reparameterizing $\theta^{(l)}$, $h^{(l)}$ and $f^{(l)}$ at each layer~\citep{Kingma2014,Rezende2014}---the last $f^{(L)}$ then becomes a function of the variational parameters~$\psi$ of all the distributions $q_\psi(\theta^{(l)})$ and $q_\psi(h^{(l)})$ from $l=1$ up to $L$. The expected log likelihood term in Equation~\eqref{eq:elbo_lv_deep} can then be readily approximated via samples obtained from randomly propagating $X$ separately multiple times through all the layers.

Predicting new $y^\star$ for unseen $X^\star$ in the deep VI model from above is then accomplished via:
\begin{equation}
\label{eq:pred_lv_deep}
p(y^\star | X^\star) = \int  p_\gamma(y^\star |  {f^\star}^{(L)}) q_\psi({f^\star}^{(L)})  \; \mathrm{d} {f^\star}^{(L)},
\end{equation}
where, similar to Equation~\eqref{eq:elbo_lv_deep}, $q_\psi({f^\star}^{(L)})$ refers to the marginal distribution over ${f^\star}^{(L)}$ as a result of marginalizing the approximate posterior over the latent variables $\theta^{(l)}$ and $h^{(l)}$ for all layers, as well as marginalizing over the latent variables ${f^\star}^{(l)}$ for all layers except the last. The superscript star notation $^\star$ refers to propagating new samples $X^\star$ through the approximate posterior model.

Figure~\ref{fig:vi_deep} B) illustrates the graphical model for the generative process under an i.i.d.\ data assumption $\{(y_n, X_n)\}_{n=1,..,N}$ where latent variables $h_n^{(l)}$ at each layer $l$ factorize over samples $n$ but latent functions $\theta^{(l)}$ do not. Under an i.i.d.\ data assumption, the $\ELBO$ is similar to Equation~\eqref{eq:elbo_lv_deep}:
\begin{eqnarray}
\ELBO(\gamma, \psi) &=& \sum_{n=1}^N \int q_\psi(f_n^{(L)})  \ln p_\gamma(y_n | f_n^{(L)}) \;  \mathrm{d} f_n^{(L)} \nonumber \\
&& - \sum_{n=1}^N \sum_{l=1}^L \KL\Big(q_\psi(h_n^{(l)}) \Big|\Big| p_\gamma(h_n^{(l)}) \Big) - \sum_{l=1}^L \KL\Big(q_\psi(\theta^{(l)}) \Big|\Big| p_\gamma(\theta^{(l)}) \Big), \label{eq:elbo_lv_deep_2}
\end{eqnarray}
except that there is a separate expected log likelihood term for each data point $n$, and a separate latent-variable $\KL$ term for $h_n^{(l)}$ at each layer $l$ and data point $n$. There is however only one latent-function $\KL$ term for each function $\theta^{(l)}$ at each layer $l$ because function parameters $\theta^{(l)}$ are the same for different data points $n$. Therefore, the latent-function $\KL$ terms are referred to as ``global'' whereas the the latent-variable $\KL$ terms as ``local''. 

Predicting new $y_n^\star$ given new data points $X_n^\star$ requires to adjust Equation~\eqref{eq:pred_lv_deep} accordingly, yielding:
\begin{equation}
\label{eq:pred_lv_deep_2}
p(y_1^\star, ..., y_{N^\star}^\star | X_1^\star, ..., X_{N^\star}^\star) = \int  \prod_{n=1}^{N^\star} p_\gamma(y_n^\star |  {\textbf{f}^\star}^{(L)}) q_\psi({\textbf{f}^\star}^{(L)})  \; \mathrm{d} {\textbf{f}^\star}^{(L)},
\end{equation}
by plugging in the i.i.d.\ likelihood and by computing the expectation with the approximate posterior marginal $q_\psi({\textbf{f}^\star}^{(L)})$. Here, the notation ${\textbf{f}^\star}^{(L)}$ denotes the multivariate random variable at the last layer $L$ that is obtained when jointly propagating all new evaluation points $X_n^\star$ through all the layers.

In order to keep the notation lightweight in this section, we have used non-boldface symbols where possible, which does not mean that the corresponding variables need to be scalars. We have also made mostly use of the weight space view that we thought most readers are more familiar with. 
We are going to revert back to function space view in the next section when addressing VI with sparse GPs, where we combine the contents from Section~\ref{sec:sparse_gps} with the contents from Section~\ref{sec:vi}. This will provide an overview over contemporary GP techniques for principled and flexible approximate inference in a wide variety of problems, that e.g.\ enables usage of non-Gaussian and heteroscedastic likelihoods, in both a memory and computationally efficient manner.

\section{Variational Inference with Sparse GPs}
\label{sec:vi_with_svgps}

Sparse GPs, as introduced in Section~\ref{sec:sparse_gps}, can approximate intractable posterior processes for which there is no closed-form solution (e.g.\ in classification problems like logistic regression) or for which the closed-form solution does exist but is too expensive to compute or too expensive to store in memory (e.g.\ in regression problems with train sets that comprise vast amounts of data points). Posterior process approximation in this context is typically achieved with VI where the sparse GP is parameterized via a predefined number of inducing points (or features) that control memory and computational complexity. Inducing points (and the parameters of the distributions over their corresponding inducing variables) are treated as optimization arguments of an $\ELBO$ objective that is maximized in the course of training. In the following, we assume a supervised learning setting in accordance with previous parts of this tutorial but with a particular focus on an i.i.d.\ data scenario. 

In Section~\ref{sec:shallow}, we explain how to do vanilla VI with sparse GP models, including some tricks commonly applied in practice. In Section~\ref{sec:lv_shallow}, we introduce latent variables to sparse GPs in order to increase their flexibility, which we extend in Section~\ref{sec:lv_shallow_iw} by the importance-weighting trick. After that, in Section~\ref{sec:deep}, we transition to VI with deep sparse GPs that can naturally handle functions which might be ``not smooth enough'' for shallow GP models. Deep sparse GPs are extended by latent variables in Section~\ref{sec:lv_deep} and combined with importance weighting in Section~\ref{sec:lv_deep_iw}. Finally, in Section~\ref{sec:demo}, we compare different sparse GP models on a synthetic example to highlight the practical benefit of importance-weighted deep latent-variable models when it comes to tackling a non-smooth and multimodal regression problem.

\subsection{Shallow Sparse GPs}
\label{sec:shallow}

Let's commence with Bayesian inference for supervised learning with an i.i.d.\ training data set $\{(y_n, X_n)\}_{n=1,..,N}$ of size $N$, where $y_n$ refers to a real-valued label associated with the training example $X_n$. The likelihood then factorizes over examples $n$, and the probability of observing a single $y_n$ given the corresponding $X_n$ is denoted as $p_\gamma(y_n | f(\cdot), X_n) = p_\gamma(y_n | f(X_n))$ where $f(\cdot)$ refers to an unknown function that is evaluated at $X_n$ and for which we assume some prior singleoutput GP $p_\gamma(f(\cdot))$. For convenience, we again assume that $\gamma$ refers to all generative parameters and that $f(\cdot)$ is a real-valued function (although the subsequent formulations remain valid under problems with multidimensional labels, vector-valued functions and multioutput GPs). The exact posterior process over $f(\cdot)$ can then be expressed with Bayes' rule as:
\begin{equation}
\label{eq:bayes_rule_f_space}
p_\gamma(f(\cdot)|y_1, ..., y_N, X_1, ..., X_N) = \frac{\prod_{n=1}^N p_\gamma(y_n | f(X_n)) p_\gamma(f(\cdot))}{\int \prod_{n=1}^N p_\gamma(y_n | f(X_n)) p_\gamma(f(\cdot)) \mathrm{d}f(\cdot)},
\end{equation}
which is, in the most general setting under arbitrary likelihoods, no longer guaranteed to be a GP. If the likelihood was Gaussian with $f(X_n)$ being the mean for a particular $y_n$ and under a fixed variance, then the posterior process would be a GP in closed form~\citep{Rasmussen2006}. However, this closed-form solution requires to invert a matrix with rows and columns equal to the number of training examples, which is cubic in $N$. The posterior is hence computationally intractable for large $N$ even though a closed-form solution does exist.

By resorting to VI under an approximate posterior sparse GP as given in Equation~\eqref{eq:sparse_mgp} and in the following denoted as $q_{\psi, \gamma}(f(\cdot))$, one can readily handle non-Gaussian likelihoods and control both memory and computational complexity at the same time. We ask the reader at this point to not get confused about the subscripts for variational and generative parameters $\psi$ and $\gamma$. This aspect is a bit subtle and different from ordinary VI formulations where the approximate posterior does not depend on generative parameters $\gamma$ but only on variational parameters $\psi$ as expected. The ``double dependence'' is due to the fact that the sparse GP is obtained via conditioning the prior GP on inducing variables which makes it hence depend on the prior GP's hyperparameters (which are generative). We are going to explain this in more detail in a subsequent paragraph.

The following $\ELBO$ expression is agnostic to how inducing features are chosen, i.e.\ it is valid for ordinary inducing points but also for interdomain features. Under the i.i.d.\ setting, the $\ELBO$ for sparse GPs can be written as:
\begin{eqnarray}
\ELBO(\gamma,\psi) = \sum_{n=1}^N \int q_{\psi, \gamma}(f(\cdot)) \ln p_\gamma(y_n | f(\cdot), X_n) \; \mathrm{d}f(\cdot)  - \KL\Big(q_{\psi, \gamma}(f(\cdot)) \Big|\Big| p_{\gamma}(f(\cdot)) \Big)  \label{eq:elbo_sparse_gp_iid_0} \\
 = \sum_{n=1}^N \int q_{\psi, \gamma}(f(X_n)) \ln p_\gamma(y_n | f(X_n)) \; \mathrm{d}f(X_n) - \KL\Big(q_{\psi}(\vu) \Big|\Big| p_{\psi,\gamma}(\vu) \Big) \label{eq:elbo_sparse_gp_iid}, \;
\end{eqnarray}
where the variational parameters $\psi$ refer to the mean and covariance of the multivariate Gaussian distribution $q_\psi(\vu)$ over inducing variables $\vu$---see Equation~\eqref{eq:other_marginal_u}---, as well as to inducing point locations $Z_1, ..., Z_M$ (or parameters of inducing features if they contain optimizable parameters). The generative parameters $\gamma$ typically comprise hyperparameters of the kernel and the likelihood (e.g.\ the likelihood variance in case of a homoscedastic Gaussian likelihood). The term $p_{\psi,\gamma}(\vu)$ refers to the prior distribution over inducing variables $\vu$ induced by the prior process $p_\gamma(f(\cdot))$. In the ordinary inducing point formulation, the inducing variables prior $p_{\psi,\gamma}(\vu)$ is simply the result of evaluating the prior GP at the inducing points $Z_m$. This explains the subscript $\psi$ in $p_{\psi,\gamma}(\vu)$ because $p_{\psi,\gamma}(\vu)$ implicitly depends on inducing point locations $Z_m$ that are variational parameters.

We will shortly explain how to go from Equation~\eqref{eq:elbo_sparse_gp_iid_0} to Equation~\eqref{eq:elbo_sparse_gp_iid}, but before that, highlight a peculiarity that is different from ordinary VI objectives regarding variational and generative parameters $\psi$ and $\gamma$. The approximate posterior process $q_{\psi, \gamma}(f(\cdot)) = \int p_{\psi,\gamma}(f(\cdot)|\vu) q_{\psi}(\vu) \mathrm{d} \vu$, that lead to the formulation of a sparse GP in Equation~\eqref{eq:sparse_mgp} in the first place, depends by definition not only on variational parameters $\psi$ but also on generative parameters $\gamma$. This is because the term $p_{\psi,\gamma}(f(\cdot)|\vu)$ is the prior GP $p_\gamma(f(\cdot))$ conditioned on inducing variables $\vu$, and the prior GP is part of the generative model. So parameters of the prior process, like kernel hyperparameters, impact the approximate posterior GP directly. 
Be also aware that the notation $p_{\psi,\gamma}(f(\cdot)|\vu)$ hides some ``dependencies'', e.g.\ in the inducing point formulation, the inducing variables $\vu$ are ``assigned to'' inducing points $Z_m$ which are variational parameters, hence explaining the subscript $\psi$ in $p_{\psi,\gamma}(f(\cdot)|\vu)$. 

In going from Equation~\eqref{eq:elbo_sparse_gp_iid_0} to Equation~\eqref{eq:elbo_sparse_gp_iid}, we have replaced integrals over infinite-dimensional random functions with integrals over finite-dimensional random variables, explained as follows. The expected log likelihood term in Equation~\eqref{eq:elbo_sparse_gp_iid} is merely a result of marginalization over $f(\cdot)$ as a consequence of the i.i.d.\ setting and the  functional form imposed on the likelihood: $y_n$ is conditioned on $f(\cdot)$ evaluated at $X_n$, and does not depend on function values at evaluation points other than $X_n$. 

The second $\KL$ term however requires a bit of explanation. It represents the $\KL$ divergence between $f(\cdot)$ under the approximate posterior $q_{\psi,\gamma}(f(\cdot))$ and under the prior $p_{\gamma}(f(\cdot))$, and is mathematically equivalent to the finite-dimensional $\KL$ between the variational distribution $q_\psi(\vu)$ over the inducing variables $\vu$ and the distribution over $\vu$ under the prior process $p_{\psi,\gamma}(\vu)$. The latter equivalence can be shown following~\cite{Matthews2016} that provides a mathematically rigorous treatment of handling integrals over uncountably infinite objects like $f(\cdot)$. We provide here a more concise but ad hoc derivation that contains integrals over $f(\cdot)$ which then vanish due to marginalization:
\begin{eqnarray}
\KL\Big(q_{\psi,\gamma}(f(\cdot)) \Big|\Big| p_{\gamma}(f(\cdot)) \Big) = \int q_{\psi,\gamma}(f(\cdot)) \ln \frac{q_{\psi,\gamma}(f(\cdot))}{p_{\gamma}(f(\cdot))}  \; \mathrm{d}f(\cdot) \label{eq:kl_derivation_0} \\
=\int q_{\psi,\gamma}(f(\cdot)) \ln \frac{p_\psi(\vu | f(\cdot))q_{\psi,\gamma}(f(\cdot))}{p_\psi(\vu | f(\cdot))p_{\gamma}(f(\cdot))}  \; \mathrm{d}f(\cdot) \; \;  \; \; \; \;  \; \; \;  \; \; \; \;  \; \; \; \; \; \; \; \; \label{eq:kl_derivation_0.125} \\
=\int \int q_{\psi,\gamma}(f(\cdot),\vu) \ln \frac{q_{\psi,\gamma}(f(\cdot),\vu)}{p_{\psi,\gamma}(f(\cdot),\vu)} \; \mathrm{d}\vu \; \mathrm{d}f(\cdot) \; \;  \; \; \; \;  \; \; \; \;  \; \; \; \; \; \; \; \; \; \; \label{eq:kl_derivation_0.25} \\
 =\int \int p_{\psi,\gamma}(f(\cdot)|\vu) q_\psi(\vu) \ln \frac{\cancel{p_{\psi,\gamma}(f(\cdot)|\vu)} q_\psi(\vu)}{\cancel{p_{\psi,\gamma}(f(\cdot)|\vu)} p_{\psi,\gamma}(\vu)} \; \mathrm{d}\vu \; \mathrm{d}f(\cdot) \label{eq:kl_derivation_0.5} \\
 = \int q_\psi(\vu) \ln \frac{q_\psi(\vu)}{p_{\psi,\gamma}(\vu)} \; \mathrm{d}\vu \; = \; \KL\Big(q_\psi(\vu) \Big|\Big| p_{\psi,\gamma}(\vu) \Big) \label{eq:kl_derivation}. \; \; \; \; \; \; \; \; \; \; \; 
\end{eqnarray}
The crucial part in the above equations is to understand how to get from Equation~\eqref{eq:kl_derivation_0} to~\eqref{eq:kl_derivation_0.25} where we have introduced the joint distributions over $f(\cdot)$ and $\vu$ under the approximate posterior $q_{\psi,\gamma}(f(\cdot), \vu)$ and the prior $p_{\psi,\gamma}(f(\cdot), \vu)$. This step looks unintuitive because of the way the integration over $\vu$ is introduced. We start by amending the fraction inside the log of Equation~\eqref{eq:kl_derivation_0} by the conditional distribution $p_\psi(\vu|f(\cdot))$, which is a Dirac delta function that is induced by a linear transformation through the interdomain features $\phi_m$ (or $Z_m$ in the ordinary inducing point formulation), hence the subscript $\psi$ because the interdomain transformation can contain variational parameters. Remember that $\vu$ is a function of $f(\cdot)$ and completely determined by~$f(\cdot)$, and that the log of Equation~\eqref{eq:kl_derivation_0.125} contains inside the fraction of the joint over $f(\cdot)$ and $\vu$ under the approximate posterior and the prior (because the conditional $p_\psi(\vu|f(\cdot))$ is the same for both). The integral over the inducing variables $\vu$ can therefore be introduced in Equation~\eqref{eq:kl_derivation_0.25} where the joint assigns all probability mass to one particular $\vu$ for a given $f(\cdot)$. In Equation~\eqref{eq:kl_derivation_0.5}, we then express the joints over $f(\cdot)$ and $\vu$ inside the log ``the other way around'' with the other conditional-marginal pairs. Because of the way the approximate posterior has been defined, the term $p_{\psi,\gamma}(f(\cdot)|\vu)$ cancels. Since the log then no longer depends on $f(\cdot)$, we can marginalize over $f(\cdot)$ yielding a finite-dimensional integral over $\vu$.

\begin{figure}[h!]
\centering
\includegraphics[trim=0 100 260 30,clip,width=0.8\textwidth]{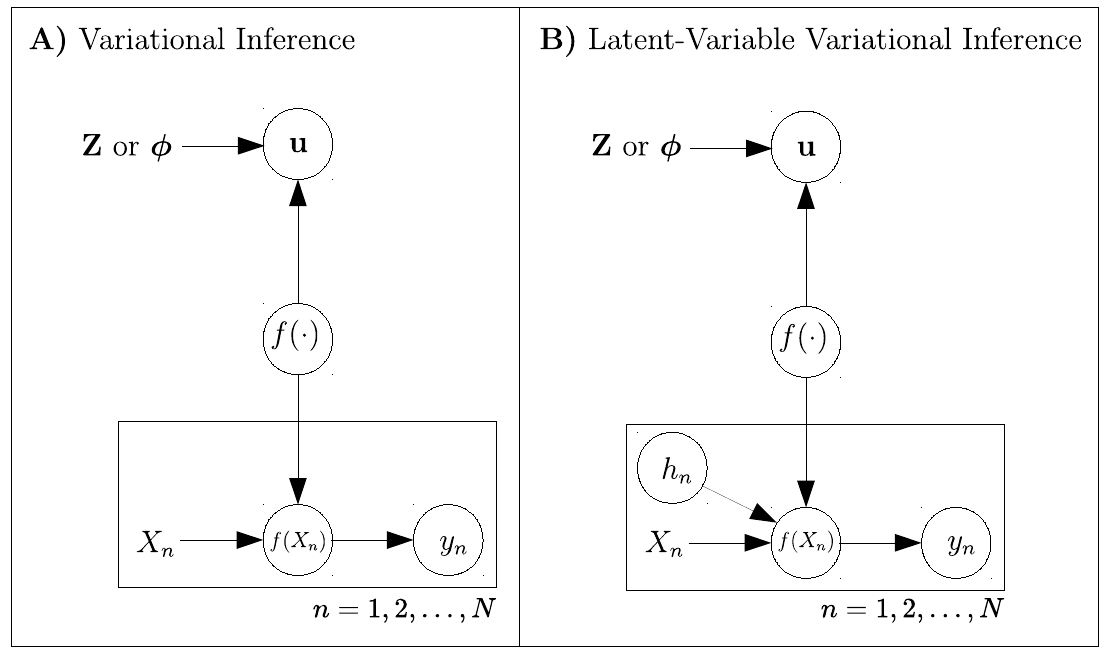}
\caption{Graphical models for VI with sparse GPs in an i.i.d.\ supervised learning setting. $X_n$ are training examples and $y_n$ are labels. Both, random functions $f(\cdot)$ and inducing variables $\vu$, are treated as latent variables. Evaluating a random function $f(\cdot)$ at a specific input location $X_n$ yields the random variable $f(X_n)$. The latter determines the likelihood of the label $y_n$ that corresponds to $X_n$. Inducing variables $\vu$ are associated with inducing points $\textbf{Z}$ (in the inducing point formulation) or with inducing features $\bm{\phi}$ (in the interdomain formulation). The bold-face notation means that all $M$ inducing points (or features) and their corresponding inducing variables are collapsed into one variable. This is necessary because the individual elements of the vector $\vu$ are not pairwise independent. For the sake of memory and computational efficiency, the number of inducing points or features $M$ is usually much smaller than the number of training examples $N$ (indexed by $n$). The graphical model behind ordinary VI with sparse GPs is depicted in \textbf{A}) and subject of Section~\ref{sec:shallow}. The graphical model behind the corresponding latent-variable formulation, that requires another latent variable $h_n$, is depicted in \textbf{B})---this is subject of Section~\ref{sec:lv_shallow}. Also note that inducing points/features are not part of the generative model: given $X_n$, $\vu$ is not required to generate $y_n$ under the prior. The technical reason is that the node for $f(\cdot)$ has two outgoing arrows. Inducing points/features are merely part of the approximate posterior to infer $f(\cdot)$ and are hence variational parameters.}
\label{fig:vi_sgp}
\end{figure}

At this point, it might be insightful to present the graphical model for sparse GP approximate inference as illustrated in Figure~\ref{fig:vi_sgp}~A). We have deliberately omitted the graphical model so far because we feel it is not necessarily intuitive nor helpful to start with for educational purposes. The reason is that the graphical model normally only contains nodes for random variables that are part of the generative process. The graphical model for sparse GP approximate inference however also contains a ``variational'' node for the inducing variable $\vu$ to demonstrate how the sparse GP connects to the prior GP that is part of the generative process.

A practical advantage of the $\ELBO$ in Equation~\eqref{eq:elbo_sparse_gp_iid} is that the $\KL$ term has an analytical expression because of the Gaussian assumptions. In case of a Gaussian likelihood, the expected log likelihood term also enjoys a closed form expression. However, if the number of training data points $N$ is too large, summing over all examples might be infeasible. One can then make use of minibatching in order to obtain an unbiased estimate of the expected log likelihood term~\citep{Hensman2013}. 

Under other likelihood models, for which no closed-form expressions exist, one has to resort to Monte Carlo methods or Gauss-Hermite quadrature~\citep{Hensman2015}. In the latter case, the Gaussian random variable $f(X_n)$ is reparameterized~\citep{Kingma2014,Rezende2014} for optimization purposes. Also note that for singleoutput settings, where $f(X_n)$ is just a scalar, there is normally not much loss of accuracy incurred by Gauss-Hermite quadrature when approximating the expected log likelihood term.

Predictions $\{y_n^\star\}_{n=1,..,N^\star}$ for a new data set $\{X_n^\star\}_{n=1,..,N^\star}$ are straightforward with sparse GPs by multiplying the likelihood with the approximate posterior GP and integrating over function values:
\begin{equation}
\label{eq:pred_sparse_gp_iid}
p(y_1^\star, ..., y_{N^\star}^\star | X_1^\star, ..., X_{N^\star}^\star) =  \int \prod_{n=1}^{N^\star} p_\gamma(y^\star_n | f(X^\star_n)) q_{\psi,\gamma}(\textbf{f}^\star) \; \mathrm{d}\textbf{f}^\star ,
\end{equation}
where $q_{\psi,\gamma}(\textbf{f}^\star)$ denotes the multivariate Gaussian obtained when evaluating the approximate posterior GP $q_{\psi,\gamma}(f(\cdot))$ at the new observations $X_1^\star, ..., X_{N^\star}^\star$. For a Gaussian likelihood, the predictive distribution has a closed form solution because it is Gaussian. For other likelihoods, one needs to resort once more to Monte Carlo methods or Gauss-Hermite quadrature.

At this point, the reader should have obtained a good understanding of VI with sparse GP models. We proceed with concluding this section by some parameterization tricks for inducing variables $\vu$ that come in handy for practical applications. First, let's remind ourselves of how the variational distribution $q_\psi(\vu)$ over inducing variables is parameterized, namely as a multivariate Gaussian with a mean vector $\qu$ and a covariance matrix $\Quu$---see Equation~\eqref{eq:other_marginal_u} from the very beginning of this manuscript. The variable $\qu$ enters the definition of a sparse GP in Equation~\eqref{eq:sparse_mgp} in terms of a difference $\qu - \muu$ in the mean function, where $\muu$ is the mean of $\vu$ under the prior. We can also define an alternative inducing variable $\vv:= \vu - \muu$ that is corrected under the prior mean $\muu$, and parameterized via a mean vector $\qv$ (the covariance matrix $\Quu$ stays the same), yielding:
\begin{equation}
\label{eq:sparse_mgp_v2}
f(\cdot) \ \sim \ \mathcal{GP} \Big( \mu(\cdot) + \kfu \Kuu^{-1} \qv, k(\cdot, \cdot^\prime) - \kfu \Kuu^{-1} (\Kuu - \Quu) \Kuu^{-1} \kuf \Big).
\end{equation}
One might wonder what is the advantage of this alternative parameterization. The answer is that replacing $\qu - \muu$ with $\qv$ in Equation~\eqref{eq:sparse_mgp} is particularly helpful in the interdomain formulation introduced in Section~\ref{sec:interdomain_gps}, where computing $\muu$ according to Equation~\eqref{eq:muui_id} would require to integrate the product between the mean function $\mu(\cdot)$ and each feature function $\phi_m(\cdot)$ over the input domain $\mathcal{X}$. These integrals do, in general, not have closed-form expressions for any mean function. Under the alternative parameterization with $\vv$ however, such integration becomes unnecessary---using arbitrary mean functions is hence no longer a problem. 

The ``problematic'' term $\muu$ also appears in the $\KL$ term in the $\ELBO$ from Equation~\eqref{eq:elbo_sparse_gp_iid}. This $\KL$ term however becomes more convenient to evaluate as $\KL\Big(q_{\psi}(\vv) \Big|\Big| p_{\gamma}(\vv) \Big)$ under the $\vv$ parameterization. This is possible because the $\KL$ is invariant under the ``translational'' transformation of the random variable $\vu$ into $\vv$. Under the ``transformed'' $\KL$, the distribution $p_{\gamma}(\vv)$ over $\vv$ under the prior has zero mean with covariance matrix $\Kuu$.

Exploiting that the $\KL$ is in fact invariant under ``change-of-variables'' transformations (not just translational transformations) of the random variable~$\vu$ (or~$\vv$ respectively), we can go one step further and define the inducing variable $\vw := \chol(\Kuu) ^{-1} \vv$ with mean vector $\qw$ and covariance matrix $ \Qww$, and where $\chol$ refers to Cholesky decomposition. This yields $\KL\Big(q_{\psi}(\vw) \Big|\Big| p_{\gamma}(\vw) \Big)$ in the $\ELBO$ from Equation~\eqref{eq:elbo_sparse_gp_iid}, where $p_{\gamma}(\vw)$ becomes the standard normal multivariate Gaussian (of dimension~$M$). The latter parameterization requires in Equation~\eqref{eq:sparse_mgp_v2} to replace $\qv$ with $\chol(\Kuu) \qw$ and $\Quu$ with $\chol(\Kuu) \Qww \chol(\Kuu)^{\top}$. Denoting $\chol(\Kuu)$ as $\Luu$, we arrive at the following sparse GP formulation:
\begin{equation}
\label{eq:sparse_mgp_v3}
f(\cdot) \ \sim \ \mathcal{GP} \Big( \mu(\cdot) + \kfu \Luu^{-\top} \qw, k(\cdot, \cdot^\prime) - \kfu \Luu^{-\top} (\mathbf{I} - \Qww) \Luu^{-1} \kuf \Big),
\end{equation}
where we have made use of the identity $\Kuu^{-1} = \Luu^{-\top} \Luu^{-1}$. Note that this parameterization of the inducing variable is a specific form of whitening that can facilitate optimizing the $\ELBO$ from Equation~\eqref{eq:elbo_sparse_gp_iid} and is standard functionality provided in contemporary GP software frameworks~\citep{Matthews2017}. 

Another common trick is to initialize generative parameters (like kernel lengthscales for example) to ``reasonable'' values (if such values are known), and then optimize initially for the variational parameters only while keeping the generative parameters fixed. Once reasonable values for the variational parameters have been identified (e.g.\ after a fixed number of training iterations), generative parameters are ``unclamped'' and jointly optimized together with the variational parameters. Initializing generative parameters, like kernel hyperparameters, such that they are very far from optimal values can lead to large gradients that can impede the optimization process.

Finally, note that, as mentioned earlier, all formulations presented in this section remain valid under arbitrary likelihoods, problems with multidimensional labels, vector-valued functions and multioutput GPs. We can for example formulate a regression problem with a heteroscedastic Gaussian likelihood straightforwardly as follows~\citep{Saul2016}. Under vector-valued random functions $\mathbf{f}(\cdot)$ with two outputs, we can define a Gaussian likelihood as $p(y_n|\textbf{f}(X_n))$ with mean $\mu_{\textrm{lik}} = \textbf{f}(X_n)[1]$ and variance $\upsilon_{\textrm{lik}} = g(\textbf{f}(X_n)[2])$, where~$1$ and~$2$ refer to both output indexes and where $g(\cdot)$ is a strictly positive function with real-line support (because GP function values are real-valued but variances need to be positive). This example is the conceptual function space equivalent to the neural network example presented earlier for VI under the weight space formulation in Section~\ref{sec:vanilla_vi}.

\subsection{Latent-Variable Shallow Sparse GPs}
\label{sec:lv_shallow}

According to Section~\ref{sec:lv_vi}, one can introduce another latent variable $h_n$ for each training example $n$ to make the generative model underlying a sparse GP more flexible. The corresponding graphical model is depicted in Figure~\ref{fig:vi_sgp} B). To that end, it is assumed that the prior over $f(\cdot)$ and all $h_n$ factorizes into $p_\gamma(f(\cdot))$ and $\prod_{n=1}^N p_\gamma(h_n)$. The likelihood for a single example $(X_n,y_n)$ is then defined as $p_\gamma(y_n|f(\cdot),h_n,X_n) =  p_\gamma(y_n|f(h_n,X_n))$, where the notation $f(h_n,X_n)$ means that the GP operates on the concatenated domain of the input $X_n$ and the latent variable $h_n$.

To understand why this increases modelling flexibility, imagine a homoscedastic Gaussian likelihood with mean $\mu_{\textrm{lik}} = f(h_n,X_n)$ and fixed variance. If there is no latent variable $h_n$, then the distribution over $y_n$ for a given $f(\cdot)$ and $X_n$ is unimodal. Under the latent-variable formulation however, the likelihood of $y_n$ given $f(\cdot)$ and $X_n$ can be multimodal (when integrated over $h_n$), hence becoming more expressive. When doing VI, we then need to perform inference over all $h_n$ as well. The general assumption is that the approximate posterior factorizes as follows $q_{\psi,\gamma}(f(\cdot)) \prod_{n=1}^N q_\psi(h_n)$, where $q_{\psi,\gamma}(f(\cdot))$ depends on variational parameters $\psi$ and generative parameters $\gamma$, as explained earlier, and $q_\psi(h_n)$ refers to the approximate posterior over $h_n$. The $\ELBO$ is then expressed as follows:
\begin{eqnarray}
\ELBO(\gamma, \psi) &=& \sum_{n=1}^N \int  \int q_{\psi,\gamma}(f(\cdot)) q_\psi(h_n) \ln p_\gamma(y_n | f(\cdot), h_n, X_n) \; \mathrm{d}h_n  \;  \mathrm{d}f(\cdot) \nonumber \\
&& -\sum_{n=1}^N \KL\Big(q_\psi(h_n) \Big|\Big| p_\gamma(h_n) \Big) -\KL\Big(q_{\psi,\gamma}(f(\cdot)) \Big|\Big| p_\gamma(f(\cdot)) \Big) \label{eq:elbo_sparse_gp_lv_iid_0} \\
 &=& \sum_{n=1}^N \int q_\psi(h_n) \int q_{\psi,\gamma}(f(h_n,X_n)) \ln p_\gamma(y_n | f(h_n, X_n)) \; \mathrm{d}f(h_n, X_n) \; \mathrm{d}h_n  \nonumber \\
&& -\sum_{n=1}^N \KL\Big(q_\psi(h_n) \Big|\Big| p_\gamma(h_n) \Big) -\KL\Big(q_{\psi,\gamma}(\vu) \Big|\Big| p_{\psi,\gamma}(\vu) \Big). \label{eq:elbo_sparse_gp_lv_iid}
\end{eqnarray}
The derivation is similar to VI without an additional latent variable as in Equation~\eqref{eq:elbo_sparse_gp_iid}. Note that the integration over $f$ and $h_n$ has ``swapped'' between Equation~\eqref{eq:elbo_sparse_gp_lv_iid_0} and~\eqref{eq:elbo_sparse_gp_lv_iid} in the expected log likelihood term. In line with earlier sections, one can integrate over $f(h_n, X_n)$ evaluated at specific locations $(h_n,X_n)$, as opposed to all of $f(\cdot)$. The ``swap'' then occurs because $h_n$ is part of the input to $f(\cdot)$. Also note that the innermost expectation over $f(h_n, X_n)$ in the first term of Equation~\eqref{eq:elbo_sparse_gp_lv_iid} can be often computed efficiently (e.g.\ in closed form under a Gaussian likelihood) while the outermost expectation does usually not have a closed-form expression.

The approximate posterior $q_\psi(h_n)$ can be chosen independently for each datapoint $n$, or in an amortized fashion via a parametric map $q_\psi(h_n | y_n,X_n)$ that maps training tuples $(y_n,X_n)$ probabilistically to $h_n$. Conceptually, the model presented in this section is similar to the one presented in~\cite{Dutordoir2018}, and can be interpreted as function space equivalent of a generalized version of conditional variational autoencoders~\citep{Kingma2015,Sohn2015}. Conditional variational autoencoders naturally adopt the parameter space view but identify function parameters as point estimates rather than through inference.

While latent-variable sparse GPs are more flexible than ordinary sparse GPs, the $\ELBO$ in Equation~\eqref{eq:elbo_sparse_gp_lv_iid} requires an additional sampling step over $h_n$. This additional sampling step leads to an increase in the variance of $\ELBO$ estimates. In the next section, we are going to present a method to alleviate this problem with the importance-weighting trick from earlier. But before that, let's finish this section with how to make predictions $\{y_n^\star\}_{n=1,..,N^\star}$ for new data examples $\{X_n^\star\}_{n=1,..,N^\star}$:
\begin{equation}
\label{eq:pred_sparse_gp_lv_iid}
p(y_1^\star, ..., y_{N^\star}^\star | X_1^\star, ..., X_{N^\star}^\star) =  \int ... \int \int \prod_{n=1}^{N^\star} p_\gamma(y^\star_n | f(h^\star_n,X^\star_n)) p_\gamma(h^\star_n) q_{\psi,\gamma}(\textbf{f}^\star) \; \mathrm{d}\textbf{f}^\star \; \mathrm{d}h^\star_{N^\star} ... \; \mathrm{d} h^\star_1,
\end{equation}
where $q_{\psi,\gamma}(\textbf{f}^\star)$ refers to the multivariate Gaussian that results from evaluating $q_{\psi,\gamma}(f(\cdot))$ at all $(h^\star_n,X^\star_n)$-pairs. Note that the integral over latent variables $h^\star_n$ is outermost because the multivariate random variable $\textbf{f}^\star$ implicitly depends on all $h^\star_n$---since $f(\cdot)$ receives both $X^\star_n$ and $h^\star_n$ as joint input. Also note that the integral over latent variables usually does not have a closed-form expression and requires approximation.

Finally, remember that the prior $p_\gamma(h^\star_n)$ over latent variables $h^\star_n$ is required when predicting $y_n^\star$ for new data points $X_n^\star$. The latter is because it is naively not possible to evaluate the approximate posterior over latent variables $h_n^\star$ for previously unseen data points $X^\star_n$---neither under the naive formulation $q_\psi(h_n)$ where there is an approximate posterior for each training example $X_n$ but none for unseen data points $X^\star_n$, nor under the amortized formulation where the approximate posterior $q_\psi(h_n | y_n, X_n)$ depends on labels that are unknown for novel data points at prediction time.

\subsection{Importance-Weighted Latent-Variable Shallow Sparse GPs}
\label{sec:lv_shallow_iw}

As outlined in the previous section, one drawback of the latent-variable formulation is that another sampling step over $h_n$ is required when compared to ordinary VI\@. This additional sampling step increases the variance of $\ELBO$ estimates which leads to decreased efficiency during optimization. We have seen in Section~\ref{sec:iwlv_vi} how to exchange computational resources for a tighter lower bound to the $\ELBO$ that, at the same time, can be estimated more reliably with less estimation variance. The latter can be achieved with the importance-weighting trick as described earlier for the parameter space view in Equation~\eqref{eq:iw_lv_marg_lik_3} from Section~\ref{sec:iwlv_vi}. Applying the same kind of reasoning for shallow sparse GPs and i.i.d.\ training data yields in function space view:
\begin{eqnarray}
\ELBO_S(\gamma,\psi) &=& \sum_{n=1}^N \mathbb{E}_{\prod_{s=1}^S q_\psi(h_n^{(s)})q_{\psi,\gamma}(\textbf{f}_n)} \Bigg[ \ln \frac{1}{S} \sum_{s=1}^S \frac{p_\gamma(y_n|\textbf{f}_n[s]) p_\gamma(h_n^{(s)})}{q_\psi(h_n^{(s)})} \Bigg] \nonumber \\
&& - \KL\Big(q_\psi(\vu) \Big|\Big| p_{\psi,\gamma}(\vu) \Big) \label{eq:sparse_gp_iw_lv_marg_lik_3},
\end{eqnarray}
where $S$ denotes the number of importance-weighted replicates for $h_n$ indexed by $s$. The quantity~$\textbf{f}_n$ is the $S$-dimensional multivariate Gaussian random variable obtained when jointly evaluating the approximate posterior GP $q_{\psi,\gamma}(f(\cdot))$ at the locations $(h_n^{(s)}, X_n)$ for all $S$ replications $h_n^{(s)}$ given~$X_n$. Note that there is an implicit outer expectation in Equation~\eqref{eq:sparse_gp_iw_lv_marg_lik_3} w.r.t.\ latent variables $h_n^{(1)}$ up to $h_n^{(S)}$ and an implicit inner expectation w.r.t.\ $\textbf{f}_n$ that depends on latent variables from the outer expectation. The notation $\textbf{f}_n[s]$ refers to the $s$-th component of the vector $\textbf{f}_n$.

For sparse GPs, terms of the form $\int q_{\psi,\gamma}(f(h,X)) \ln p_\gamma(y|f(h,X)) \mathrm{d} f(h,X)$ are often efficiently computable---e.g.\ in closed form in case of a Gaussian likelihood. The problem with Equation~\eqref{eq:sparse_gp_iw_lv_marg_lik_3} is that it does not contain such an expected log likelihood expression, as opposed to the naive $\ELBO$ formulation. In order to take advantage of efficient computability, it is therefore desirable to combine importance weighting with said expected log likelihood expression (following~\cite{Salimbeni2019}). This can be achieved with a sequence of steps as detailed next. We start with introducing importance weights for latent variables $h_n$ in the marginal likelihood:
\begin{eqnarray}
p_\gamma(y_1,...,y_N|X_1,...,X_N) = \mathbb{E}_{\prod_{n=1}^N p_\gamma(h_n)} \Bigg[ p_\gamma(y_1,...,y_N|h_1,...,h_N,X_1,...,X_N)  \Bigg] \\
= \mathbb{E}_{\prod_{n=1}^N q_\psi(h_n)} \Bigg[ p_\gamma(y_1,...,y_N|h_1,...,h_N,X_1,...,X_N) \frac{\prod_{n=1}^N p_\gamma(h_n)}{\prod_{n=1}^N q_\psi(h_n)} \Bigg] \label{eq:first_step_iw_lv_sgp},
\end{eqnarray}
where importance weights are introduced via an approximate posterior $\prod_{n=1}^N q_\psi(h_n)$ over all $h_n$. Note that $p_\gamma(y_1,...,y_N|h_1,...,h_N,X_1,...,X_N)$ is the likelihood of all labels $\{y_1, ..., y_N\}$ conditioned on all latent variables $\{h_1, ..., h_N\}$ and training examples $\{X_1, ...,X_N\}$ but, importantly, marginalized over random functions $f(\cdot)$.

Next, we leverage the ordinary $\ELBO$ to lower-bound $\ln p_\gamma(y_1,...,y_N|h_1,...,h_N,X_1,...,X_N)$ as:
\begin{eqnarray}
\ln p_\gamma(y_1,...,y_N|h_1,...,h_N,X_1,...,X_N) &\geq& \sum_{n=1}^N \mathbb{E}_{q_{\psi,\gamma}(f(h_n,X_n))}  \Bigg[ \ln p_\gamma(y_n|f(h_n,X_n)) \Bigg]  \nonumber \\
&&- \KL\Big(q_\psi(\vu) \Big|\Big| p_{\psi,\gamma}(\vu) \Big) \label{eq:second_step_iw_lv_sgp}.
\end{eqnarray}
Taking the exp of Equation~~\eqref{eq:second_step_iw_lv_sgp} gives a lower bound to $p_\gamma(y_1,...,y_N|h_1,...,h_N,X_1,...,X_N)$ which we can plug into Equation~\eqref{eq:first_step_iw_lv_sgp}. After taking the log and some rearrangements, one arrives at:
\begin{eqnarray}
\ln p_\gamma(y_1,...,y_N|X_1,...,X_N) &\geq& \sum_{n=1}^N \ln \mathbb{E}_{q_\psi(h_n)} \Bigg[ \explik_{\gamma,\psi}(y_n,h_n,X_n) \frac{ p_\gamma(h_n)}{q_\psi(h_n)} \Bigg] \nonumber \\
&& - \KL\Big(q_\psi(\vu) \Big|\Big| p_{\psi,\gamma}(\vu) \Big) \label{eq:third_step_iw_lv_sgp} ,
\end{eqnarray}
where, in order to preserve a clear view, we needed to define the helper term $\explik_{\gamma,\psi}(y_n,h_n,X_n)$ for the exponential of the expected log likelihood term over function values $f(h_n,X_n)$:
\begin{equation}
\label{eq:explik}
\explik_{\gamma,\psi}(y_n,h_n,X_n) := \exp \Bigg( \int q_{\psi,\gamma}(f(h_n,X_n)) \ln p_\gamma(y_n|f(h_n,X_n)) \; \mathrm{d} f(h_n,X_n) \Bigg) .
\end{equation}
The importance-weighting trick can finally be applied to the first log term on the r.h.s.\ of Equation~\eqref{eq:third_step_iw_lv_sgp} by introducing $S$ replicates indexed by $s$ for each $h_n$, as in Equation~\eqref{eq:sparse_gp_iw_lv_marg_lik_3}, yielding:
\begin{eqnarray}
\ln p_\gamma(y_1,...,y_N|X_1,...,X_N) &\geq& \sum_{n=1}^N \mathbb{E}_{\prod_{s=1}^S q_\psi(h_n^{(s)})} \Bigg[ \ln \frac{1}{S} \sum_{s=1}^S \explik_{\gamma,\psi}(y_n,h_n^{(s)},X_n) \frac{ p_\gamma(h^{(s)}_n)}{q_\psi(h^{(s)}_n)} \Bigg] \nonumber \\
&&- \KL\Big(q_\psi(\vu) \Big|\Big| p_{\psi,\gamma}(\vu) \Big) , \label{eq:ugly_log}
\end{eqnarray}
that combines importance weighting on the level of latent variables $h_n^{(s)}$ with the benefit of efficient computability of the expected log likelihood term over function values $f(h_n^{(s)},X_n)$ inside the expression $\explik_{\gamma,\psi}(y_n,h_n,X_n)$ as introduced in Equation~\eqref{eq:explik}. The next couple of sections deal with deep sparse GPs that are obtained by stacking shallow sparse GPs on top of each other, but we are going to return back to the trick of combining efficient computability and importance weighting in Section~\ref{sec:lv_deep_iw} later on after having introduced latent-variable deep sparse GPs.

\subsection{Deep Sparse GPs}
\label{sec:deep}

Following Section~\ref{sec:deep_gps} and more precisely Figure~\ref{fig:deep_rbf}, deep GPs hold the potential to model functions that are less smooth and more abruptly changing (with which an ordinary GP might have problems). In line with the conceptual logic of Bayesian deep learning from Section~\ref{sec:bayesian_layers}, one can create a generative model via stacking multioutput GPs on top of each other and feeding the final outcome into a likelihood. Imagine to that end $L$ vector-valued random functions $\{\textbf{f}^{(1)}(\cdot), ..., \textbf{f}^{(L)}(\cdot)\}$ that are distributed according to $L$ independent prior multioutput GPs with a joint distribution $p_\gamma(\textbf{f}^{(1)}(\cdot)) \cdots p_\gamma(\textbf{f}^{(L)}(\cdot))$. Note that the input domain of the first GP is $\mathcal{X}$ and the input dimension of each other GP equals the output dimension of the previous GP\@. Practically, we recommend to use an identity mean function in prior distributions (where possible, or linear otherwise) in all layers except the last. This is to encourage that different training inputs are mapped to different latent representations, which helps with the learning progress. Mapping different training inputs to the same latent representation on the other hand (as encouraged by a zero-mean function for example) could impede the learning progress.

As explained previously in Section~\ref{sec:deep_gps}, a single data point $X_n$ is propagated through a deep GP as follows (see the second paragraph in Section~\ref{sec:deep_gps} for multiple data points). A random function $\textbf{f}^{(1)}(\cdot)$ from the prior GP $p_\gamma(\textbf{f}^{(1)}(\cdot))$ is evaluated at $X_n$ yielding the vector-valued random variable $\textbf{f}_n^{(1)}$. A single sample from this random variable $\textbf{f}_n^{(1)}$ serves as input to evaluate the second prior GP $p_\gamma(\textbf{f}^{(2)}(\cdot))$ resulting in the random variable $\textbf{f}_n^{(2)}$ (a single sample of which serves as input to the next prior GP, and so on). This ultimately yields the random variable $\textbf{f}_n^{(L)}$, which determines the likelihood $p_\gamma(y_n | \textbf{f}_n^{(L)})$ for the label $y_n$ and the input $X_n$. Under an approximate posterior that factorizes over layers $l \in \{1, ..., L\}$, i.e.\ $q_{\psi,\gamma}(\textbf{f}^{(1)}(\cdot)) \cdots q_{\psi,\gamma}(\textbf{f}^{(L)}(\cdot))$ where each $q_{\psi,\gamma}(\textbf{f}^{(l}(\cdot))$ is a multioutput sparse GP, the $\ELBO$ is then defined as~\citep{Salimbeni2017}:
\begin{eqnarray}
\ELBO(\gamma, \psi) &=& \sum_{n=1}^N \mathbb{E}_{q_{\psi,\gamma}(\textbf{f}_n^{(L-1)})} \Bigg[ \int q_{\psi,\gamma}(\textbf{f}_n^{(L)} | \textbf{f}_n^{(L-1)})  \ln p_\gamma(y_n | \textbf{f}_n^{(L)}) \;  \mathrm{d}\textbf{f}_n^{(L)} \Bigg] \nonumber \\
&&-\sum_{l=1}^L \KL\Big(q_\psi(\vU^{(l)}) \Big|\Big| p_{\psi,\gamma}(\vU^{(l)}) \Big), \label{eq:elbo_deep_sgp}
\end{eqnarray}
where inducing variables $\vU^{(l)}$ are represented in matrix form in line with the output-as-output view. The first dimension of $\vU^{(l)}$ corresponds to the number of inducing features (assuming the same number of inducing features per output for convenience) and the second dimension to the number of outputs at the layer~$l$. It is practically recommended to initialize $q_\psi(\vU^{(l)})$ with low variance in non-terminal layers to ensure a ``deterministic'' information flow during initial stages of training. 

The quantity $q_{\psi,\gamma}(\textbf{f}_n^{(L-1)})$ is the marginal distribution over $\textbf{f}_n^{(L-1)}$, marginalized over all previous random variables from $\textbf{f}_n^{(1)}$ up to $\textbf{f}_n^{(L-2)}$. It is important to note that $q_{\psi,\gamma}(\textbf{f}_n^{(L-1)})$ is not Gaussian and that there is no closed form expression for it---practically, one therefore needs to resort to Monte Carlo methods. On the other side, the term $q_{\psi,\gamma}(\textbf{f}_n^{(L)} | \textbf{f}_n^{(L-1)})$ is Gaussian by construction: it is the conditional probability of the last-layer random variable $\textbf{f}_n^{(L)}$ conditioned on the the second-to-last random variable $\textbf{f}_n^{(L-1)}$. The Gaussianity is simply due to the fact that $q_{\psi,\gamma}(\textbf{f}_n^{(L)} | \textbf{f}_n^{(L-1)})$ refers to the last approximate posterior GP $q_{\psi,\gamma}(f(\cdot))$ evaluated at $\textbf{f}_n^{(L-1)}$. In ordinary non-deep notation, we usually omit conditioning explicitly on evaluation locations $X_n$ because they are normally not considered as random variables. In a deep GP however, for all GPs except the first one, the inputs are random variables over which one needs to average. 

We could have chosen to represent Equation~\eqref{eq:elbo_deep_sgp} in terms of the marginal over $\textbf{f}_n^{(L)}$ without explicitly mentioning $\textbf{f}_n^{(L-1)}$ in the first place. However, because $q_{\psi,\gamma}(\textbf{f}_n^{(L)} | \textbf{f}_n^{(L-1)})$ is Gaussian, the inner expectation of the log likelihood in Equation~\eqref{eq:elbo_deep_sgp} can often be computed efficiently, e.g.\ in case of a Gaussian likelihood. This would have been hidden if we had not emphasized the different expectations of the last and second-to-last layer. In order to obtain an end-to-end differentiable system, it is standard procedure to reparameterize the variables $\textbf{f}^{(l)}$ at each layer $l$ following~\citep{Kingma2014,Rezende2014} in accordance with earlier sections.

Predictions $\{y_n^\star\}_{n=1,..,N^\star}$ for new data points $\{X_n^\star\}_{n=1,..,N^\star}$ are then obtained as follows:
\begin{equation}
\label{eq:pred_deep_sparse_gp_iid}
p(y_1^\star, ..., y_{N^\star}^\star | X_1^\star, ..., X_{N^\star}^\star) =  \mathbb{E}_{q_{\psi,\gamma}({\textbf{F}^\star}^{(L-1)})} \Bigg[ \int \prod_{n=1}^{N^\star} p_\gamma(y^\star_n | {\textbf{F}^\star}^{(L)}[n,:]) q_{\psi,\gamma}({\textbf{F}^\star}^{(L)} | {\textbf{F}^\star}^{(L-1)}) \; \mathrm{d} {\textbf{F}^\star}^{(L)} \Bigg],
\end{equation}
where upper-case bold-face notation $\textbf{F}^\star$ represents a matrix-valued random variable (which naturally arises when a multioutput GP is evaluated at multiple data points in the output-as-output view---see Section~\ref{sec:deep_gps} for a detailed explanation). The first dimension of ${\textbf{F}^\star}^{(l)}$ corresponds to the number of evaluation points (here $N^\star$) and the second dimension to the number of output heads at the layer~$l$. We again separate the expectation over ${\textbf{F}^\star}^{(L-1)}$ and ${\textbf{F}^\star}^{(L)}$ to highlight that the inner expected log likelihood term can often be computed efficiently. 
The notational form ${\textbf{F}^\star}^{(L)}[n,:]$ refers to the $n$-th row of the random variable ${\textbf{F}^\star}^{(L)}$ that sits in the last layer indexed with $L$.
Note how this is different from the notation in Equation~\eqref{eq:elbo_deep_sgp} where we used $\textbf{f}_n^{(L)}$ with a subscript $n$ instead. This is not arbitrary. In Equation~\eqref{eq:elbo_deep_sgp}, every data point $n$ is propagated independently through the deep GP because of marginalization due to the functional form imposed on the likelihood. In Equation~\eqref{eq:pred_deep_sparse_gp_iid} however, all samples $n$ need to be propagated jointly through the deep GP, as is also the case with predictions using shallow GP models presented earlier in Equation~\eqref{eq:pred_sparse_gp_iid}.

In the next section, we will introduce latent variables $h_n^{(l)}$ for each layer $l \in \{1,..., L\}$ to make deep sparse GP models more flexible. In this context, we are going to illustrate the graphical model behind the generative process in Figure~\ref{fig:vi_deep_sgp_latentv}. This graphical model also holds for the ordinary deep sparse GP model from this section if we ignore the nodes for latent variables $h_n^{(l)}$.

\subsection{Latent-Variable Deep Sparse GPs}
\label{sec:lv_deep}

In line with earlier sections, one can increase the flexibility of a deep GP's modelling capabilities by introducing latent variables $h_n^{(l)}$ for each data point $n$ and each layer $l$, distributed according to the joint prior $\prod_{n=1}^N p_\gamma(h_n^{(1)}) \cdots p_\gamma(h_n^{(L)})$. The generative model is similar to the previous section, except that the input domain of each prior GP $p_\gamma(\textbf{f}^{(l)}(\cdot))$ is concatenated with the domain where $h_n^{(l)}$ lives, explained in more detail as follows in the context of propagating a single example $X_n$ through the deep latent-variable GP. 

Evaluating the first prior GP $p_\gamma(\textbf{f}^{(1)}(\cdot))$ requires to draw a sample from the the first latent variable $h_n^{(1)} \sim p_\gamma(h_n^{(1)})$ and concatenate it with $X_n$. The resultant multivariate random variable $\textbf{f}^{(1)}_n$ determines together with $h_n^{(2)}$ where to evaluate the second prior GP $p_\gamma(\textbf{f}^{(2)}(\cdot))$, and so on. The final random variable $\textbf{f}_n^{(L)}$ eventually determines the likelihood $p_\gamma(y_n | \textbf{f}_n^{(L)})$ of the label $y_n$ associated with $X_n$. Under the assumption of a factorized approximate sparse posterior process $q_{\psi,\gamma}(\textbf{f}^{(1)}(\cdot)) \cdots q_{\psi,\gamma}(\textbf{f}^{(L)}(\cdot))$, as in the previous section, and after introducing a factorized approximate posterior for the additional latent variables $\prod_{n=1}^N q_\psi(h_n^{(1)}) \cdots q_\psi(h_n^{(L)})$, the $\ELBO$ can be expressed as~\citep{Salimbeni2019}:
\begin{eqnarray}
\ELBO(\gamma, \psi) &=& \sum_{n=1}^N \mathbb{E}_{q_{\psi,\gamma}(\textbf{f}_n^{(L-1)})} \Bigg[ \int q_\psi(h_n^{(L)}) \int q_{\psi,\gamma}(\textbf{f}_n^{(L)} | h_n^{(L)}, \textbf{f}_n^{(L-1)}) \ln p_\gamma(y_n | \textbf{f}_n^{(L)}) \;  \mathrm{d}\textbf{f}_n^{(L)} \;  \mathrm{d}h_n^{(L)} \Bigg] \nonumber \\
&&-\sum_{n=1}^N \sum_{l=1}^L \KL\Big(q_\psi(h_n^{(l)}) \Big|\Big| p_\gamma(h_n^{(l)}) \Big) -\sum_{l=1}^L \KL\Big(q_\psi(\vU^{(l)}) \Big|\Big| p_{\psi,\gamma}(\vU^{(l)}) \Big), \label{eq:elbo_deep_lv_sgp}
\end{eqnarray}
where there is an additional latent-variable $\KL$ term for each data point and each layer because of the factorization assumptions in both the prior and the approximate posterior over $h_n^{(l)}$. Note that $q_{\psi,\gamma}(\textbf{f}_n^{(L-1)})$ refers to the marginal distribution over $\textbf{f}_n^{(L-1)}$, marginalized over $\textbf{f}_n^{(1)}$ up to $\textbf{f}_n^{(L-2)}$ but also marginalized over $h_n^{(1)}$ up to $h_n^{(L-1)}$---which has no closed form and needs to be evaluated sampling-based, similarly to the previous section. 

\begin{figure}[h!]
\centering
\includegraphics[trim=140 10 180 100,clip,width=0.7\textwidth]{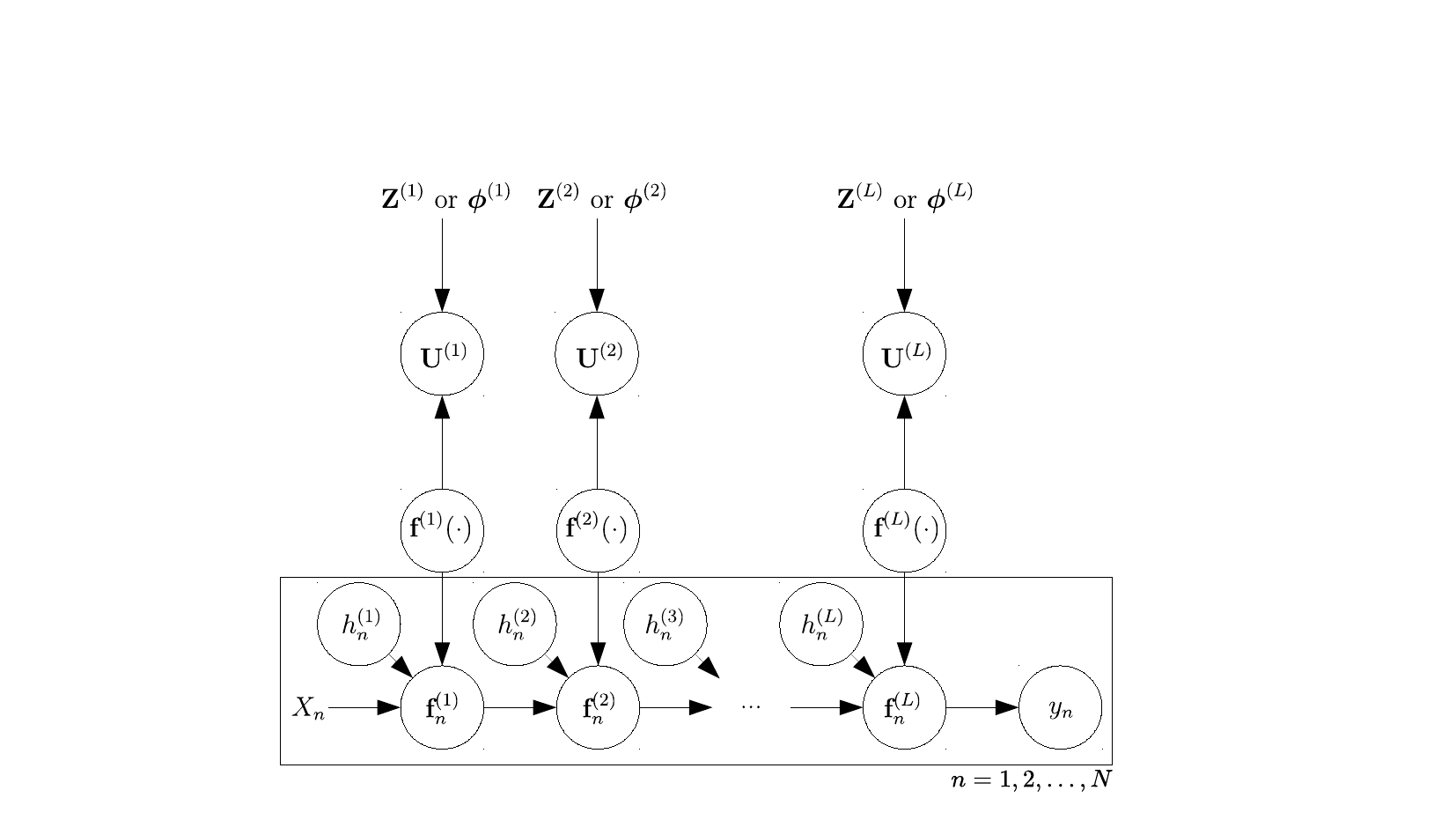}
\caption{Graphical model of a latent-variable deep sparse GP in an i.i.d.\ supervised learning setting. Labels and data points are denoted as $y_n$ and $X_n$ respectively. An input $X_n$ determines together with a sample from the first latent variable $h_n^{(1)}$ where to evaluate the first GP over vector-valued functions $\textbf{f}^{(1)}(\cdot)$. This yields the random variable $\textbf{f}_n^{(1)}$, a sample from which together with a sample from the second latent variable $h_n^{(2)}$ determines where to evaluate the second multioutput GP over $\textbf{f}_n^{(2)}(\cdot)$, and so forth. A sample from the last random variable $\textbf{f}_n^{(L)}$ in this cascade finally determines the probability of $y_n$. Inducing features are denoted as $\bm{\phi}$ and contain inducing points $\textbf{Z}$ as a special case (for Dirac features). Here, bold-face notation means that all $M$ inducing features/points are stored in one variable for each layer $l$ and shared across outputs for notational convenience. If we assume that inducing points/features are not shared across outputs, then each output would have its own set of $M$ inducing points/features. Under the ordinary inducing point formulation for example, $\mathbf{Z}$ would then be a three-dimensional tensor ($M \times \text{number of input dimensions} \times \text{number of outputs}$) rather than a matrix ($M \times \text{number of input dimensions}$). The associated inducing variables $\vU$ are matrices where the first dimension corresponds to the number of inducing features $M$ and the second dimension to the number of outputs (irrespective of whether inducing features are shared across outputs or not). The total number of inducing features $M$ is usually much smaller than the number of training examples $N$ for reasons of computational and memory efficiency.}
\label{fig:vi_deep_sgp_latentv}
\end{figure}

In Equation~\eqref{eq:elbo_deep_lv_sgp}, we again express the innermost expected log likelihood averaged over $\textbf{f}_n^{(L)}$ explicitly to highlight that this term can be computed efficiently under certain conditions---e.g.\ a Gaussian likelihood. The graphical model behind the generative process of a latent-variable deep sparse GP is depicted in Figure~\ref{fig:vi_deep_sgp_latentv}. It contains the generative model of an ordinary deep sparse GP without additional latent variables from the previous section as a special case when ignoring the nodes $h_n^{(l)}$.

Predicting new labels $\{y_n^\star\}_{n=1,..,N^\star}$ for new data $\{X_n^\star\}_{n=1,..,N^\star}$ is then similar to Equation~\eqref{eq:pred_deep_sparse_gp_iid}:
\begin{equation}
\label{eq:pred_deep_sparse_gp_iid_lv}
p(y_1^\star, ..., y_{N^\star}^\star | X_1^\star, ..., X_{N^\star}^\star) =  \mathbb{E}_{q_{\psi,\gamma}({\textbf{F}^\star}^{(L-1)}_{\text{incl.} \; h})} \Bigg[ \int \prod_{n=1}^{N^\star} p_\gamma(y^\star_n | {\textbf{F}^\star}^{(L)}[n,:]) q_{\psi,\gamma}({\textbf{F}^\star}^{(L)} | {\textbf{F}^\star}^{(L-1)}_{\text{incl.} \; h}) \; \mathrm{d} {\textbf{F}^\star}^{(L)} \Bigg] ,
\end{equation}
where the notation $\textbf{F}^\star$ refers to matrix-valued random variables that result from jointly propagating all new data points $\{X_n^\star\}_{n=1,..,N^\star}$ through the deep GP, and where the short-hand notation ${\textbf{F}^\star}^{(L-1)}_{\text{incl.} \; h} := (h_n^{(L)}, {\textbf{F}^\star}^{(L-1)})$ is introduced in an attempt to preserve a clear view.
At this stage, the reader should have obtained a good understanding of VI for deep sparse GP models with additional latent variables. What remains to be addressed is how to make latent-variable deep sparse GP models more optimization-efficient using the importance-weighting trick, as discussed next.

\subsection{Importance-Weighted Latent-Variable Deep Sparse GPs}
\label{sec:lv_deep_iw}

We have seen in Section~\ref{sec:lv_shallow_iw} how to use the importance-weighting trick for efficient VI in shallow sparse GP models with an additional latent variable. The motivation behind this trick is to trade computational resources to obtain a tighter lower bound to the log marginal likelihood and to achieve less variance during optimization when computing $\ELBO$ estimates. The idea is to introduce importance weights in such a way that the innermost expected log likelihood term of the $\ELBO$ objective is preserved, since this term can often be computed efficiently. 

In a deep sparse GP, the same logic applies to all latent variables from $h_n^{(1)}$ up to $h_n^{(L)}$ in accordance with Equation~\eqref{eq:ugly_log} from Section~\ref{sec:lv_shallow_iw}. A detailed derivation for a two-layer deep sparse GP with an additional latent variable in the first layer only (but not in the second) can be found in~\cite{Salimbeni2019}. Building on top of the latter, we present here the more general case for $L$ sparse GP layers. We stick with the setting of only one additional latent variable at the first layer (extending the formulation to an additional latent variable for each layer is conceptually straightforward but notationally cumbersome, and there is currently no empirical evidence that having more than one additional latent variable is beneficial). The derivation is a bit more subtle but similar to Section~\ref{sec:lv_shallow_iw} and follows the same steps as the two-layer setting from~\cite{Salimbeni2019}---the end result is:
\begin{eqnarray}
\ELBO_S(\gamma, \psi) &=& \sum_{n=1}^N \mathbb{E}_{\prod_{s=1}^S q_\psi(h_n^{(s)}) q_{\psi,\gamma}(\textbf{F}_n^{(L-1)} | \textbf{h}_n)} \Bigg[ \ln \frac{1}{S} \sum_{s=1}^S \explik_{\gamma,\psi}(y_n,\textbf{F}_n^{(L-1)}[s,:])  \frac{p_\gamma(h^{(s)}_n)}{q_\psi(h^{(s)}_n)} \Bigg] \nonumber \\
&& -\sum_{l=1}^L \KL\Big(q_\psi(\vU^{(l)}) \Big|\Big| p_{\psi,\gamma}(\vU^{(l)}) \Big) ,\label{eq:ugly_log_2}
\end{eqnarray}
where the notation $h_n^{(s)}$ refers to the $s$-th replication of the additional latent variable $h_n$ in the first layer for the data point $n$ (and $\textbf{h}_n$ is shorthand for the latent variable vector storing all replications for a given $n$). The matrix-valued random variable $\textbf{F}_n^{(L-1)}$ (of shape $S \times$ number of outputs) denotes the output of the deep GP at the second-to-last layer when evaluated jointly at all $S$ replications $(h_n^{(s)},X_n)$. In order to preserve a clear view, $q_{\psi,\gamma}(\textbf{F}_n^{(L-1)} | \textbf{h}_n)$ refers to the marginal distribution over the output $\textbf{F}_n^{(L-1)}$ at the second-to-last layer $L-1$ when marginalizing over all previous GP outputs, but conditioning on $\textbf{h}_n$ that sits in the first layer. Similarly to earlier, and to ease the notation, we needed to introduce the helper function $\explik_{\gamma,\psi}(y_n,\textbf{F}_n^{(L-1)}[s,:])$ defined as:
\begin{equation}
\label{eq:explik_2}
\explik_{\gamma,\psi}(y_n,\textbf{F}_n^{(L-1)}[s,:]) := \exp \Bigg( \int q_{\psi,\gamma}(\textbf{f}_n^{(L,s)} | \textbf{F}_n^{(L-1)}[s,:]) \ln p_\gamma(y_n|\textbf{f}_n^{(L,s)}) \; \mathrm{d} \textbf{f}_n^{(L,s)} \Bigg) ,
\end{equation} 
where $\textbf{f}_n^{(L,s)}$ refers to the output of the deep GP at the last layer $L$ and $q_{\psi,\gamma}(\textbf{f}_n^{(L,s)} | \textbf{F}_n^{(L-1)}[s,:])$ to its conditional distribution conditioned on the output of the second-to-last layer $\textbf{F}_n^{(L-1)}$ for a particular replication~$s$. Note that the GP sitting at the last layer $L$ can be evaluated for each individual replication $s$ independently because of the factorized form of the likelihood, whereas all other latent layers require a joint evaluation across all $S$ replications.

It can be insightful to continue at this point with a practical example that demonstrates the differences in modelling flexibility between shallow and deep sparse GPs\@. We do so in the next section where we also illustrate the effect of additional latent variables $h_n$ on the model's expressiveness.

\begin{figure}[h!]
\begin{subfigure}{.2\textwidth}
  \centering
  \text{Data}
  \includegraphics[width=.9\linewidth, height=.7\linewidth]{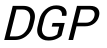}
  \label{fig:demo:gp}
\end{subfigure}%
\begin{subfigure}{.2\textwidth}
  \centering
  \text{GP}
  \includegraphics[width=.9\linewidth]{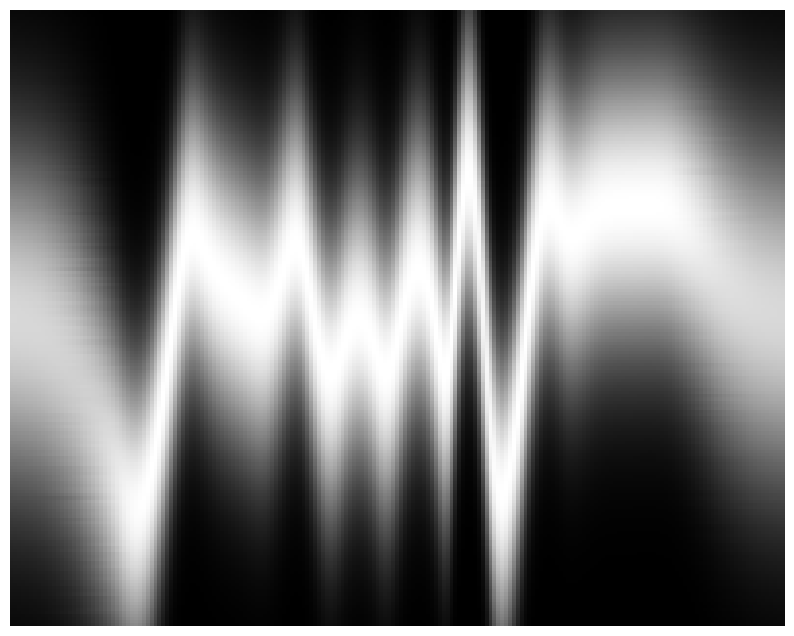}
  \label{fig:demo:gp}
\end{subfigure}%
\begin{subfigure}{.2\textwidth}
  \centering
  \text{DGP}
  \includegraphics[width=.9\linewidth]{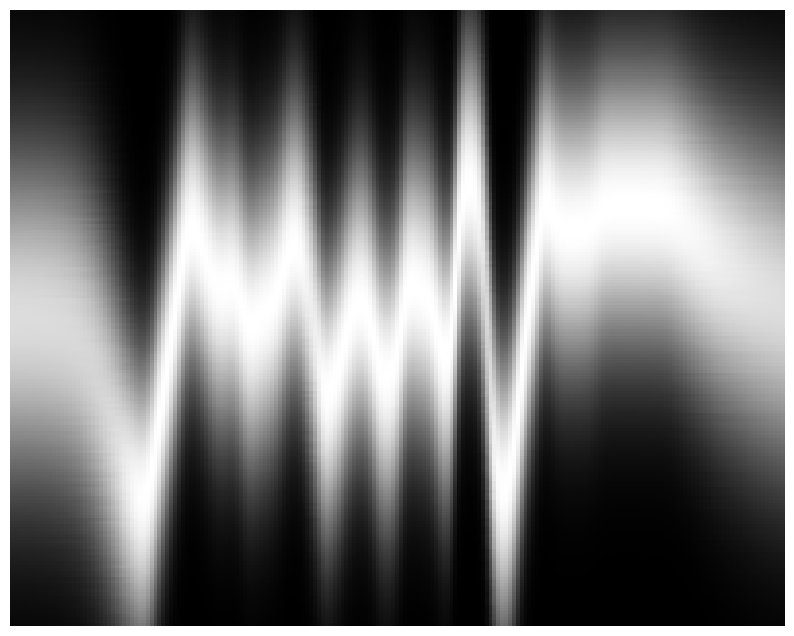}
  \label{fig:demo:gp-gp}
\end{subfigure}%
\begin{subfigure}{.2\textwidth}
  \centering
  \text{LV-GP}
  \includegraphics[width=.9\linewidth]{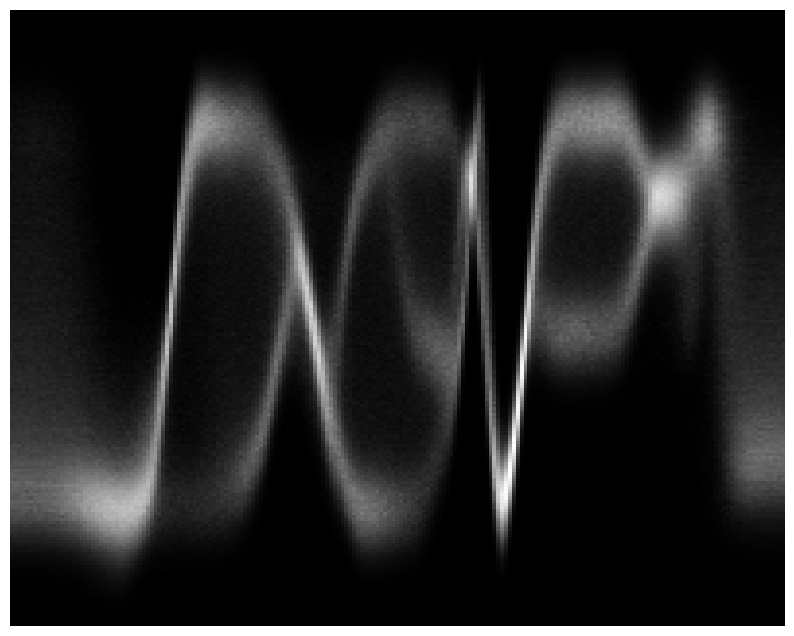}
  \label{fig:demo:lv-gp}
\end{subfigure}%
\begin{subfigure}{.2\textwidth}
  \centering
  \text{LV-DGP}
  \includegraphics[width=.9\linewidth]{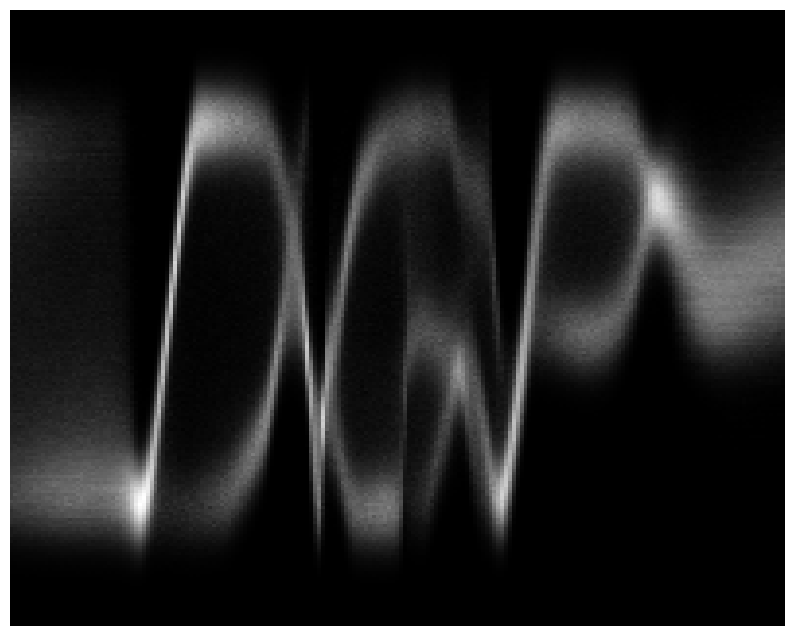}
  \label{fig:demo:lv-gp-gp}
\end{subfigure}%
\caption{Different GP models are compared to each other on a non-smooth multimodal regression problem following \cite{Salimbeni2019}. The first plot shows the training data set: data points (i.e.\ $(X,y)$-pairs) are given by the black pixels of the letters ``\textit{DGP}''. The second plot shows predictive samples from a fitted shallow sparse GP model using a prior zero-mean function and a prior RBF kernel (in white on a black background). The third plot shows predictions from a deep GP consisting of two GPs stacked on top of each other (DGP). The fourth plot shows a shallow latent-variable GP (LV-GP). The last plot extends the two-layer deep GP from the third plot by adding a latent variable to the first layer (LV-DGP). The plots demonstrate the importance of latent variables in complex tasks: the quality of the fit increases from the left to the right achieving best results when combining deep GPs and latent variables. See \url{https://github.com/vdutor/Toy_DGP_experiment} for code to reproduce the individual plots from this figure.}
\label{fig:demo}
\end{figure}

\subsection{Comparing Different Sparse GP Models on Synthetic Data}
\label{sec:demo}

While the previous sections introduced the theory behind sparse GP models and extensions thereof, this section aims at providing an illuminating example to demonstrate the effect of deep and latent-variable GP models. To this end, we fit four different types of models on a non-smooth and multimodal regression problem. The four models are
\begin{enumerate}
    \item a shallow sparse GP according to Section~\ref{sec:shallow},
    \item a two-layer deep sparse GP according to Section~\ref{sec:deep},
    \item a shallow sparse latent-variable GP according to Section~\ref{sec:lv_shallow_iw},
    \item and a two-layer deep sparse GP with a latent variable at the first layer (Section~\ref{sec:lv_deep_iw}).
\end{enumerate}
Note that the latent-variable models are trained with the importance-weighting trick as outlined earlier. The results are shown in Figure~\ref{fig:demo} and demonstrate the expressiveness obtained when combining deep GPs with latent variables (achieving the best result in this example). While the qualitative difference between shallow latent-variable and deep latent-variable sparse GPs might look minor in this example, the empirical results in~\cite{Salimbeni2019} provide strong evidence of a significant performance boost across a wide range of regression problems when making latent-variable GPs deep. Also note that for a deep GP, latent variables could be added to any layer (not just to the first as in our example). However, we follow here \cite{Salimbeni2019} and leave it to future work to investigate whether adding more latent variables improves performance even further.

\section{Summary}
\label{sec:conclusion}

The aim of this tutorial is to provide access to and an overview over sparse GPs, VI and how to do VI with sparse GP models, targeting a broad audience of readers that are not familiar with neither GPs nor VI\@. The idea behind the outline of the manuscript is to introduce GPs and sparse GPs in Section~\ref{sec:sparse_gps} explicitly as stand-alone models outside the scope of exact or approximate inference. Inference is then introduced in Section~\ref{sec:vi} and covers the main idea behind approximate inference with VI, however mostly focusing on weight space models for educational reasons. How to do VI with sparse GPs (that are function space models) is then finally covered in Section~\ref{sec:vi_with_svgps}. Note that while Section~\ref{sec:vi_with_svgps} requires Sections~\ref{sec:sparse_gps} and~\ref{sec:vi} as necessary prerequisites, Sections~\ref{sec:sparse_gps} and~\ref{sec:vi} can both be studied independently and separate from one another.

Section~\ref{sec:sparse_gps} introduces sparse GPs starting with multivariate Gaussian distributions and conditioning operations in multivariate Gaussians (that also yield Gaussians). This is necessary to understand sparse GPs which are essentially the consequence of conditioning GPs on a finite set of variables. It turns out that a proper exposition to the subject enables access to more recent advances in the field of GPs, such as interdomain GPs to provide more flexible features as well as computational gains, multioutput GPs that can handle problems with multidimensional labels, and deep GPs that hold the potential to model more abruptly changing functions (which is problematic for ordinary GPs).

Section~\ref{sec:vi} introduces the idea behind approximate inference using VI, with an emphasis on weight space models (such as neural networks for example) that we feel a large group of readers is familiar with (as opposed to function space models). We start with exact inference, its problems, and how to provide a remedy with vanilla VI\@. But we subsequently also cover more advanced topics like importance-weighted VI to increase the quality of the inference routine (at the cost of an increased computational complexity), latent-variable VI to enable more expressive modelling, up to presenting a generic framework for Bayesian deep learning with Bayesian layers (that stack approximate inference blocks on top of each other).

VI with sparse GP models is eventually introduced in Section~\ref{sec:vi_with_svgps}. We begin with VI for ordinary sparse GPs and continue with extensions to latent-variable sparse GPs, importance-weighted sparse GPs, and deep sparse GPs. In the end, we culminate in a challenging multimodal synthetic example that demonstrates the flexibility of modeling when combining all the previously mentioned extensions, more precisely importance-weighting and latent variables in combination with deep GPs. Along the way, we provide several useful recommendations for experimenters for how to train sparse GP models in practice.

\acknowledgments{Many thanks to Vincent Adam for providing feedback to an early version of this manuscript.}


\appendix

\section{Appendix}

\subsection{Latent-Variable Variational Inference in Unsupervised Learning}
\label{sec:unsupervised_vi}

\begin{figure}[h!]
\centering
\includegraphics[trim=190 130 220 100,clip,width=0.6\textwidth]{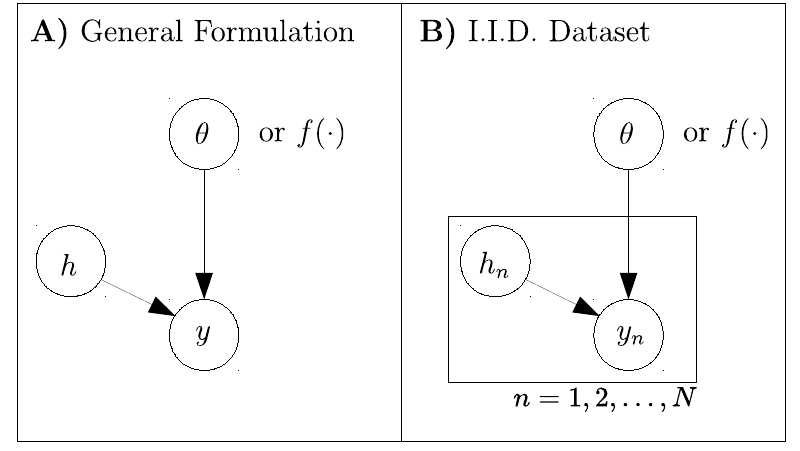}
\caption{Graphical models for latent-variable VI in unsupervised learning settings. Unknown functions are denoted as $\theta$ (in parameter space view) or $f(\cdot)$ (in function space view). They receive the additional latent variable $h$ as input and map it probabilistically to the observed output $y$. \textbf{A)} refers to the ordinary setting and \textbf{B)} to the i.i.d.\ setting where $N$ different examples are index by $n$.}
\label{fig:vi_unsup}
\end{figure}

In unsupervised learning, only observations $y$ are available but no inputs $X$, and the goal is to infer parameters of a generative model that can mimic the true underlying distribution of the observations~$y$. Throughout this tutorial, we have assumed a supervised learning scenario because it contains the unsupervised setting (where input variables $X$ are missing) as a special case. Deriving the vanilla $\ELBO$ from Section~\ref{sec:vanilla_vi} for unsupervised learning is hence trivial, since all we need to do is to ``delete'' $X$ in Equation~\eqref{eq:elbo}. While the same is true for latent-variable VI, we nevertheless decided to dedicate a separate section to this topic at the end of this tutorial because of its importance in contemporary machine learning literature. The advantage of generative latent-variable models is that they can represent challenging distributions over $y$, e.g.\ containing multiple modes. The corresponding graphical model is, not surprisingly, similar to latent-variable VI for supervised learning from Section~\ref{sec:lv_vi}---compare Figure~\ref{fig:vi_unsup} A) to Figure~\ref{fig:vi} B) that only differ by the input variable $X$. The corresponding $\ELBO$ can be readily obtained from Equation~\eqref{eq:elbo_lv} by ``ignoring'' $X$, yielding:
\begin{eqnarray}
\ELBO(\gamma, \psi) &=& \int \int q_\psi(\theta) q_\psi(h) \ln p_\gamma(y | \theta, h)  \; \mathrm{d} h \;  \mathrm{d}\theta \nonumber \\
&& -\KL\Big(q_\psi(h) \Big|\Big| p_\gamma(h) \Big) -\KL\Big(q_\psi(\theta) \Big|\Big| p_\gamma(\theta) \Big), \label{eq:elbo_lv_unsup}
\end{eqnarray}
where the likelihood term $p_\gamma(y | \theta, h)$ does not contain any other context variable apart from $\theta$ and $h$.
Generating new samples $y^\star$ is then achieved via:
\begin{equation}
\label{eq:pred_lv_unsup}
p(y^\star) =  \int \int p_\gamma(y^\star | \theta, h) p_\gamma(h) \; \mathrm{d} h \; q_\psi(\theta) \; \mathrm{d}\theta .
\end{equation}

Practically, one usually assumes i.i.d.\ data $\{y_n\}_{n=1,..,N}$ as indicated by the graphical model in Figure~\ref{fig:vi_unsup}~B). In line with the supervised formulation, the additional latent variable $h_n$ is also assumed i.i.d.\ across training examples and is indexed with $n$. The latter leads to a factorized likelihood which induces the following $\ELBO$:
\begin{eqnarray}
\ELBO(\gamma, \psi) &=& \sum_{n=1}^N \int \int q_\psi(\theta) q_\psi(h_n) \ln p_\gamma(y_n | \theta, h_n) \; \mathrm{d} h_n \;  \mathrm{d}\theta \nonumber \\
&& -\sum_{n=1}^N \KL\Big(q_\psi(h_n) \Big|\Big| p_\gamma(h_n) \Big) -\KL\Big(q_\psi(\theta) \Big|\Big| p_\gamma(\theta) \Big), \label{eq:elbo_lv_iid_unsup}
\end{eqnarray}
Generating new observations $\{y_n^\star\}_{n=1,..,N^\star}$ is then accomplished via:
\begin{equation}
\label{eq:pred_lv_iid_unsup}
p(y_1^\star, ..., y_{N^\star}^\star) =  \int \prod_{n=1}^{N^\star}  \int p_\gamma(y_n^\star | \theta, h_n) p_\gamma(h_n) \; \mathrm{d} h_n \; q_\psi(\theta) \; \mathrm{d}\theta ,
\end{equation}
where $h_n$ needs to be integrated out with the prior $p_\gamma(h_n)$. This is in line with the supervised setting, where there is either a separate approximate posterior $q_\psi(h_n)$ for each individual training example $y_n$ or an amortized approximate posterior $q_\psi(h_n|y_n)$ that maps observations $y_n$ to latent variables $h_n$. Both parameterizations do not readily generalize to new observations $y_n^\star$ as they are either implicitly or explicitly conditioned on training observations $y_n$.
The approximate posterior $q_\psi(h_n)$ serves hence only as auxiliary training tool and is typically ``thrown away'' after learning.

Similar to earlier sections, it can be insightful to provide some concrete examples after presenting a general VI formulation. Imagine to that end an i.i.d.\ scenario where $\bm{\theta}$ represents the weights of a deep neural network in vectorized form, and where the prior $p(\bm{\theta})$ can be a multivariate Gaussian with diagonal covariance matrix. Imagine furthermore that $\textbf{h}_n$ is a multivariate random variable as well, with a multivariate standard normal Gaussian prior $p(\textbf{h}_n)$. Let the likelihood $p_\gamma(y_n | \bm{\theta}, \textbf{h}_n)$ be homoscedastic Gaussian with variance $\upsilon_{\textrm{lik}}^{(\gamma)}$ and where $\gamma$ indicates generative parameters as previously. Importantly, let the mean of the likelihood $\mu_{\textrm{lik}}(\textbf{h}_n) = f_{\bm{\theta}}(\textbf{h}_n)$ be a neural net function that receives $\textbf{h}_n$ as input. We could then parameterize the approximate posterior over network weights $q_\psi(\bm{\theta})$ as a mean field multivariate Gaussian, and the approximate posterior over the additional latent variables $q_\psi(\textbf{h}_n | y_n)$ in an amortized way via a neural net that maps $y_n$ probabilistically to $\textbf{h}_n$. In accordance with previous examples, $\psi$ refers to the entirety of all variational parameters. We have just arrived at the general formulation of a variational auto-encoder according to the appendix of~\cite{Kingma2014}, although using a homoscedastic rather than a heteroscedastic likelihood to ease the exposition. We can readily replace the neural net prior $p(\bm{\theta})$ and the corresponding approximate posterior $q_\psi(\bm{\theta})$ with GPs that operate on the domain where the additional latent variables $\textbf{h}_n$ reside. This requires us to replace $f_{\bm{\theta}}(\textbf{h}_n)$ with $f(\textbf{h}_n)$ in the likelihood where $f(\cdot)$ is a GP random function. The latter leads us to a latent-variable GP model similar to the work of~\cite{Damianou2016}.




\bibliography{gp_tutorial}  

\begin{thebibliography}{48}
\providecommand{\natexlab}[1]{#1}
\providecommand{\url}[1]{\texttt{#1}}
\expandafter\ifx\csname urlstyle\endcsname\relax
  \providecommand{\doi}[1]{doi: #1}\else
  \providecommand{\doi}{doi: \begingroup \urlstyle{rm}\Url}\fi

\bibitem[Adam et~al.(2020)Adam, Eleftheriadis, Artemev, Durrande, and
  Hensman]{Adam2020}
V~Adam, S~Eleftheriadis, A~Artemev, N~Durrande, and J~Hensman.
\newblock {Doubly sparse variational Gaussian processes}.
\newblock In \emph{Proceedings of the International Conference on Artificial
  Intelligence and Statistics}, 2020.

\bibitem[Alvarez and Lawrence(2008)]{Alvarez2008}
M~A Alvarez and N~D Lawrence.
\newblock {Sparse convolved Gaussian processes for multi-output regression}.
\newblock In \emph{Advances in Neural Information Processing Systems}, 2008.

\bibitem[Alvarez et~al.(2009)Alvarez, Luengo, and Lawrence]{Alvarez2009}
M~A Alvarez, D~Luengo, and N~D Lawrence.
\newblock {Latent force models}.
\newblock In \emph{Proceedings of the International Conference on Artificial
  Intelligence and Statistics}, 2009.

\bibitem[Alvarez et~al.(2010)Alvarez, Luengo, Titsias, and
  Lawrence]{Alvarez2010}
M~A Alvarez, D~Luengo, M~Titsias, and N~D Lawrence.
\newblock {Efficient multioutput Gaussian processes through variational
  inducing kernels}.
\newblock In \emph{Proceedings of the International Conference on Artificial
  Intelligence and Statistics}, 2010.

\bibitem[Alvarez et~al.(2012)Alvarez, Rosasco, and Lawrence]{Alvarez2012}
M~A Alvarez, L~Rosasco, and N~D Lawrence.
\newblock {Kernels for vector-valued functions: a review}.
\newblock \emph{Foundations and Trends in Machine Learning}, 2012.

\bibitem[Bishop(2006)]{Bishop2006}
C~M Bishop.
\newblock \emph{{Pattern Recognition and Machine Learning}}.
\newblock Springer, 2006.

\bibitem[Blomqvist et~al.(2019)Blomqvist, Kaski, and Heinonen]{Blomqvist2019}
K~Blomqvist, S~Kaski, and M~Heinonen.
\newblock {Deep convolutional Gaussian processes}.
\newblock In \emph{Proceedings of the European Conference on Machine Learning
  and Principles and Practice of Knowledge Discovery in Databases}, 2019.

\bibitem[Blundell et~al.(2015)Blundell, Cornebise, Kavukcuoglu, and
  Wierstra]{Blundell2015}
C~Blundell, J~Cornebise, K~Kavukcuoglu, and D~Wierstra.
\newblock {Weight uncertainty in neural networks}.
\newblock In \emph{Proceedings of the International Conference on Machine
  Learning}, 2015.

\bibitem[Borovitskiy et~al.(2020)Borovitskiy, Terenin, Mostowsky, and
  Deisenroth]{Borovitskiy2020}
V~Borovitskiy, A~Terenin, P~Mostowsky, and M~P Deisenroth.
\newblock {Matern Gaussian processes on Riemannian manifolds}.
\newblock \emph{arXiv}, 2020.

\bibitem[Boyle and Frean(2004)]{Boyle2004}
P~Boyle and M~Frean.
\newblock {Dependent Gaussian processes}.
\newblock In \emph{Advances in Neural Information Processing Systems}, 2004.

\bibitem[Bui et~al.(2017)Bui, Yan, and Turner]{Bui2017}
T~D Bui, J~Yan, and R~E Turner.
\newblock {A unifying framework for Gaussian process pseudo-point
  approximations using power expectation propagation}.
\newblock \emph{Journal of Machine Learning Research}, 2017.

\bibitem[Burda et~al.(2016)Burda, Grosse, and Salakhutdinov]{Burda2016}
Y~Burda, R~B Grosse, and R~Salakhutdinov.
\newblock {Importance weighted autoencoders}.
\newblock In \emph{Proceedings of the International Conference on Learning
  Representations}, 2016.

\bibitem[Burt et~al.(2019)Burt, Rasmussen, and van~der Wilk]{Burt2019}
D~R Burt, C~E Rasmussen, and M~van~der Wilk.
\newblock {Rates of convergence for sparse variational Gaussian process
  regression}.
\newblock In \emph{Proceedings of the International Conference on Machine
  Learning}, 2019.

\bibitem[Burt et~al.(2020)Burt, Rasmussen, and van~der Wilk]{Burt2020}
D~R Burt, C~E Rasmussen, and M~van~der Wilk.
\newblock {Variational orthogonal features}.
\newblock \emph{arXiv}, 2020.

\bibitem[Damianou and Lawrence(2013)]{Damianou2013}
A~Damianou and N~D Lawrence.
\newblock {Deep Gaussian processes}.
\newblock In \emph{Proceedings of the International Conference on Artificial
  Intelligence and Statistics}, 2013.

\bibitem[Damianou et~al.(2016)Damianou, Titsias, and Lawrence]{Damianou2016}
A~Damianou, M~Titsias, and N~D Lawrence.
\newblock {Variational inference for latent variables and uncertain inputs in
  Gaussian processes}.
\newblock \emph{Journal of Machine Learning Research}, 2016.

\bibitem[Domke and Sheldon(2018)]{Domke2018}
J~Domke and D~Sheldon.
\newblock {Importance weighting and variational inference}.
\newblock In \emph{Advances in Neural Information Processing Systems}, 2018.

\bibitem[Dutordoir et~al.(2018)Dutordoir, Salimbeni, Deisenroth, and
  Hensman]{Dutordoir2018}
V~Dutordoir, H~Salimbeni, M~P Deisenroth, and J~Hensman.
\newblock {Gaussian process conditional density estimation}.
\newblock In \emph{Advances in Neural Information Processing Systems}, 2018.

\bibitem[Dutordoir et~al.(2020{\natexlab{a}})Dutordoir, Durrande, and
  Hensman]{Dutordoir2020}
V~Dutordoir, N~Durrande, and J~Hensman.
\newblock {Sparse Gaussian processes with spherical harmonic features}.
\newblock In \emph{Proceedings of the International Conference on Machine
  Learning}, 2020{\natexlab{a}}.

\bibitem[Dutordoir et~al.(2020{\natexlab{b}})Dutordoir, van~der Wilk, Artemev,
  and Hensman]{Dutordoir2020b}
V~Dutordoir, M~van~der Wilk, A~Artemev, and J~Hensman.
\newblock {Bayesian image classification with deep convolutional Gaussian
  processes}.
\newblock In \emph{Proceedings of the International Conference on Artificial
  Intelligence and Statistics}, 2020{\natexlab{b}}.

\bibitem[Duvenaud et~al.(2014)Duvenaud, Rippel, Adams, and
  Ghahramani]{Duvenaud2014}
D~Duvenaud, O~Rippel, R~P Adams, and Z~Ghahramani.
\newblock {Avoiding pathologies in very deep networks}.
\newblock In \emph{Proceedings of the International Conference on Artificial
  Intelligence and Statistics}, 2014.

\bibitem[Gardner et~al.(2018)Gardner, Pleiss, Bindel, Weinberger, and
  Wilson]{Gardner2018}
J~R Gardner, G~Pleiss, D~Bindel, K~Q Weinberger, and A~G Wilson.
\newblock {GPyTorch: Blackbox matrix-matrix Gaussian process inference with GPU
  acceleration}.
\newblock In \emph{Advances in Neural Information Processing Systems}, 2018.

\bibitem[Hensman et~al.(2013)Hensman, Fusi, and Lawrence]{Hensman2013}
J~Hensman, N~Fusi, and N~D Lawrence.
\newblock {Gaussian processes for big data}.
\newblock In \emph{Proceedings of the Conference on Uncertainty in Artificial
  Intelligence}, 2013.

\bibitem[Hensman et~al.(2015{\natexlab{a}})Hensman, Matthews, Filippone, and
  Ghahramani]{Hensman2015b}
J~Hensman, A~G~G Matthews, M~Filippone, and Z~Ghahramani.
\newblock {MCMC for variationally sparse Gaussian processes}.
\newblock In \emph{Advances in Neural Information Processing Systems},
  2015{\natexlab{a}}.

\bibitem[Hensman et~al.(2015{\natexlab{b}})Hensman, Matthews, and
  Ghahramani]{Hensman2015}
J~Hensman, A~G~G Matthews, and Z~Ghahramani.
\newblock {Scalable variational Gaussian process classification}.
\newblock \emph{Journal of Machine Learning Research}, 2015{\natexlab{b}}.

\bibitem[Hensman et~al.(2018)Hensman, Durrande, and Solin]{Hensman2018}
J~Hensman, N~Durrande, and A~Solin.
\newblock {Variational Fourier features for Gaussian processes}.
\newblock \emph{Journal of Machine Learning Research}, 2018.

\bibitem[Higdon(2002)]{Higdon2002}
D~Higdon.
\newblock {Space and space-time modeling using process convolutions}.
\newblock \emph{Quantitative Methods for Current Environmental Issues}, 2002.

\bibitem[Higgins et~al.(2017)Higgins, Matthey, Pal, Burgess, Glorot, Botvinick,
  Mohamed, and Lerchner]{Higgins2017}
I~Higgins, L~Matthey, A~Pal, C~Burgess, X~Glorot, M~Botvinick, S~Mohamed, and
  A~Lerchner.
\newblock {Beta-VAE: Learning basic visual concepts with a constrained
  variational framework}.
\newblock In \emph{Proceedings of the International Conference on Learning
  Representations}, 2017.

\bibitem[Journel and Huijbregts(1978)]{Journel1978}
A~G Journel and C~J Huijbregts.
\newblock \emph{{Mining Geostatistics}}.
\newblock Academic Press, 1978.

\bibitem[Kingma and Welling(2014)]{Kingma2014}
D~P Kingma and M~Welling.
\newblock {Auto-encoding variational Bayes}.
\newblock In \emph{Proceedings of the International Conference on Learning
  Representations}, 2014.

\bibitem[Kingma et~al.(2015)Kingma, Rezende, Mohamed, and Welling]{Kingma2015}
D~P Kingma, D~J Rezende, S~Mohamed, and M~Welling.
\newblock {Semi-supervised learning with deep generative models}.
\newblock In \emph{Advances in Neural Information Processing Systems}, 2015.

\bibitem[Lazaro-Gredilla and Figueiras-Vidal(2009)]{Lazaro2009}
M~Lazaro-Gredilla and A~Figueiras-Vidal.
\newblock {Inter-domain Gaussian processes for sparse inference using inducing
  features}.
\newblock In \emph{Advances in Neural Information Processing Systems}, 2009.

\bibitem[Matthews et~al.(2016)Matthews, Hensman, Turner, and
  Ghahramani]{Matthews2016}
A~G~G Matthews, J~Hensman, R~Turner, and Z~Ghahramani.
\newblock {On sparse variational methods and the Kullback-Leibler divergence
  between stochastic processes}.
\newblock In \emph{Proceedings of the International Conference on Artificial
  Intelligence and Statistics}, 2016.

\bibitem[Matthews et~al.(2017)Matthews, van~der Wilk, Nickson, Fujii,
  Boukouvalas, Leon-Villagra, Ghahramani, and Hensman]{Matthews2017}
A~G~G Matthews, M~van~der Wilk, T~Nickson, K~Fujii, A~Boukouvalas,
  P~Leon-Villagra, Z~Ghahramani, and J~Hensman.
\newblock {GPflow: A Gaussian process library using TensorFlow}.
\newblock \emph{Journal of Machine Learning Research}, 2017.

\bibitem[Micchelli and Pontil(2005)]{Micchelli2005}
C~A Micchelli and M~Pontil.
\newblock {On learning vector-valued functions}.
\newblock \emph{Neural Computation}, 2005.

\bibitem[O'Hagan(1992)]{OHagan1992}
A~O'Hagan.
\newblock {Some Bayesian numerical analysis}.
\newblock \emph{Bayesian Statistics}, 1992.

\bibitem[Rasmussen and Williams(2006)]{Rasmussen2006}
C~E Rasmussen and C~K~I Williams.
\newblock \emph{{Gaussian Processes for Machine Learning}}.
\newblock MIT Press, 2006.

\bibitem[Rezende et~al.(2014)Rezende, Mohamed, and Wierstra]{Rezende2014}
D~J Rezende, S~Mohamed, and D~Wierstra.
\newblock {Stochastic backpropagation and approximate inference in deep
  generative models}.
\newblock In \emph{Proceedings of the International Conference on Machine
  Learning}, 2014.

\bibitem[Riutort-Mayol et~al.(2020)Riutort-Mayol, Buerkner, Andersen, Solin,
  and Vehtari]{Riutort2020}
G~Riutort-Mayol, P-C Buerkner, M~R Andersen, A~Solin, and A~Vehtari.
\newblock {Practical Hilbert space approximate Bayesian Gaussian processes for
  probabilistic programming}.
\newblock \emph{Computing Research Repository}, 2020.

\bibitem[Salimbeni and Deisenroth(2017)]{Salimbeni2017}
H~Salimbeni and M~P Deisenroth.
\newblock {Doubly stochastic variational inference for deep Gaussian
  processes}.
\newblock In \emph{Advances in Neural Information Processing Systems}, 2017.

\bibitem[Salimbeni et~al.(2019)Salimbeni, Dutordoir, Hensman, and
  Deisenroth]{Salimbeni2019}
H~Salimbeni, V~Dutordoir, J~Hensman, and M~P Deisenroth.
\newblock {Deep Gaussian processes with importance-weighted variational
  inference}.
\newblock In \emph{Proceedings of the International Conference on Machine
  Learning}, 2019.

\bibitem[Saul et~al.(2016)Saul, Hensman, Vehtari, and Lawrence]{Saul2016}
A~D Saul, J~Hensman, A~Vehtari, and N~D Lawrence.
\newblock {Chained Gaussian processes}.
\newblock In \emph{Proceedings of the International Conference on Artificial
  Intelligence and Statistics}, 2016.

\bibitem[Sohn et~al.(2015)Sohn, Lee, and Yan]{Sohn2015}
K~Sohn, H~Lee, and X~Yan.
\newblock {Learning structured output representation using deep conditional
  generative models}.
\newblock \emph{Advances in Neural Information Processing Systems}, 2015.

\bibitem[Titsias(2009)]{Titsias2009}
M~Titsias.
\newblock {Variational learning of inducing variables in sparse Gaussian
  processes}.
\newblock In \emph{Proceedings of the International Conference on Artificial
  Intelligence and Statistics}, 2009.

\bibitem[Tran et~al.(2019)Tran, Dusenberry, van~der Wilk, and Hafner]{Tran2019}
D~Tran, M~W Dusenberry, M~van~der Wilk, and D~Hafner.
\newblock {Bayesian layers: A module for neural network uncertainty}.
\newblock In \emph{Advances in Neural Information Processing Systems}, 2019.

\bibitem[van~der Wilk and Rasmussen(2017)]{vanderWilk2017}
M~van~der Wilk and C~E Rasmussen.
\newblock {Convolutional Gaussian processes}.
\newblock In \emph{Advances in Neural Information Processing Systems}, 2017.

\bibitem[van~der Wilk et~al.(2020)van~der Wilk, Dutordoir, John, Artemev, Adam,
  and Hensman]{vanderWilk2020}
M~van~der Wilk, V~Dutordoir, S~T John, A~Artemev, V~Adam, and J~Hensman.
\newblock {A framework for interdomain and multioutput Gaussian processes}.
\newblock \emph{arXiv}, 2020.

\bibitem[Wenzel et~al.(2020)Wenzel, Roth, Veeling, Swiatkowski, Tran, Mandt,
  Snoek, Salimans, Jenatton, and Nowozin]{Wenzel2020}
F~Wenzel, K~Roth, B~Veeling, J~Swiatkowski, L~Tran, S~Mandt, J~Snoek,
  T~Salimans, R~Jenatton, and S~Nowozin.
\newblock {How good is the Bayes posterior in deep neural networks really?}
\newblock In \emph{Proceedings of the International Conference on Machine
  Learning}, 2020.

\end{thebibliography}

\end{document}